\documentclass[10pt,letterpaper]{article}

\usepackage[hmargin=1.1in,vmargin=1in,headheight=18pt,headsep=14pt,footskip=28pt]{geometry}

\usepackage[T1]{fontenc}
\usepackage{amsmath,amsthm,mathtools}
\usepackage{newpxtext}\usepackage{newpxmath}
\usepackage[semibold]{sourcesanspro}
\usepackage[varqu,scaled=0.95]{zi4}

\usepackage{array}
\usepackage{booktabs}
\usepackage{graphicx}
\usepackage{wrapfig}
\usepackage{enumitem}
\usepackage{needspace}
\usepackage{microtype}
\usepackage{algorithm}
\usepackage[noend]{algpseudocode}
\usepackage{xcolor}
\usepackage[most]{tcolorbox}
\usepackage{fontawesome5}
\usepackage{titlesec}

\definecolor{surface}{HTML}{FFFFFF}
\definecolor{titlebg}{HTML}{F9F2EE}
\definecolor{surfaceelev}{HTML}{F1F2EF}
\definecolor{border}{HTML}{DDDFDA}
\definecolor{bordersubtle}{HTML}{E9EAE6}
\definecolor{textprimary}{HTML}{1D1F21}
\definecolor{textsecondary}{HTML}{5F646B}
\definecolor{textmuted}{HTML}{858A90}
\definecolor{accent}{HTML}{A94F24}
\definecolor{accenthover}{HTML}{8E401D}
\definecolor{accentstrong}{HTML}{783719}
\definecolor{success}{HTML}{397A54}
\definecolor{warning}{HTML}{956A22}
\definecolor{danger}{HTML}{A94444}
\definecolor{info}{HTML}{477396}

\algrenewcommand{\algorithmiccomment}[1]{\hfill\textcolor{accent}{\#\ #1}}

\newcommand{\gdelta}[1]{{\scriptsize\textcolor{textmuted}{(#1)}}}

\newcommand{\tdot}{{\color{warning}\Large\textbullet}}
\newcommand{\tdash}{{\color{textmuted}$-$}}
\newcommand{\tgood}{{\color{success}\ensuremath{\checkmark}}}
\newcommand{\tno}{{\color{danger}\ensuremath{\times}}}

\titleformat{\section}{\sffamily\large\bfseries}{\textcolor{accent}{\thesection}}{0.6em}{}
\titleformat{\subsection}{\sffamily\normalsize\bfseries}{\textcolor{accent}{\thesubsection}}{0.6em}{}
\titleformat{\subsubsection}{\sffamily\normalsize\bfseries}{\textcolor{accent}{\thesubsubsection}}{0.5em}{}
\titleformat{\paragraph}[runin]{\sffamily\normalsize\bfseries}{}{0pt}{}
\titlespacing*{\section}{0pt}{16pt plus 4pt minus 2pt}{6pt plus 2pt}
\titlespacing*{\subsection}{0pt}{12pt plus 3pt minus 2pt}{4pt plus 1pt}
\titlespacing*{\subsubsection}{0pt}{10pt plus 2pt minus 2pt}{3pt}
\titlespacing*{\paragraph}{0pt}{6pt plus 2pt minus 1pt}{0.7em}

\makeatletter
\def\ps@firstpage{%
  \def\@oddhead{\hfill
    \raisebox{-0.5pt}{\includegraphics[height=14pt]{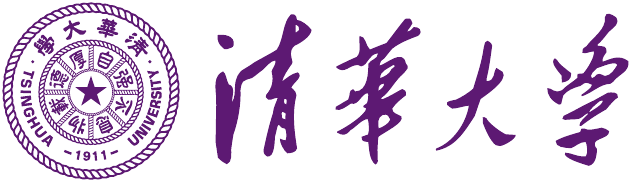}}%
    \hspace{9pt}{\color{border}\rule[-1pt]{0.5pt}{13pt}}\hspace{9pt}%
    \raisebox{2.6pt}{\includegraphics[height=7pt]{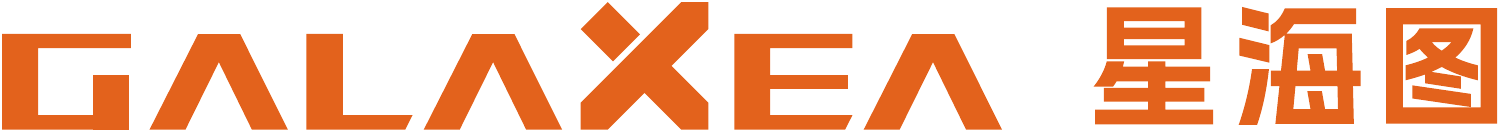}}%
    \llap{\color{textmuted}\rule[-7pt]{\textwidth}{0.4pt}}}%
  \let\@evenhead\@oddhead
  \def\@oddfoot{\hfil{\sffamily\small\color{textmuted}\thepage}\hfil}%
  \let\@evenfoot\@oddfoot}
\def\ps@body{%
  \let\@oddhead\@empty \let\@evenhead\@empty
  \def\@oddfoot{\hfil{\sffamily\small\color{textmuted}\thepage}\hfil}%
  \let\@evenfoot\@oddfoot}
\makeatother
\usepackage[font=small,labelfont={sf,bf,color=accent}]{caption}
\usepackage{natbib}
\setcitestyle{numbers,square,citesep={,}}
\usepackage{hyperref}
\usepackage{url}
\hypersetup{
  colorlinks=true,
  citecolor=accent,
  linkcolor=info,
  urlcolor=accenthover,
  pdftitle={ActionCodec2: A Streaming Action Tokenizer},
  pdfauthor={Zibin Dong, Yicheng Liu, Shiduo Zhang, Tianyuan Yuan, Baijun Ye, Jinqiu Wang, Hang Zhao},
}

\newtcolorbox{titlepanel}{
  enhanced, colback=titlebg, frame hidden, boxrule=0pt, arc=5pt,
  left=18pt, right=18pt, top=16pt, bottom=12pt,
  before skip=0pt, after skip=14pt,
}
\newcommand{\linkbtn}[3]{\href{#1}{\tcbox[on line, enhanced, colback=white, colframe=accent!35!white,
  boxrule=0.5pt, arc=6pt, boxsep=0pt, left=7pt, right=8pt, top=3pt, bottom=3pt]%
  {{\sffamily\small\color{accentstrong}{#2}\,#3}}}}
\newcommand{\metaitem}[2]{{\sffamily\small\textbf{#1:}~#2}}

\newif\ifpflink \pflinktrue
\newcommand{\pfdest}{proofs}
\newcommand{\proofin}[1]{\renewcommand{\pfdest}{#1}\pflinktrue}
\newcommand{\proofhere}{\pflinkfalse}
\newtheoremstyle{linked}
  {\topsep}{\topsep}
  {\itshape}{}
  {\bfseries}{.}
  {5pt plus 1pt minus 1pt}
  {\thmname{#1}\thmnumber{ \ifpflink\hyperlink{\pfdest}{#2}\else#2\fi}\thmnote{ (#3)}}
\theoremstyle{linked}
\newtheorem{theorem}{Theorem}
\newtheorem{proposition}[theorem]{Proposition}
\newtheorem{lemma}[theorem]{Lemma}
\newtheorem{corollary}[theorem]{Corollary}
\theoremstyle{definition}

\theoremstyle{remark}
\newtheorem{remark}[theorem]{Remark}

\newcommand{\R}{\mathbb{R}}

\newcommand{\SO}{\operatorname{SO}}

\newcommand{\supp}{\operatorname{supp}}
\newcommand{\Exp}{\operatorname{Exp}}

\newcommand{\argmin}{\operatorname*{arg\,min}}

\newcommand{\Vol}{\operatorname{Vol}}

\begin{document}
\color{textprimary}
\thispagestyle{firstpage}

\begin{titlepanel}
{\raggedright\sffamily\bfseries\fontsize{20}{24}\selectfont ActionCodec2: A Streaming Action Tokenizer\par}
\vspace{10pt}
{\raggedright\sffamily\bfseries
\mbox{Zibin Dong\textsuperscript{1,2}},\quad
\mbox{Yicheng Liu\textsuperscript{1,2}},\quad
\mbox{Shiduo Zhang\textsuperscript{2}},\quad
\mbox{Tianyuan Yuan\textsuperscript{1,2}},\quad
\mbox{Baijun Ye\textsuperscript{1,2}},\quad
\mbox{Jinqiu Wang\textsuperscript{2,3}},\quad
\mbox{Hang Zhao\textsuperscript{1,2,\textdagger}}\par}
\vspace{4pt}
{\sffamily\small\color{textsecondary}\textsuperscript{1}Tsinghua University,\quad \textsuperscript{2}Galaxea AI,\quad \textsuperscript{3}Fudan University\par}
\vspace{1pt}
{\sffamily\footnotesize\color{textmuted}\textsuperscript{\textdagger}Corresponding author\par}
\vspace{10pt}
Most action tokenizers for autoregressive vision-language-action (VLA) models compress motion into chunks that must be decoded in full before their actions become executable. We introduce ActionCodec2, a non-neural streaming tokenizer that releases actions incrementally while keeping trajectory reconstruction error within a prescribed tolerance. Its encoder quantizes each step relative to the decoder’s reconstructed state, so the selected motion primitive corrects residual error rather than allowing errors from independently quantized increments to accumulate. Finding the smallest such primitive set is a geometric covering problem; we derive the exact minimum for translation and constant-factor-tight bounds for rotation. Set-BPE then losslessly merges co-occurring primitive symbols across coordinates and neighboring steps, without a fixed coordinate ordering. A universal version fitted on 20k hours of data from 24 embodiments averages 8.8 tokens per second for 15 Hz end-effector motion on this corpus. With the same VLA backbone, ActionCodec2 outperforms the compared tokenizers on three simulation benchmarks and in zero-shot real-robot evaluation. Its streaming representation allows policies to incorporate state feedback within an action chunk, substantially improving performance over open-loop execution under randomized dynamics, external disturbances, and moving objects. The same tokens also can serve as a direct control interface for VLM agents, which compose successful action sequences into reusable macros. These agent-built macros reduce control decisions and thinking-token usage across attempts and transfer to a held-out task. More results are available at~ \url{https://zibindong.github.io/actioncodec2-page/}.

\par\vspace{10pt}
{\color{accent!25!white}\rule{\linewidth}{0.4pt}}\par\vspace{0pt}
\metaitem{Date}{October 2026}\par\vspace{7pt}
\linkbtn{https://huggingface.co/collections/ZibinDong/actioncodec2}{\raisebox{-1.5pt}{\includegraphics[height=1.1em]{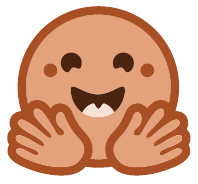}}}{Hugging Face}\hspace{6pt}%
\linkbtn{https://zibindong.github.io/actioncodec2-page/}{\color{accent}\faGlobe}{Project Page}\hspace{6pt}%
\linkbtn{https://github.com/ZibinDong/actioncodec2}{\color{accent}\faGithub}{GitHub}\par
\end{titlepanel}

\section{Introduction}
\label{sec:intro}

\begin{figure}[t]
\centering
\scalebox{1.0}{\includegraphics[width=\linewidth]{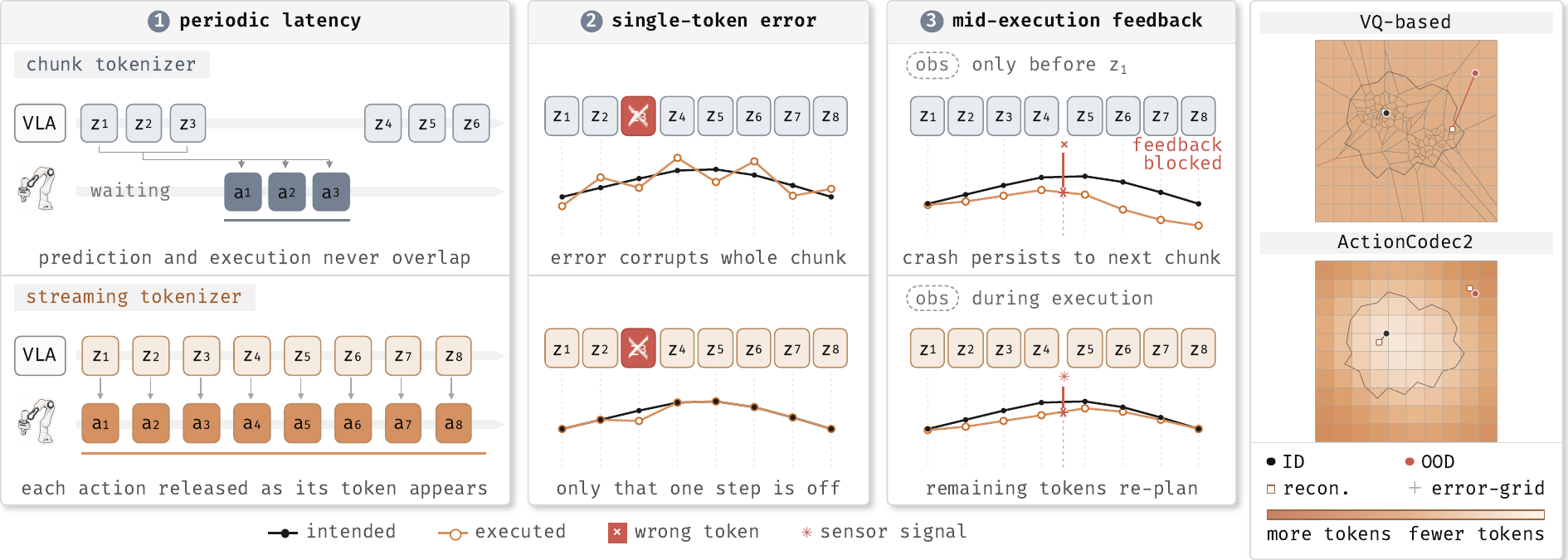}}
\caption{\textbf{Chunkwise and streaming tokenization.} Left: a chunk tokenizer releases actions only after the whole chunk is decoded, and an erroneous token can affect multiple actions within the chunk; a streaming tokenizer releases each action with its own tokens, so the policy can use feedback at every step. Right: a fixed VQ codebook allocates its codes according to the training distribution and provides no distribution-independent reconstruction bound; ActionCodec2 covers the reachable range and uses more tokens where needed.}
\label{fig:teaser}
\vspace{-5pt}
\end{figure}

In an autoregressive (AR) VLA, the action tokenizer encodes continuous actions into discrete tokens for training and decodes predicted tokens into executable actions \citep{rt2, openvla, fast, dong2026actioncodec}. Most existing action tokenizers encode one to two seconds of motion as a chunk (Figure~\ref{fig:teaser}, left), following a formulation widely used in diffusion policies \citep{chi2023diffusionpolicy, zhao2023aloha, bharadhwaj2024roboagent}. For example, FAST \citep{fast} exploits temporal correlations by compressing DCT coefficients with BPE, while VQ-VAE-based tokenizers \citep{mete2024quest, vq-vla, oat, x-tokenizer, dong2026actioncodec} learn compact latent representations of action chunks. With chunkwise tokenization, the robot can react to new observations only at chunk boundaries, and an erroneous token may affect multiple actions within a chunk. Real-time chunking \citep{black2025real} allows earlier reactions by replanning before a chunk ends, but reacting at every control step would require repeatedly predicting overlapping chunks. A streaming tokenizer instead releases actions as their symbols become available, allowing the policy to incorporate new observations at each control step without regenerating entire chunks.

\begin{figure}[tb]
\centering
\scalebox{0.7}{%
\begin{minipage}[t]{0.50\linewidth}
  \centering
  \includegraphics[width=\linewidth,trim={0 202.5 0 0},clip]{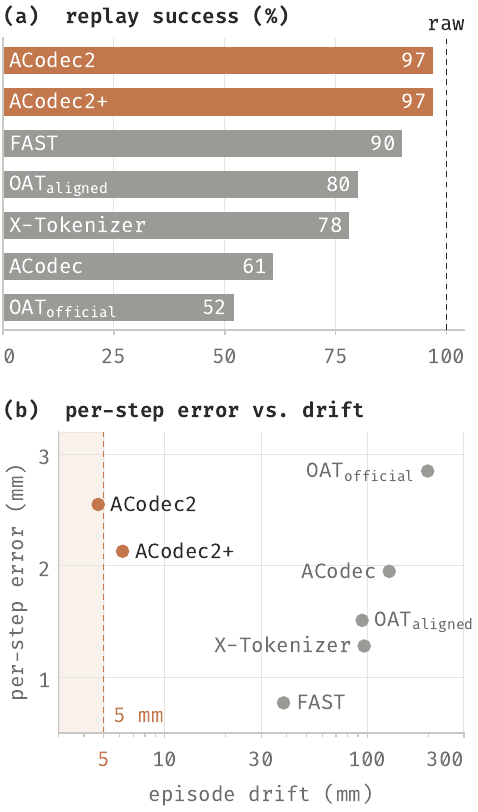}
\end{minipage}\hfill
\begin{minipage}[t]{0.46\linewidth}
  \centering
  \includegraphics[width=\linewidth,trim={0 0 0 187},clip]{fig/replay_audit.pdf}
\end{minipage}%
}
\caption{\textbf{Round-trip fidelity of action tokenizers on LIBERO demonstrations} (first-order variants). (a)~Demonstrations are tokenized, decoded, and replayed open loop; raw demonstrations score $100\%$. (b)~Per-step error versus mean episode drift.}
\label{fig:replay-audit}
\vspace{-5pt}
\end{figure}

The simplest streaming tokenizer is per-dimension binning, as in OpenVLA \citep{openvla}, which assigns one token to each action dimension. For 7-dimensional actions at $15\,\mathrm{Hz}$, this requires 105 tokens per second, substantially increasing autoregressive decoding latency. An alternative is to shorten the chunks of learned VQ-based tokenizers to one or two steps \citep{mete2024quest, vq-vla, liu2026faster, oat}. However, their fixed codebooks concentrate reconstruction accuracy on the training distribution, with no guarantee for unfamiliar motions (Figure~\ref{fig:teaser}, right). More importantly, accurate reconstruction of individual actions does not guarantee accurate trajectories, as small errors in incremental actions accumulate over time. In a replay test on LIBERO (Figure~\ref{fig:replay-audit}), FAST reconstructs individual actions to within $1\,\mathrm{mm}$ but accumulates about $40\,\mathrm{mm}$ of drift per episode, achieving only $90\%$ replay success. The other chunkwise baselines reach at most $80\%$, whereas ActionCodec2 keeps the drift below $5\,\mathrm{mm}$ and achieves $97\%$ replay success.

ActionCodec2 addresses trajectory drift by accounting for the quantization error left by previous steps. The encoder maintains the same reconstructed state as the decoder and uses it to select the next motion primitive, keeping the reconstructed trajectory within a prescribed tolerance of the original. Rather than allowing small errors to accumulate, each step corrects the error remaining from earlier steps. This reduces the range that needs to be quantized to the possible one-step motions and the residual tracking error, rather than the robot's entire workspace. We derive the minimum number of discrete motion symbols required to guarantee this error bound.

The motion primitives guarantee bounded tracking error, but representing every possible combination of primitives with a separate VLA token would require an enormous vocabulary. Representing each coordinate separately avoids this problem but produces long token sequences. We therefore compress these symbols without changing the reconstructed actions. Standard BPE \citep{bpe, fast} merges only adjacent symbols under a fixed coordinate ordering, limiting its ability to capture correlations across dimensions. We introduce Set-BPE, which losslessly merges co-occurring symbols across coordinates and neighboring steps, regardless of coordinate order. At $15\,\mathrm{Hz}$, ActionCodec2 requires only $8.8$ tokens per second for end-effector control and $10.8$ for joint control.

ActionCodec2 is fitted without gradient-based optimization: its primitive sets are determined by the motion bounds and prescribed tolerances, while Set-BPE learns its merges by counting symbol co-occurrences. In addition to benchmark-specific tokenizers fitted on each dataset's training corpus, we fit universal variants on about $20$k hours of data from $24$ embodiments for plug-and-play use across robots and tasks. Our contributions are:
\begin{itemize}[topsep=0pt, itemsep=2pt, leftmargin=1.1em, labelsep=0.5em, labelwidth=0.6em, align=left]
    \item We formulate bounded-error streaming action tokenization as a covering problem and prove the optimality of fixed motion primitives among one-symbol-per-step decoders, with an exact solution for translation and constant-factor-tight bounds for rotation.
    \item We introduce Set-BPE, which losslessly merges co-occurring primitive symbols across coordinates and neighboring steps, without relying on a fixed coordinate ordering. This allows merged tokens to capture coordinated motion patterns that standard BPE may miss, while reducing token count by $6.6$--$10.7\%$ on held-out data.
    \item Using the same G0.5 VLA \citep{g05}, ActionCodec2 achieves $98.2\%$ on LIBERO, $93.8\%$ on SimplerEnv, and $79.5\%$ on LIBERO-Plus, outperforming FAST, OAT, X-Tokenizer, and ActionCodec. It also achieves $73.9\%$ task progress across four zero-shot Franka tasks.
    \item We show that ActionCodec2 tokens can serve as a direct control interface for VLM agents, which store successful token sequences as reusable macros that carry over across attempts and to new tasks.
\end{itemize}

\vspace{-5pt}
\section{Preliminaries}
\label{sec:prelim}

Let $q_t$ be the configuration that the policy commands at control step $t$, such as an end-effector position, an end-effector orientation, or a vector of joint angles. Consecutive configurations differ by a relative motion $\delta_t\in G$, so that $q_{t+1}=q_t\delta_t$. The group operation is addition for position $G=(\R^3,+)$ and joint angles $G=(\R^n,+)$, and matrix multiplication for orientation $G=\SO(3)$. The feasible motions of one step form a set $\mathcal A\subseteq G$, for example the set permitted by the actuator limits. A tokenizer encodes $q_{0:T}$ into symbols from an alphabet of size $K$, and the decoder reconstructs a trajectory $\hat q_{0:T}$ with $\hat q_0=q_0$. Here and throughout the theoretical analysis, a \emph{symbol} refers to a primitive-level action symbol; Set-BPE later packs these symbols into the tokens predicted by the VLA. Let $\nu(t)$ be the number of symbols consumed when the decoder outputs step $t$. We call the symbol representation \emph{streaming} if $\nu(t)>\nu(t-1)$ for every $t\ge1$, so that every step has symbols of its own, and \emph{chunkwise} with length $\tau$ if $\nu(t)$ is constant within each block of $\tau$ consecutive steps. We define the \emph{tracking error} as $\xi_t=\hat q_t^{-1}q_t$, which reduces to $q_t-\hat q_t$ for position and joint angles. Let $\mathcal E\subseteq G$ be a tolerance set containing the identity, for example $[-5\mathrm{mm},5\mathrm{mm}]^3$ for position. A tokenizer has \emph{bounded error} if $\xi_t\in\mathcal E$ at every step of every feasible trajectory. We ask for the smallest alphabet that can guarantee bounded error when the decoder consumes one symbol per step.

\vspace{-5pt}
\section{ActionCodec2}
\label{sec:actioncodec2}

\begin{figure}[t]
\centering
 \scalebox{0.95}{\includegraphics[width=\linewidth]{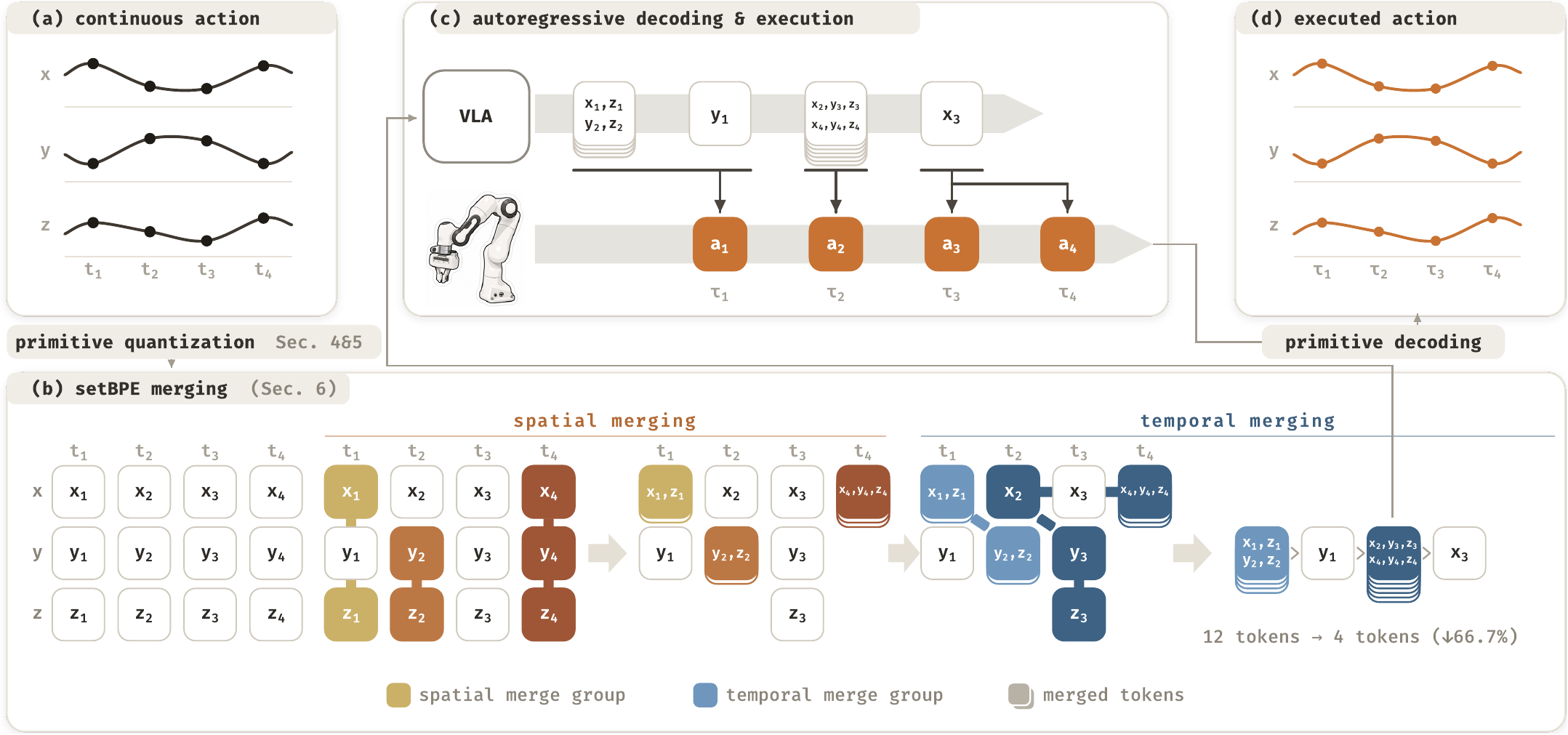}}
\caption{\textbf{ActionCodec2.} (a)~Each step of a continuous action is quantized into primitive atoms, one per coordinate. (b)~Set-BPE merges atoms that co-occur within a step and then across nearby steps, independent of coordinate order. (c)~The VLA emits the tokens autoregressively and releases each action once its step is complete. (d)~The decoded primitives track the trajectory within $\epsilon$.}
\label{fig:setbpe}
\vspace{-5pt}
\end{figure}

ActionCodec2 consists of two components (Figure~\ref{fig:setbpe}). The first represents each control step by a primitive, a fixed relative motion that the decoder composes with its current reconstruction. The encoder mirrors this deterministic update, computes the current tracking error, and selects a primitive that returns the next error to $\mathcal E$. Sections~\ref{sec:general} and~\ref{sec:euclidean} derive the minimum primitive alphabet required by this construction; the primitive sets we use factorize across coordinates, so a primitive is written as one atom per coordinate. The second component, Set-BPE (Section~\ref{sec:setbpe}), losslessly merges co-occurring atoms within and across steps into the token sequence predicted by the VLA.

\subsection{Minimum alphabet for bounded error}
\label{sec:general}

We begin with one-dimensional translation, where the main idea is easy to see. Let one step move by at most $b$, so $\mathcal A=[-b,b]$, let the tolerance be $\mathcal E=[-\epsilon,\epsilon]$, and let symbol $k$ apply a fixed motion $u_k$, so that $\hat q_{t+1}=\hat q_t+u_{z_t}$. The error evolves as $\xi_{t+1}=\xi_t+\delta_t-u_{z_t}$. If $\xi_t\in\mathcal E$, then before correction the error can lie anywhere in $\mathcal E+\mathcal A=[-(\epsilon+b),\ \epsilon+b]$. A primitive $u$ returns the error to $\mathcal E$ exactly when the pre-correction error lies in $u+\mathcal E=[u-\epsilon,\ u+\epsilon]$. Bounded error therefore requires the intervals $u_k+\mathcal E$ to cover $\mathcal E+\mathcal A$. The smallest such cover uses $\lceil1+b/\epsilon\rceil$ primitives; for example, a uniform grid over $[-b,b]$ achieves this with adjacent primitives separated by at most $2\epsilon$. Thus the required alphabet depends on the one-step motion range relative to the tolerance, not on the workspace size.

The error must be fed back for this to hold. Take $\epsilon=5\,\mathrm{mm}$, primitives spaced $2\epsilon=10\,\mathrm{mm}$ apart, and three consecutive motions of $3.4\,\mathrm{mm}$. If each motion is quantized on its own, the nearest primitive is $0$ every time, the decoder does not move, and the error grows as $3.4$, $6.8$, and $10.2\,\mathrm{mm}$, leaving $\mathcal E$ at the second step. If the encoder instead quantizes the pre-correction error $\xi_t+\delta_t$, it is $3.4$ at $t=0$ (primitive $0$, $\xi_1=3.4$), $6.8$ at $t=1$ (primitive $10$, $\xi_2=-3.2$), and $0.2$ at $t=2$ (primitive $0$, $\xi_3=0.2$). Independent quantization violates the tolerance at the second step, whereas feedback quantization stays within $5\,\mathrm{mm}$ throughout.

We next extend this argument to arbitrary configuration groups and arbitrary decoders that read one symbol per step. Consider a decoder that updates its reconstruction using one of $K$ maps,
\begin{equation}
\hat q_{t+1}=y_{z_t}(\hat q_t),
\qquad z_t\in\{1,\ldots,K\},
\qquad\hat q_0=q_0 .
\label{eq:general-step}
\end{equation}

One-step scalar binning~\citep{openvla} and vector-quantized tokenizers with a learned codebook~\citep{vqvae} can be written in this form. The one-dimensional requirement extends directly to the general decoder. Define the accepted pairs
$\mathcal S=\{(q,\hat q):\hat q^{-1}q\in\mathcal E\}$ and the feasible successors $\mathcal F(q)=q\mathcal A$. We require that, from every accepted pair and every feasible next configuration, at least one symbol keeps the next pair accepted:
\begin{equation}
\forall(q,\hat q)\in\mathcal S,\quad
\forall q^+\in\mathcal F(q),\quad
\exists k\le K:\ \bigl(q^+,y_k(\hat q)\bigr)\in\mathcal S .
\label{eq:full-tolerance}
\end{equation}

We call this the \emph{recoverability condition}, and let $K^\star$ be the smallest $K$ for which a decoder of the form \eqref{eq:general-step} satisfies it.

\proofin{proof:lem:onestep}
\begin{lemma}[One step suffices]
\label{lem:onestep}
If the recoverability condition holds, then from every exact start
$\hat q_0=q_0$ and along every feasible trajectory, one can select a
symbol at each step such that $\xi_t\in\mathcal E$ for all $t$.
The symbol at step $t$ can be selected using only
$(q_t,\hat q_t,q_{t+1})$.
\end{lemma}

We now consider a simpler decoder class in which each symbol applies a fixed relative motion. Given primitives $\mathcal U=\{u_1,\ldots,u_K\}\subseteq G$, define
$y_k(\hat q)=\hat q u_k$.
We call this a \emph{fixed-primitive decoder}. The one-dimensional construction above is a special case. On a general group, its error evolves as
$\xi_{t+1}=u_{z_t}^{-1}\xi_t\delta_t$.
Before correction, the error can therefore lie anywhere in
$\mathcal E\mathcal A:=\{\xi\delta:\xi\in\mathcal E,\ \delta\in\mathcal A\}$.
A primitive $u$ returns the error to $\mathcal E$ exactly when the pre-correction error lies in
$u\mathcal E:=\{u\eta:\eta\in\mathcal E\}$.
Hence a fixed-primitive decoder satisfies the recoverability condition exactly when
$\mathcal E\mathcal A\subseteq\bigcup_{u\in\mathcal U}u\mathcal E$.
The following theorem shows that restricting the decoder to fixed primitives does not increase the minimum alphabet size.

\proofin{proof:thm:composition-optimal}
\begin{theorem}[Optimality of fixed primitives]
\label{thm:composition-optimal}
Under the recoverability condition:
\begin{enumerate}
\item
A fixed-primitive decoder with primitive set
$\mathcal U\subseteq G$
satisfies the condition if and only if
\[
\mathcal E\mathcal A
\subseteq
\bigcup_{u\in\mathcal U}u\mathcal E.
\]

\item
Every decoder of the form \eqref{eq:general-step} that satisfies the
condition uses at least
\[
\min\left\{
|\mathcal U| :
\mathcal U\subseteq G \text{ finite},
\quad
\mathcal E\mathcal A
\subseteq
\bigcup_{u\in\mathcal U}u\mathcal E
\right\}
\]
symbols, where the minimum over an empty family is $+\infty$.

\item
Consequently, $K^\star$ equals this minimum, and a fixed-primitive decoder
attains it whenever the minimum is finite.
\end{enumerate}
\end{theorem}

The lower bound places no restriction on the encoder and therefore also applies when the entire trajectory is available in advance. The matching fixed-primitive construction, however, needs only the current state and the next configuration, so access to future actions cannot reduce $K^\star$ under the recoverability condition.

\subsection{Translation and rotation}
\label{sec:euclidean}

\begin{figure}[t]
\centering
\scalebox{1.0}{\includegraphics[width=0.9\linewidth]{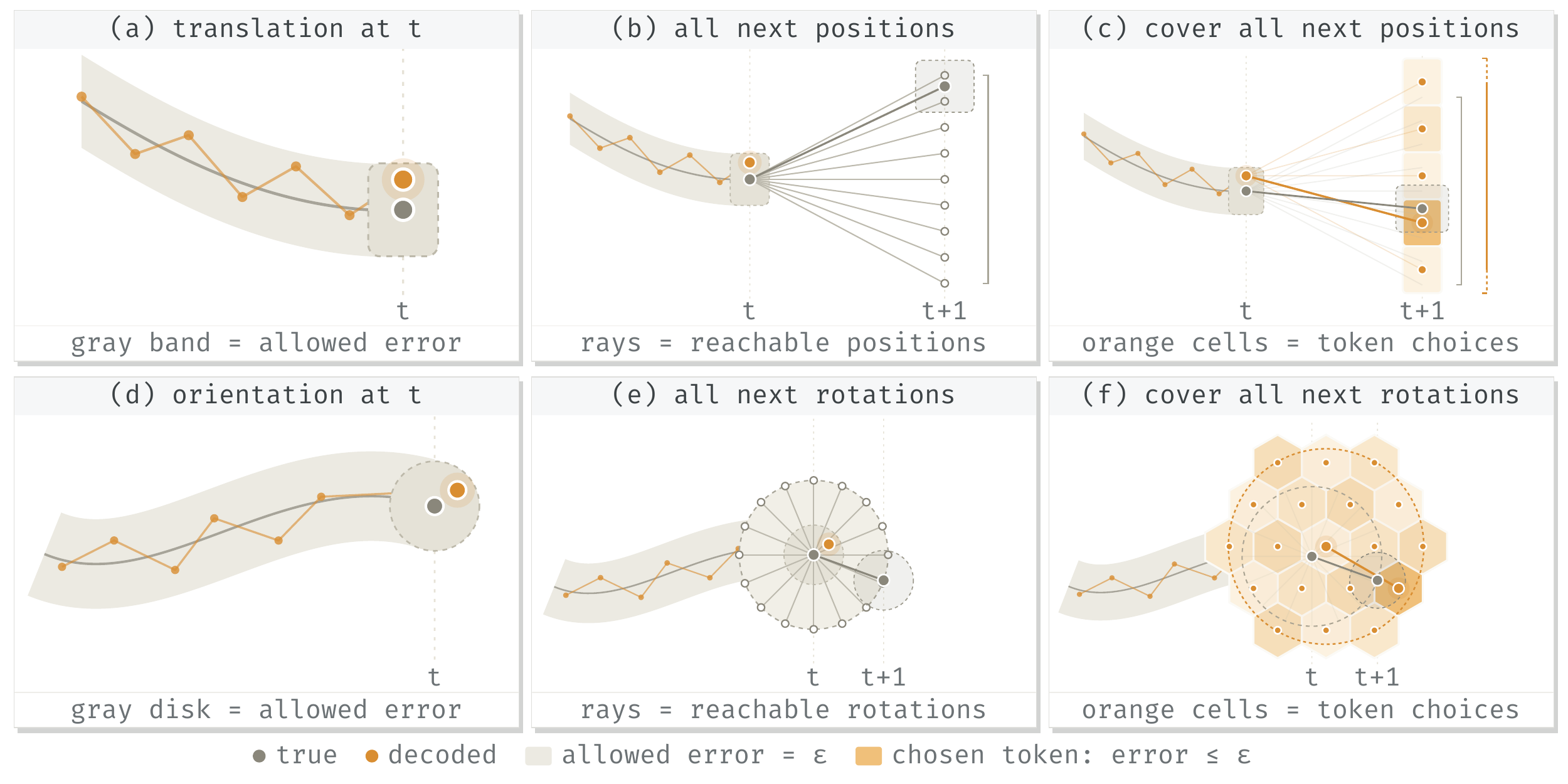}}
\caption{\textbf{Covering primitives for translation and orientation.} The decoded state initially lies within the prescribed tolerance of the true state. Candidate primitives cover the one-step reachable error set, and the selected primitive returns the tracking error to within $\epsilon$.}
\label{fig:primitives}
\end{figure}

For translation, $\mathcal E$ and $\mathcal A$ are boxes, and the covering number has a closed form (Figure~\ref{fig:primitives}, top).

\proofin{proof:thm:translation-exact}
\begin{corollary}[Translation]
\label{thm:translation-exact}
Let $G=(\R^d,+)$ with box motion set $\mathcal A=\prod_{j=1}^{d}[\underline a_j,\overline a_j]$ and tolerance $\mathcal E=\prod_{j=1}^{d}[-\epsilon_j,\epsilon_j]$, writing $\Delta a_j:=\overline a_j-\underline a_j$ with $\epsilon_j>0$.  In the additive group $(\R^d,+)$, the covering condition becomes $\mathcal E+\mathcal A\subseteq\bigcup_{u\in\mathcal U}(u+\mathcal E)$, and the minimum alphabet size is
\begin{equation}
K^\star_{\rm trans}=\prod_{j=1}^{d}N_j,
\qquad
N_j=\left\lceil1+\frac{\Delta a_j}{2\epsilon_j}\right\rceil.
\label{eq:translation-nj}
\end{equation}
This minimum is attained by the axiswise uniform grid $\mathcal U=\prod_j\mathcal U_j$ in Algorithm~\ref{alg:primitives}, where $\mathcal U_j=\{\underline a_j\}$ if $\Delta a_j=0$ and otherwise
\begin{equation}
\mathcal U_j
=
\left\{
\underline a_j+\frac{\Delta a_j(i-1)}{N_j-1}
:i=1,\ldots,N_j
\right\}.
\label{eq:translation-grid}
\end{equation}
If feasible trajectories are restricted to a workspace, clipping each reconstructed successor to the workspace preserves the error guarantee. The same alphabet therefore remains sufficient, although exploiting the workspace constraint may reduce the minimum required size.
\end{corollary}

As a concrete example, an arm moving at up to $1.2\,\mathrm{m/s}$ under $15\,\mathrm{Hz}$ control can move by at most $80\,\mathrm{mm}$ along one axis per step, giving $\Delta a_j=160\,\mathrm{mm}$. With $\epsilon_j=5\,\mathrm{mm}$, Equation~\eqref{eq:translation-nj} gives $N_j=\lceil1+160/10\rceil=17$ primitives, spaced $10\,\mathrm{mm}$ apart. By comparison, directly binning absolute positions over a $4\,\mathrm{m}$ workspace at the same tolerance requires $400$ bins.

For orientation, the tracking error is itself a rotation, measured by its intrinsic rotation angle. The tolerance and motion sets are therefore geodesic balls in $\SO(3)$, whose volumes provide bounds on the covering number (Figure~\ref{fig:primitives}, bottom). We write $\mathrm{Rot}(v)$ for the rotation by angle $\lVert v\rVert$ about the axis $v/\lVert v\rVert$, with $\mathrm{Rot}(0)=I$.

\proofin{proof:prop:so3-exact}
\begin{corollary}[Orientation]
\label{prop:so3-exact}
Let $G=\SO(3)$ with $R_{t+1}=R_t\Delta_t$ and $\hat R_{t+1}=\hat R_tU$, and let $\theta(Q)=\arccos\bigl((\operatorname{tr}Q-1)/2\bigr)\in[0,\pi]$ be the rotation angle. Let $\mathcal B_R(r)=\{Q:\theta(Q)\le r\}$, $\mathcal E=\mathcal B_R(\epsilon_R)$, and $\mathcal A=\mathcal B_R(b_R)$, with $0<\epsilon_R\le\pi$ and $0\le b_R\le\pi$. Then $\mathcal E\mathcal A=\mathcal B_R(\rho_R)$ with $\rho_R=\min\{\pi,\epsilon_R+b_R\}$. Define $\beta_R=b_R/\epsilon_R$ and let $V_R(r)=(r-\sin r)/\pi$ denote the normalized Haar volume of $\mathcal B_R(r)$. Then
\begin{equation}
\left\lceil\frac{V_R(\rho_R)}{V_R(\epsilon_R)}\right\rceil
\le K^\star_{\rm rot}
\le
\left\lfloor\frac{V_R(\min\{\pi,\rho_R+\epsilon_R/2\})}{V_R(\epsilon_R/2)}\right\rfloor,
\label{eq:so3-bounds}
\end{equation}
so $K^\star_{\rm rot}=\Theta\bigl((1+\beta_R)^3\bigr)$ whenever $\rho_R+\epsilon_R/2\le r_0<\pi$, with constants depending on $r_0$ only. The lattice $\{\mathrm{Rot}(v):v\in L^3\}$, where $L$ is a uniform grid on $[-\rho_R,\rho_R]$ with spacing at most $2\epsilon_R/\sqrt3$, satisfies the covering condition.
\end{corollary}

With $n_R$ grid points per axis, the lattice contains $n_R^3\approx(\sqrt3\,\rho_R/\epsilon_R)^3$ points. This is within a constant factor of the optimal covering rate, but includes additional points because the grid covers the cube $[-\rho_R,\rho_R]^3$ rather than only the ball of admissible rotation vectors. We nevertheless use the lattice because it factorizes the rotation vector into three coordinate indices, each taking $n_R$ values, exactly as in the translation case. A direct cover of the rotation ball can use fewer primitives, but each primitive must then be represented by a separate joint rotation index. For $\epsilon_R=0.02\,\mathrm{rad}$ and $\rho_R=1\,\mathrm{rad}$, this requires more than $10^5$ such indices, far beyond a vocabulary of a few thousand tokens. The three indices parameterize the coordinates of a single rotation vector; they do not represent three sequential rotations about the coordinate axes. We choose $n_R$ to be odd so that $0\in L$, giving an exact identity-rotation primitive.

In our implementation, rotation primitives act on the left, $\hat R_{t+1}=U\hat R_t$, corresponding to world-frame increments. The minimum primitive count is unchanged:

\proofin{proof:prop:so3-frames}
\begin{proposition}[Frame equivalence]
\label{prop:so3-frames}
Let increments act on the left, $R_{t+1}=\Delta^w_tR_t$ and $\hat R_{t+1}=U\hat R_t$, so that the error is $\Xi^w_t=R_t\hat R_t^\top$ and the covering condition is $\mathcal A^w_R\mathcal B_R(\epsilon_R)\subseteq\bigcup_k\mathcal B_R(\epsilon_R)U_k$. Its minimum equals the body-frame value $K^\star_{\rm rot}$ evaluated on the inverted motion set $(\mathcal A^w_R)^\top$. Hence the world- and body-frame minima are identical whenever $\mathcal A^w_R$ is closed under inversion, as is the case for a geodesic ball.
\end{proposition}

Algorithm~\ref{alg:primitives} constructs both primitive sets. At each step, the encoder computes the pre-correction error $\hat q_t^{-1}q_{t+1}$ and selects the nearest grid value independently along each coordinate; for rotation, this selection is performed in the corresponding rotation-vector coordinates.

\begin{algorithm}[t]
\caption{Primitive lists for $(\R^d,+)$ and $\SO(3)$: a uniform grid on each translation axis and a lattice of rotation vectors, with the nearest-primitive encoder}
\label{alg:primitives}
\small
\scalebox{1.0}{\parbox{\linewidth}{%
\begin{algorithmic}[1]
\Procedure{Translation}{$\underline a,\overline a,\epsilon$}
  \Require $\underline a_j\le\overline a_j$, $\epsilon_j>0$ for all $j$
  \For{$j=1,\ldots,d$}
    \State $\Delta a_j\gets\overline a_j-\underline a_j$ \Comment{one-step motion range}
    \State $N_j\gets\lceil1+\Delta a_j/(2\epsilon_j)\rceil$ \Comment{minimum number of bins}
    \State $\mathcal U_j\gets\{\underline a_j\}$
    \If{$N_j>1$}
      \State $\mathcal U_j\gets\bigl\{\underline a_j+\tfrac{\Delta a_j(i-1)}{N_j-1}\bigr\}_{i=1}^{N_j}$ \Comment{uniform binning}
    \EndIf
  \EndFor
  \State \Return $\mathcal U_1\times\cdots\times\mathcal U_d$
\EndProcedure
\Procedure{Rotation}{$\epsilon_R,b_R$}
  \Require $0<\epsilon_R\le\pi$, $0\le b_R\le\pi$
  \State $\rho_R\gets\min\{\pi,\epsilon_R+b_R\}$ \Comment{maximum error before correction}
  \State $n_R\gets\lceil\sqrt3\,\rho_R/\epsilon_R\rceil+1$, increased by one if even \Comment{odd, so that $0\in L$}
  \State $L\gets\bigl\{-\rho_R+\tfrac{2\rho_R(i-1)}{n_R-1}\bigr\}_{i=1}^{n_R}$ \Comment{spacing at most $2\epsilon_R/\sqrt3$}
  \State \Return $\{\mathrm{Rot}(v):v\in L^3\}$ \Comment{$n_R^3$ primitives, one index per coordinate}
\EndProcedure
\Procedure{Encode}{$\hat q_t,q_{t+1}$}
  \State $w\gets\hat q_t^{-1}q_{t+1}$ \Comment{error before correction}
  \If{translation}
    \State \Return $\bigl(\argmin_{u\in\mathcal U_j}|w_j-u|\bigr)_{j=1}^{d}$
  \Else
    \State $v\gets\mathrm{Rot}^{-1}(w)$ \Comment{rotation vector, $\lVert v\rVert\le\pi$}
    \State \Return $\bigl(\argmin_{l\in L}|v_j-l|\bigr)_{j=1}^{3}$
  \EndIf
\EndProcedure
\end{algorithmic}}}
\end{algorithm}

\subsection{Set-BPE: order-free lossless packing}
\label{sec:setbpe}

The preceding construction reduces the number of discrete motion primitives needed to guarantee bounded tracking error, but a smaller primitive alphabet does not necessarily mean fewer tokens for an AR model. We therefore turn to lossless compression, which preserves the original motion primitives and their tracking guarantee. Standard BPE \citep{bpe} provides one such approach, but it merges only symbols that are adjacent under a fixed ordering of action coordinates. For example, translation along $z$ may co-occur with rotation about $z$, yet no fixed ordering can make all correlated coordinates adjacent. This limitation becomes more pronounced across embodiments with different coordinate layouts. Set-BPE merges sets of coordinate assignments. The assignment at step $t$ is $S_t=\{(j,z_{t,j})\}_{j=1}^{D}$. Although $S_t$ can take $\prod_{j=1}^{D}N_j$ possible values, it is composed from only $\sum_j N_j$ distinct \emph{atoms}, each specifying one coordinate and one bin. Initially every atom is a token. A \emph{spatial} token is a set of atoms on distinct coordinates, and the spatial tokens of a step partition $S_t$. A \emph{spatial merge} replaces two co-occurring tokens on disjoint coordinate sets by their union; for example, $\{(z,+1)\}$ and $\{(r_z,+1)\}$ merge into $\{(z,+1),(r_z,+1)\}$. A \emph{temporal} token contains spatial tokens at offsets $0,\ldots,\tau-1$, with exactly one spatial token at offset $0$, designated as its \emph{root}. A \emph{temporal merge} combines two token instances $\delta<\tau$ steps apart into one token with span at most $\tau$ and the root of the earlier instance; during a steady motion, for example, it replaces a spatial token repeated on two consecutive steps ($\delta=1$) by one token.

Token instances are serialized first by start step and, among tokens with the same start step, by the smallest coordinate of the root. This ordering makes step boundaries recoverable from the token stream alone: the next root is assigned to the lowest unassigned coordinate of the earliest incomplete step, and a step is released once all of its coordinates have been assigned. No explicit timestamps are therefore required. Moreover, every valid prefix admits a continuation because a primitive atom is always available for each unassigned coordinate. Since every merge is invertible, the resulting token stream $Z$ is in bijection with the original atom representation $Y$, so $H(Y)=H(Z)$ and both the decoded trajectory and its error guarantee are unchanged (Appendix~\ref{sec:lossless}).

Set-BPE is fitted greedily by repeatedly committing the merge with the largest exact gain, measured by the number of instance pairs it replaces. To avoid rescanning the full corpus at every round, identical steps are aggregated into weighted records and counters are maintained only for candidates that actually occur. Candidate scores that become stale remain valid upper bounds, allowing a lazy priority queue to recover the exact maximum without recomputing every score. As a result, each round requires work proportional to the instances affected by the committed merge rather than to the full corpus or the square of the vocabulary size (Algorithm~\ref{alg:setbpe}; Appendix~\ref{app:bpe-training} gives the exact counting rules).

\begin{algorithm}[t]
\caption{Set-BPE: exact greedy fitting and streaming coding. $\Exp$ expands a spatial token to its atoms and $\Exp_T$ a temporal token to its (offset, spatial token) pairs; $a$ is the start step of an instance in \textsc{Encode} and the earliest incomplete step in \textsc{Decode}, and $C_0$ is the root.}
\label{alg:setbpe}
\small
\begin{algorithmic}[1]
\Procedure{Fit}{weighted steps $\{(r,w)\}$, span $\tau$, budget $V$}
  \State counters $\gets$ exact scores of observed candidates \Comment{one vectorized pass; spatial first, then temporal}
  \State heap $\gets$ observed candidates keyed by score $U$
  \While{the vocabulary is below budget}
    \State $(c,U)\gets\textsc{PopMax}(\text{heap})$
    \State $f\gets$ exact score of $c$ \Comment{occurrence-record intersection; greedy matching for a self-rule}
    \If{$f<U$}
      \State push $(c,f)$; \textbf{continue} \Comment{stale bound: requeue}
    \EndIf
    \State commit $c$: replace the instances it fuses, update the incident counters
  \EndWhile
\EndProcedure
\Procedure{Encode}{$S_{1:T}$}
  \State apply the spatial then the temporal merges in training order
  \State \Return the instances sorted by $\bigl(a,\min\supp C_0\bigr)$
\EndProcedure
\Procedure{Decode}{token stream}
  \State $a\gets1$; \ $\hat S_t\gets\emptyset$ for every $t$
  \ForAll{$Q$ in stream order}
    \While{$|\hat S_a|=D$}
      \State emit $q_a$; \ $a\gets a+1$
    \EndWhile
    \State $\hat S_{a+\delta}\gets\hat S_{a+\delta}\uplus\Exp(C)$ for $(\delta,C)\in\Exp_T(Q)$
  \EndFor
  \State emit the remaining completed steps in order
\EndProcedure
\end{algorithmic}
\end{algorithm}

\subsection{Fitting a Universal ActionCodec2}
\label{sec:fitting}

\begin{wrapfigure}{r}{0.38\linewidth}
\vspace{-40pt}
\centering
\includegraphics[width=\linewidth]{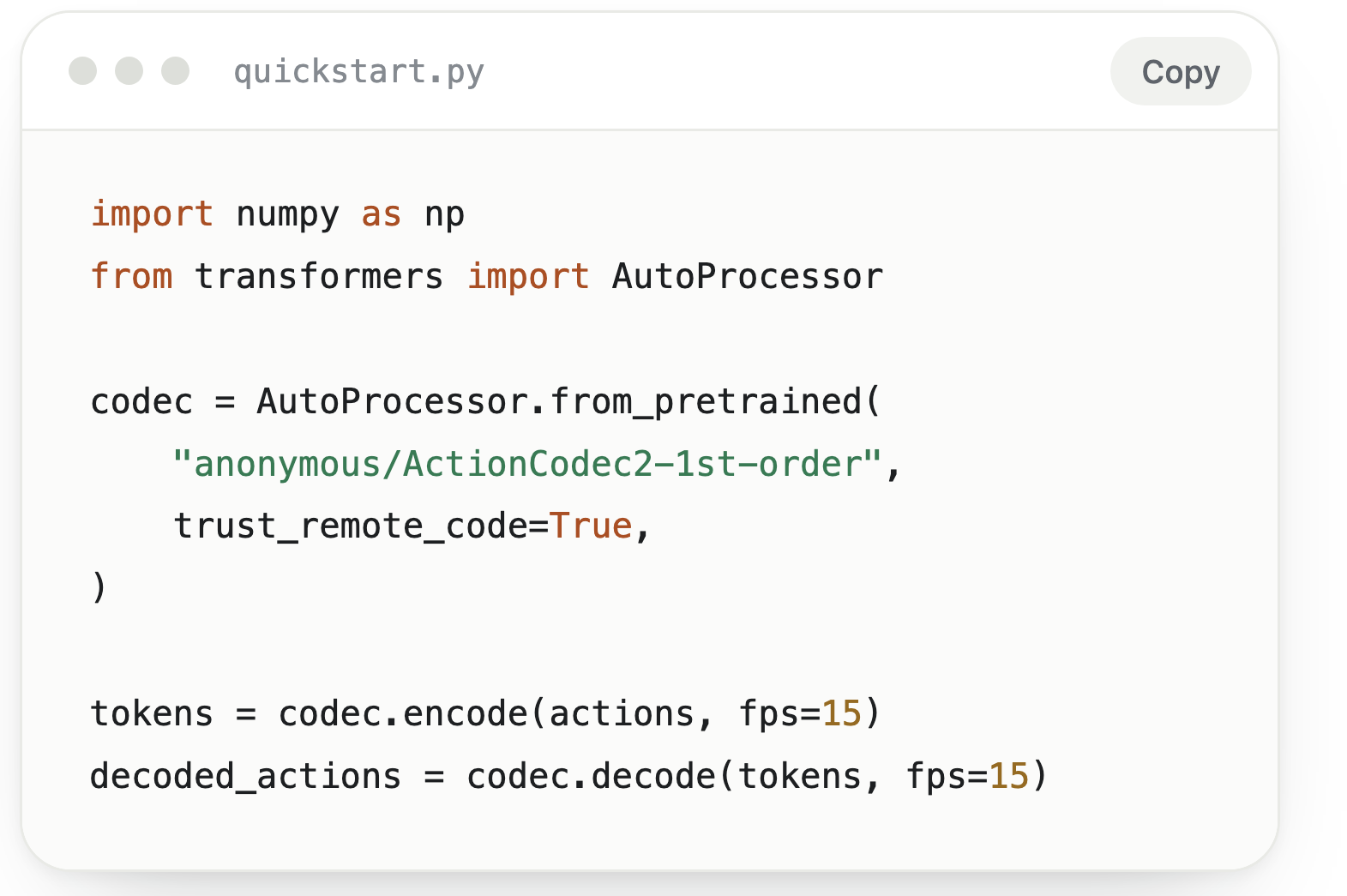}
\vspace{-10pt}
\caption{\textbf{ActionCodec2 as a Hugging~Face \texttt{AutoProcessor}.}}
\label{fig:quickstart}
\vspace{-20pt}
\end{wrapfigure}

To enable general-purpose use across robots and tasks, we fit universal ActionCodec2 tokenizers on about $20$k hours of robot data from $24$ embodiments, covering both household and industrial settings. The corpus contains $105$M end-effector actions from $146$k trajectories and $204$M joint actions from $233$k trajectories. We fit separate tokenizers for end-effector and joint control. The tolerances are set to $5\,\mathrm{mm}$ for position, $0.02\,\mathrm{rad}$ for orientation, and $4$--$20\,\mathrm{mrad}$ for joint angles, scaled according to each joint's influence on end-effector position, estimated from the URDF kinematics of the pretraining robots at their average configurations. This keeps the joint-space tolerances approximately consistent with the $5\,\mathrm{mm}$ end-effector position tolerance. We also include an exact zero on each primitive axis to represent stationary motion, adding at most one primitive per axis while preserving the covering condition.

Set-BPE is then fitted on the resulting primitive streams to obtain a vocabulary of $4{,}096$ tokens. At $15\,\mathrm{Hz}$, the first-order tokenizers require only $8.8$ tokens per second for end-effector control and $10.8$ for joint control on the pretraining dataset. The entire fitting process requires no gradient-based optimization and runs on a single CPU node. On held-out trajectories, tracking errors remain within the prescribed tolerances, and the token streams decode losslessly to the original primitive sequences. We also fit second-order variants that maintain a velocity state and encode changes in velocity, further reducing the number of primitive atoms from $176$ to $86$ for end-effector control and from $353$ to $273$ for joint control (Appendix~\ref{sec:second-general}). We denote the universal first- and second-order tokenizers by \textbf{ACodec2+} and \textbf{ACodec2+$^{\mathrm{II}}$}, respectively. For controlled comparisons, we additionally fit benchmark-specific versions, denoted by \textbf{ACodec2} and \textbf{ACodec2$^{\mathrm{II}}$}. The fitted tokenizers are available through the HuggingFace \texttt{AutoProcessor} interface for direct use in downstream applications (Figure~\ref{fig:quickstart}).\footnote{Additional universal variants support position tolerances of 1, 5, and 10 mm and vocabulary sizes of 1k, 2k, 4k, 8k, and 16k; see Appendix~\ref{app:configs} and \url{https://huggingface.co/ZibinDong/ActionCodec2}.}

\section{Experiments}
\label{sec:experiments}

For each tokenizer, we replace the original action tokenizer of the pretrained AR VLA G0.5 \citep{g05} and fine-tune the resulting model with the same recipe (Appendix~\ref{app:tokenizers}). The original G0.5 tokenizer is not included among the compared methods.

\subsection{Setup}
\label{sec:exp-setup}

\textbf{Tokenizers.}
We abbreviate ActionCodec2 and ActionCodec \citep{dong2026actioncodec} as \textbf{ACodec2} and \textbf{ACodec}, respectively; the ACodec2 variants follow the notation of Section~\ref{sec:fitting}. The baselines are \textbf{BEAST} \citep{beast}, which quantizes B-spline control points; \textbf{FAST} \citep{fast}, which applies a DCT followed by BPE to action chunks; \textbf{X-Tokenizer} \citep{x-tokenizer}, a VQ tokenizer pretrained on $2$B action frames; and \textbf{OAT} \citep{oat} and \textbf{ACodec}, VQ tokenizers with Perceiver-style architectures. OAT is evaluated both at its official token budget ($\text{OAT}_{\text{official}}$) and at one matched to the average token rate of ACodec2 ($\text{OAT}_{\text{aligned}}$).

\begin{table}[t]
\centering
\small
\setlength{\tabcolsep}{2.5pt}
\caption{\textbf{Qualitative comparison of the evaluated action tokenizers.} \tdot~=~attribute applies; \tdash~=~not applicable; \tgood~=~yes; \tno~=~no. \emph{Lossless} refers to the compression stage only: quantization always loses. \emph{Tracking bound}: a worst-case bound on the tracking error of the decoded trajectory at every step. \emph{Streaming}: actions are represented stepwise along physical time rather than as a jointly represented fixed-horizon chunk. \emph{Pretrained}: once on a large multi-embodiment corpus. \emph{Held-out universal}: zero-shot on unseen corpora or embodiments.}
\label{tab:tokenizer-properties}
\scalebox{0.85}{%
\begin{tabular}{lccccccc}
\toprule
\textbf{Tokenizer} & \textbf{Neural} & \textbf{Rule-based} & \textbf{Lossless} & \textbf{Track.\ bound} & \textbf{Streaming} & \textbf{Pretrained} & \textbf{Held-out univ.} \\
\midrule
FAST \citep{fast}                     & \tdash & \tdot & \tgood & \tno & \tno   & \tgood & \tgood \\
X-Tokenizer \citep{x-tokenizer}       & \tdot  & \tdash & \tno  & \tno   & \tno   & \tgood & \tno   \\
OAT \citep{oat}                       & \tdot  & \tdash & \tno  & \tno   & \tno   & \tno   & \tno   \\
ACodec \citep{dong2026actioncodec}    & \tdot  & \tdash & \tno  & \tno   & \tno   & \tno   & \tno   \\
\midrule
ACodec2 (ours, corpus-fit)            & \tdash & \tdot & \tgood & \tgood & \tgood & \tno   & \tno   \\
ACodec2+ (ours, universal)            & \tdash & \tdot & \tgood & \tgood & \tgood & \tgood & \tgood \\
\bottomrule
\end{tabular}%
}
\end{table}

\textbf{Benchmarks.}
We evaluate ActionCodec2 on three simulation benchmarks and one real-robot setting, covering standard manipulation tasks, robustness to perturbations, and generalization to unseen environments. LIBERO \citep{libero} contains four task suites (spatial, object, goal, and long-horizon), while LIBERO-Plus \citep{libero-plus} evaluates these tasks under seven categories of perturbation. SimplerEnv \citep{simplerenv} evaluates policies trained on BridgeData \citep{walke2023bridgedata} across four WidowX simulation tasks. For real-robot evaluation, we train policies on DROID \citep{droid} and test them on a physical Franka arm across four task suites: color-discriminated pick-and-place, snack storing and serving, nested-cup arrangement, and towel unfolding. Except for LIBERO, all evaluation scenes are unseen during training (Appendix~\ref{app:benchmarks}).

\subsection{Main results on simulation benchmarks}
\label{sec:exp-main}

Using the same G0.5 backbone, ACodec2+$^{\mathrm{II}}$ achieves the highest average performance on all three simulation benchmarks: $98.2\%$ on LIBERO, $79.5\%$ on LIBERO-Plus, and $93.8\%$ on SimplerEnv (Tables~\ref{tab:libero-simpler} and~\ref{tab:libero-plus}). The advantage is particularly large under robot-initial-state perturbations in LIBERO-Plus, where ACodec2$^{\mathrm{II}}$ and ACodec2+$^{\mathrm{II}}$ reach $81.9\%$ and $85.3\%$, compared with $52.3$--$65.7\%$ for the other tokenizers. The pretrained ACodec2+ also outperforms the benchmark-specific ACodec2 on both out-of-distribution benchmarks. Under distribution shift, ACodec2+ produces substantially more small-magnitude actions: on LIBERO-Plus, rotations within $1^{\circ}$ account for $42\%$ of its commands, compared with $16\%$ for ACodec2; on SimplerEnv, displacements within $1\,\mathrm{mm}$ account for $18\%$ versus $0\%$. This more conservative action distribution may reduce accumulated rollout error and partly explain the advantage of pretraining.

\begin{table}[t]
\centering
\small
\setlength{\tabcolsep}{4pt}
\caption{\textbf{LIBERO and SimplerEnv results (\%).} Every row uses the same G0.5 VLA backbone~\citep{g05}, fine-tuned with the corresponding tokenizer. LIBERO: evaluated on the four LIBERO suites with $50$ official initial states per task ($2{,}000$ episodes). SimplerEnv: four WidowX tasks, $120$ initial states each ($480$ episodes). \textbf{Bold} is the best in a column, \underline{underline} the second best.}
\label{tab:libero-simpler}
\scalebox{0.93}{%
\begin{tabular}{lccccc|ccccc}
\toprule
 & \multicolumn{5}{c|}{\textbf{LIBERO}} & \multicolumn{5}{c}{\textbf{SimplerEnv}} \\
\cmidrule(lr){2-6}\cmidrule(lr){7-11}
\textbf{Tokenizer} & \textbf{Spatial} & \textbf{Object} & \textbf{Goal} & \textbf{Long} & \textbf{Avg.} & \textbf{Spoon} & \textbf{Carrot} & \textbf{Stack} & \textbf{Eggplant} & \textbf{Avg.} \\
\midrule
FAST \citep{fast}                                & 95.4 & 99.2 & 85.4 & 91.2 & 92.8 & 91.7 & 64.2 & 88.3 & \textbf{100.0} & 86.0 \\
OAT$_{\text{official}}$ \citep{oat}              & 93.8 & 95.8 & 88.4 & 85.8 & 91.0 & 9.2  & 16.7 & 2.5  & 23.3 & 12.9 \\
OAT$_{\text{aligned}}$ \citep{oat}               & 98.2 & \underline{99.8} & 95.6 & 91.8 & 96.4 & 9.2  & 7.5  & 15.0 & 44.2 & 19.0 \\
ACodec \citep{dong2026actioncodec}               & 95.8 & 99.6 & 94.8 & 90.2 & 95.1 & 78.3 & 55.0 & 47.5 & 76.7 & 64.4 \\
X-Tokenizer \citep{x-tokenizer}                  & 88.6 & 96.0 & 87.6 & 82.8 & 88.8 & 90.8 & 65.8 & 81.7 & 95.0 & 83.3 \\
BEAST \citep{beast}                              & 96.4 & 97.8 & 94.1 & 93.5 & 95.5 & 93.3 & 65.8 & \textbf{94.2} & 92.5 & 86.5 \\
\midrule
\textbf{ACodec2}                                 & \underline{98.6} & 99.2 & \underline{97.8} & 91.8 & 96.9 & 95.8 & 83.3 & 90.0 & 94.2 & 90.8 \\
\textbf{ACodec2+}                                & 97.2 & 99.6 & 95.4 & 94.4 & 96.7 & \underline{97.5} & 85.0 & 89.2 & \underline{95.8} & 91.9 \\
\textbf{ACodec2$^{\mathrm{II}}$}                 & 98.2 & \textbf{100.0} & 95.8 & \underline{95.2} & \underline{97.3} & 95.8 & \underline{95.0} & \textbf{94.2} & 87.5 & \underline{93.1} \\
\textbf{ACodec2+$^{\mathrm{II}}$}                & \textbf{98.8} & 99.4 & \textbf{98.2} & \textbf{96.4} & \textbf{98.2} & \textbf{98.3} & \textbf{96.7} & 89.2 & 90.8 & \textbf{93.8} \\
\bottomrule
\end{tabular}%
}
\end{table}

\begin{table}[t]
\centering
\footnotesize
\setlength{\tabcolsep}{3.5pt}
\caption{\textbf{LIBERO-Plus robustness (\%).} Task-level success over all $10{,}030$ perturbed tasks; each column pools the perturbations in that category. \textbf{Bold} is the best in a column, \underline{underline} the second best.}
\label{tab:libero-plus}
\scalebox{1.0}{%
\begin{tabular}{lcccccccc}
\toprule
\textbf{Tokenizer} & \textbf{Camera} & \textbf{Robot} & \textbf{Lang.} & \textbf{Light} & \textbf{Backgr.} & \textbf{Noise} & \textbf{Layout} & \textbf{Avg.} \\
\midrule
FAST \citep{fast}                       & 55.2 & 64.5 & 73.8 & 86.9 & 87.5 & 51.4 & 80.1 & 69.7 \\
OAT$_{\text{official}}$ \citep{oat}                            & 48.0 & 53.8 & 73.7 & 84.6 & 83.7 & 44.1 & 74.8 & 64.3 \\
OAT$_{\text{aligned}}$ \citep{oat}                      & 62.5 & 61.5 & \textbf{78.7} & 94.5 & 89.9 & 53.8 & 83.5 & 73.2 \\
ACodec \citep{dong2026actioncodec}     & 59.7 & 57.8 & 75.3 & 90.8 & 87.0 & 48.5 & 79.0 & 69.4 \\
X-Tokenizer \citep{x-tokenizer}        & 42.7 & 52.3 & 56.9 & 67.8 & 65.3 & 34.5 & 59.8 & 52.9 \\
\midrule
\textbf{ACodec2}                        & 58.2 & 60.9 & \underline{78.5} & 92.7 & 90.4 & 52.3 & 82.1 & 71.8 \\
\textbf{ACodec2+}                       & 64.0 & 65.7 & 76.7 & \underline{94.9} & 92.7 & \underline{57.8} & 83.3 & 74.8 \\
\textbf{ACodec2$^{\mathrm{II}}$}                & \underline{65.4} & \underline{81.9} & 78.1 & \textbf{95.4} & \underline{93.5} & 56.1 & \underline{84.1} & \underline{77.7} \\
\textbf{ACodec2+$^{\mathrm{II}}$}               & \textbf{66.4} & \textbf{85.3} & 77.5 & 90.6 & \textbf{97.1} & \textbf{63.9} & \textbf{84.8} & \textbf{79.5} \\
\bottomrule
\end{tabular}
}
\end{table}

\begin{table}[t]
\centering
\small
\setlength{\tabcolsep}{3.5pt}
\caption{\textbf{Ablations (\%).} Upper: the client runs on a $20\,\mathrm{Hz}$ wall clock and sends zero motion whenever its action queue is empty; Lower: the same ACodec2 primitive stream packed by BPE or Set-BPE; both packers are lossless, and gray values give the gain of Set-BPE. \textbf{Bold} is the best within a panel, \underline{underline} the second best.}
\label{tab:ablation-detail}
\scalebox{1.0}{%
\begin{tabular}{lccccc}
\toprule
\textbf{Tokenizer / setting} & \textbf{Spatial} & \textbf{Object} & \textbf{Goal} & \textbf{Long} & \textbf{Avg.} \\
\midrule
\multicolumn{6}{l}{\emph{Deployment: real-time control at $20\,\mathrm{Hz}$}} \\
FAST \citep{fast} / chunk                        & 35.0 & 40.0 & 47.2 & 21.4 & 35.9 \\
X-Tokenizer \citep{x-tokenizer} / chunk          & 58.8 & 72.8 & 65.6 & 45.0 & 60.6 \\
ACodec \citep{dong2026actioncodec} / chunk       & 86.0 & 94.6 & 83.8 & 73.4 & 84.5 \\
OAT$_{\text{aligned}}$ \citep{oat} / chunk       & 92.0 & 94.0 & 86.4 & 76.4 & \underline{87.2} \\
\textbf{ACodec2+} / chunk                        & 82.2 & 83.0 & 72.8 & 54.4 & 73.1 \\
\textbf{ACodec2+} / streaming                   & \textbf{94.0} & \textbf{96.8} & \textbf{88.2} & \textbf{85.0} & \textbf{91.0} \\
\midrule
\multicolumn{6}{l}{\emph{Packing: Set-BPE vs.\ order-dependent BPE}} \\
ACodec2 / order-dependent BPE                    & 96.0 & 98.8 & 94.8 & 87.4 & 94.3 \\
ACodec2 / Set-BPE  & \textbf{98.6}\gdelta{+2.6} & \textbf{99.2}\gdelta{+0.4} & \textbf{97.8}\gdelta{+3.0} & \textbf{91.8}\gdelta{+4.4} & \textbf{96.9}\gdelta{+2.6} \\
\bottomrule
\end{tabular}
}
\end{table}

\subsection{Real-robot experiments}
\label{sec:exp-real}

\begin{figure}[t]
\centering
\includegraphics[width=\linewidth]{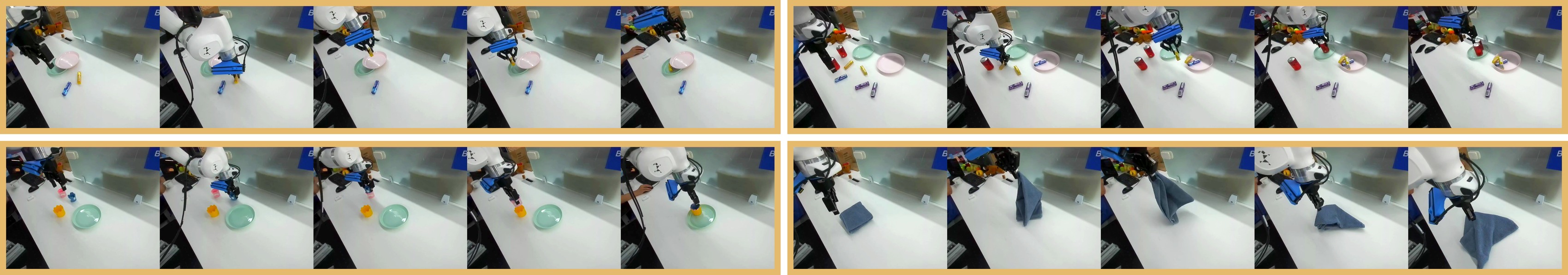}
\caption{\textbf{Example rollouts of ActionCodec2 on the self-designed DROID-Franka tasks.}}
\label{fig:droid-rollouts}
\end{figure}

The four designed DROID-Franka task suites have scenes, goals, and camera placements that do not occur in the DROID dataset. The baselines are FAST and X-Tokenizer, the two other pretrained general-purpose tokenizers. FAST and the pretrained ACodec2 variants are evaluated without additional tokenizer fitting. X-Tokenizer is fine-tuned on DROID as recommended by its authors, while the corpus-specific ACodec2 variants are fitted on DROID (Appendix~\ref{app:tokenizers}). We measure task progress as the fraction of prescribed milestones completed. Figure~\ref{fig:droid} reports the aggregate results, and Figure~\ref{fig:droid-breakdown} in Appendix~\ref{app:benchmarks} gives the milestone-level breakdown for each suite. ACodec2+ reaches $73.9\%$ task progress, compared with $50.8\%$ for FAST and $52.8\%$ for X-Tokenizer; all ACodec2 variants outperform both baselines. In the recorded rollouts, the streaming policies qualitatively show fewer periodic pauses than chunkwise decoding, and ACodec2 shows fewer corrective reversals near grasp and release events, while the baselines more often alternate between opposing adjustments. The right panel of Figure~\ref{fig:droid} reports inference latency, the wall-clock time from submitting an observation to obtaining the corresponding decoded action output: $607\,\mathrm{ms}$ for FAST and $467\,\mathrm{ms}$ for X-Tokenizer, against about $66\,\mathrm{ms}$ for ACodec2, $50\,\mathrm{ms}$ for ACodec2+, and $29\,\mathrm{ms}$ for ACodec2+$^{\mathrm{II}}$. A $15\,\mathrm{Hz}$ control period is $66.7\,\mathrm{ms}$, so the ACodec2 variants respond within one control period, whereas FAST and X-Tokenizer need about nine and seven. Representative rollouts for all four suites are shown in Figure~\ref{fig:droid-rollouts}.

\begin{figure}[t]
\centering
\scalebox{1.0}{\includegraphics[width=\linewidth]{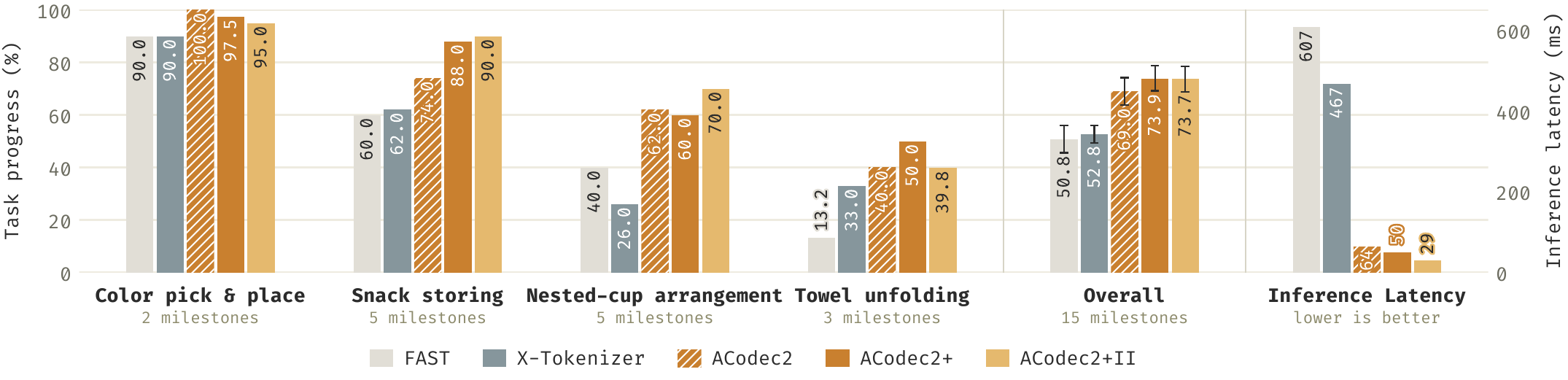}}
\caption{\textbf{DROID-Franka real-robot task progress (\%).} Five policies on the four self-designed task suites, $50$ rollouts per policy, scored by task progress (Appendix~\ref{app:benchmarks}). Policies are trained on DROID and evaluated zero-shot in unseen scenes; error bars on Overall are $90\%$ stratified bootstrap CIs. The right panel reports per-policy inference latency (ms, right axis), the wall-clock time from submitting an observation to obtaining the corresponding decoded action output.}
\label{fig:droid}
\end{figure}

\subsection{ActionCodec2 under reinforcement learning}
\label{sec:exp-rl}

\begin{figure}[t]
\centering
\scalebox{0.9}{%
\begin{minipage}[t]{0.56\textwidth}
\centering
\includegraphics[width=\linewidth]{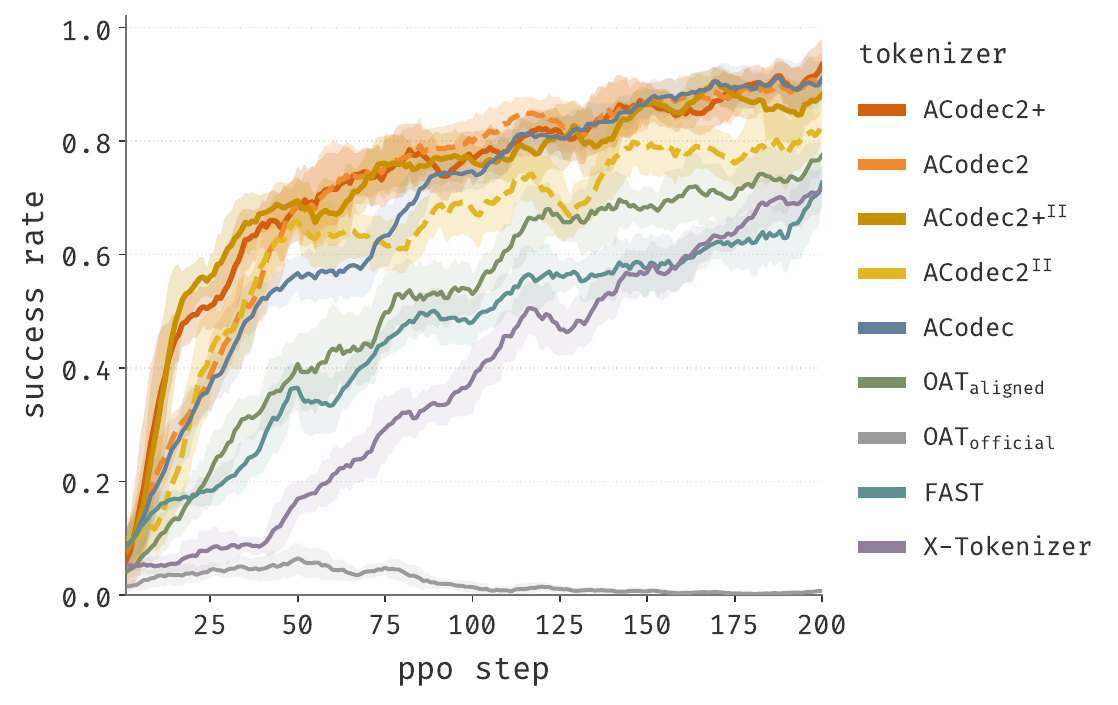}
\end{minipage}\hfill
\begin{minipage}[t]{0.39\textwidth}
\centering
\includegraphics[width=\linewidth]{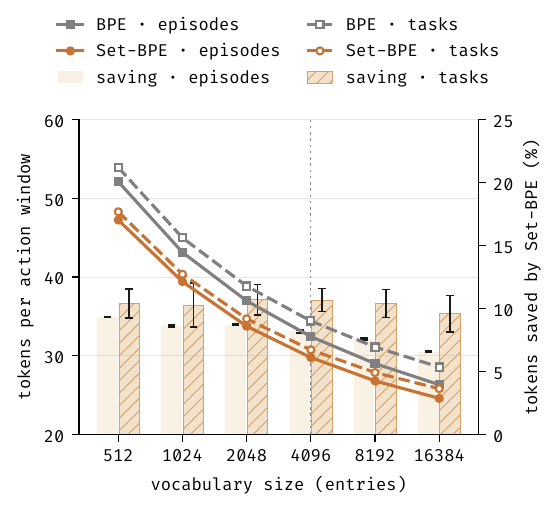}
\end{minipage}%
}
\caption{\textbf{PPO learning curves (left) and packing compression (right).} Left: hardened LIBERO-Object, with objects and basket re-placed at random; policies are initialized from checkpoints with about $5\%$ success, then trained for $200$ PPO iterations with $320$ rollouts each. Solid curves show the rollout success rate as a rolling mean, shaded bands $\pm$ one rolling standard deviation. Right: lines show mean tokens per action window, bars the token saving over BPE; \emph{unseen episodes} and \emph{unseen tasks} are held out from fitting; the dotted line marks the deployed $4{,}096$-entry vocabulary.}
\label{fig:ppo-packing}
\end{figure}

We evaluate RL fine-tuning with PPO \citep{ppo}, following prior work on AR VLAs \citep{guo2025irevla, xu2025rldg, liu2025rlvla}. We use a harder variant of LIBERO-Object in which the positions of all objects and the target basket are randomized at the start of each episode. All policies are first fine-tuned on standard LIBERO until they reach about $5\%$ success on this modified suite, and are then trained for $200$ PPO iterations with $320$ rollouts per iteration. The ACodec2 variants reach $83$--$93\%$ success after PPO training (Figure~\ref{fig:ppo-packing}, left). ACodec2$^{\mathrm{II}}$ is less stable during training and finishes near $83\%$, while the other ACodec2 variants reach about $93\%$. Among the baselines, ACodec reaches $91\%$; $\text{OAT}_{\text{aligned}}$, FAST, and X-Tokenizer reach $72$--$78\%$; and $\text{OAT}_{\text{official}}$ shows little improvement. In standard LIBERO, the target basket always appears on the left, and the initial policies consequently exhibit a strong leftward bias even when the basket is randomized. The ACodec2 variants correct this bias early in PPO training, whereas the other tokenizers adapt more slowly or retain it throughout training.

\subsection{Feedback within an action chunk}
\label{sec:exp-feedback}

\begin{figure}[t]
\centering
\includegraphics[width=\linewidth]{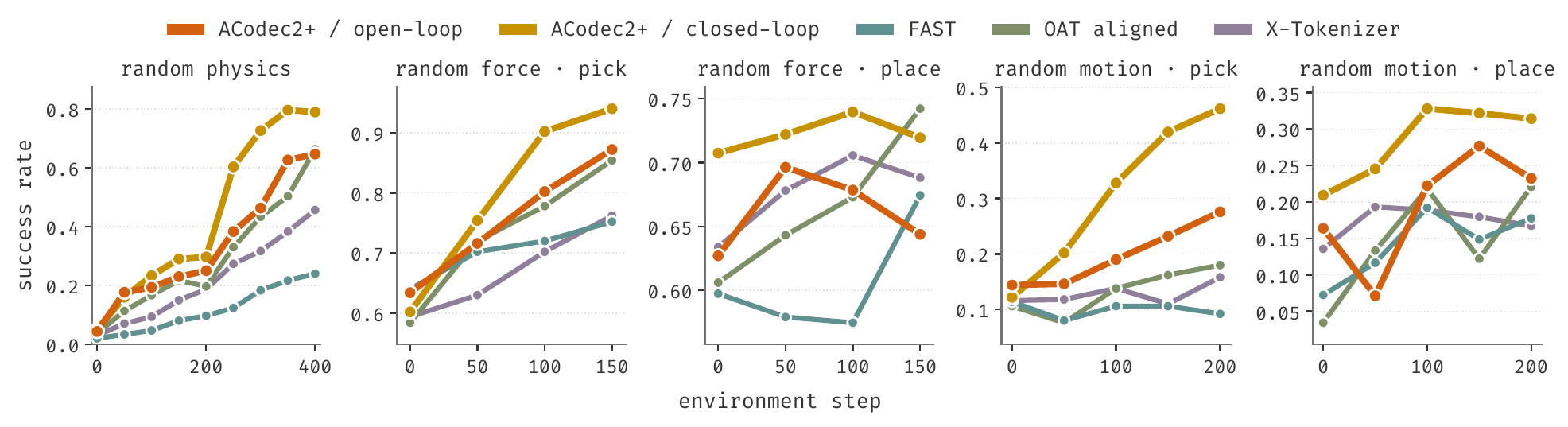}
\vspace{-20pt}
\caption{\textbf{PPO training under in-chunk perturbations.} Random physics: task success over six tasks; the first $200$ iterations train on the easy parameter ranges and the rest on the hard ranges, and every evaluation uses the hard ranges. Random force and random motion: \emph{pick} is the fraction of episodes in which the target object is grasped, and \emph{place} is the fraction of these episodes in which it is then placed in the basket; ten tasks each. Every point averages $50$ fixed seeds per task.}
\label{fig:feedback}
\end{figure}

Because ACodec2 tokens decode step by step, observations can be inserted into the context while a chunk is being generated, and the remaining actions of the chunk are then conditioned on them. We test this in three LIBERO-based suites in which a chunk executed open loop drifts from its intended outcome (Appendix~\ref{app:feedback}). \emph{Random physics} randomizes payload mass, joint damping and friction, and surface friction in six tasks. \emph{Random force} pushes the gripper once per second with a $150$--$180\,\mathrm{N}$ impulse in a random direction. \emph{Random motion} moves every object and the basket along random planar paths at $0.18$--$0.30\,\mathrm{m/s}$. ACodec2+ / open-loop and the baselines receive the robot state only at the start of a chunk; ACodec2+ / closed-loop also receives the measured robot state or object positions after every executed action, before the next action token is predicted. Each policy is trained with PPO from their LIBERO checkpoints. After training, the closed-loop variant reaches $79.0\%$ task success in random physics, compared with $64.7\%$ for open-loop ACodec2+, and grasp success of $94.0\%$ versus $87.2\%$ in random force and $46.2\%$ versus $27.6\%$ in random motion (Figure~\ref{fig:feedback}). In random physics, the PD controller of LIBERO follows its target less accurately under a heavier payload or larger damping. The closed-loop policy sees the gap between the action and the measured state and raises the remaining control signals to compensate. In random motion, it keeps adjusting its actions to the moving object, while the open-loop variant and the baselines lag behind it.

\subsection{Ablations}
\label{sec:exp-ablation}

\textbf{Streaming under real-time control.} The evaluations above suspend environment execution during policy decoding. We now evaluate under a continuous $20\,\mathrm{Hz}$ control clock (Table~\ref{tab:ablation-detail}, upper panel): actions are consumed from an asynchronous FIFO queue while the physics engine continues to advance, and zero motion is issued whenever the queue is empty. Using the same ACodec2+ checkpoint, streaming execution achieves $91.0\%$ average success on LIBERO, compared with $73.1\%$ when the same predictions are released only after a complete chunk is decoded. It also outperforms the chunkwise tokenizers FAST ($35.9\%$), X-Tokenizer ($60.6\%$), ACodec ($84.5\%$), and $\text{OAT}_{\text{aligned}}$ ($87.2\%$).

\textbf{Set-BPE packing.} Across vocabulary sizes from $512$ to $16{,}384$, Set-BPE reduces token count by $6.6$--$10.7\%$ relative to BPE on unseen episodes and unseen tasks (Figure~\ref{fig:ppo-packing}, right). With a $4{,}096$-entry vocabulary, the mean number of tokens per $20$-step window decreases from $32.45$ to $29.79$ on unseen episodes and from $34.44$ to $30.76$ on unseen tasks. On the same ACodec2 primitive stream, replacing Set-BPE by BPE lowers LIBERO success by $2.6$ points, from $96.9\%$ to $94.3\%$ (Table~\ref{tab:ablation-detail}, lower panel).

\subsection{ActionCodec2 as an agent interface}
\label{sec:exp-agent}

ActionCodec2 exposes robot control as discrete motion tokens that decode directly into executable actions. An agent can emit these tokens directly and store successful token sequences as named macros for reuse across attempts.

\begin{figure}[t]
\centering
\scalebox{0.9}{\includegraphics[width=\linewidth]{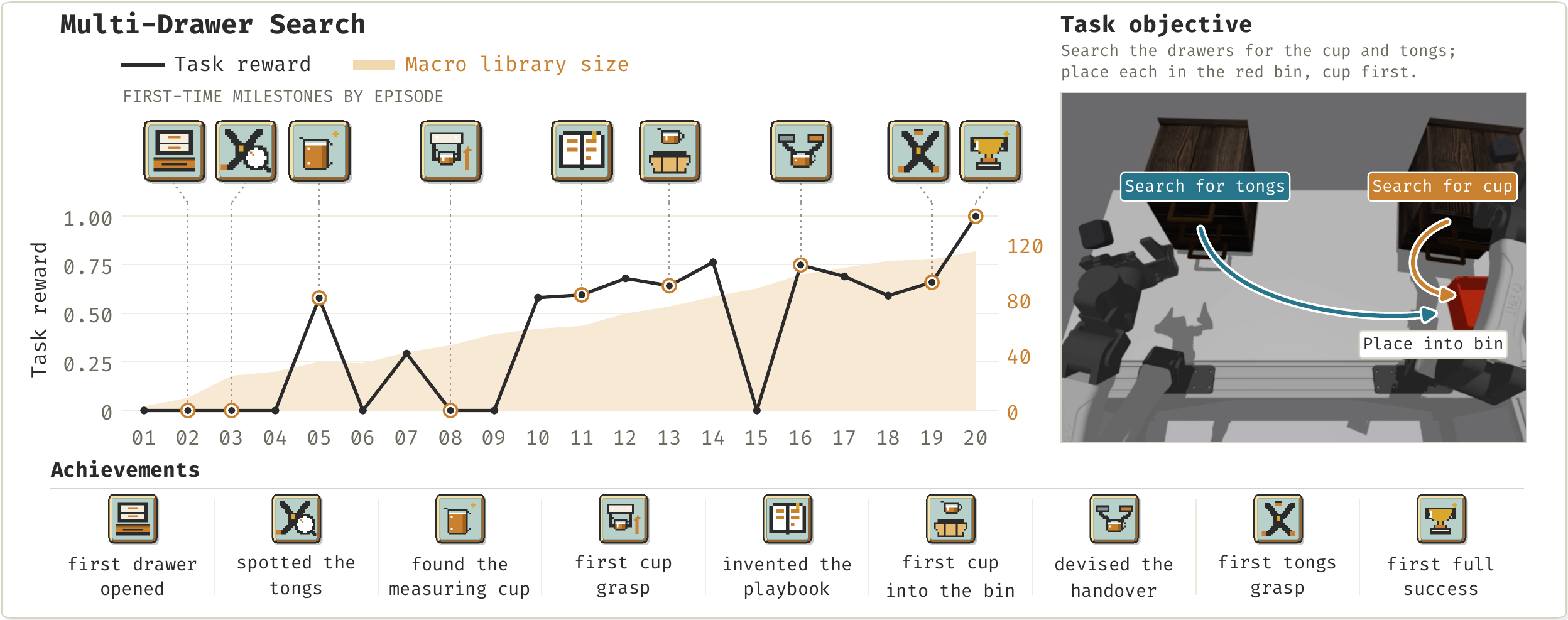}}
\caption{\textbf{Agent control through ActionCodec2.} Claude Opus 5.5 controls a simulated YAM robot in Multi-Drawer Search. The timeline shows maximum shaped reward, the size of the accumulated macro library, and selected first-time milestones; the first full task success occurs in episode~20.}
\label{fig:agent-control}
\end{figure}

\textbf{Multi-Drawer Search.}
We first test this interface in a long-horizon bimanual task. Claude Opus 5.5~\citep{claude-opus-5-5} controls a simulated YAM robot~\citep{i2rt-yam} to search two cabinets for a measuring cup and tongs and place them in a bin in the required order. Across $20$ attempts, operator-curated reflections and an accumulating library of ActionCodec2 macros are carried forward. The agent progressively acquires routines for opening drawers, grasping, transport, and handover, and first completes the full task in episode~20 (Figure~\ref{fig:agent-control}). Because both the reflections and the macro library change across attempts, this experiment demonstrates that ActionCodec2 motions can be composed and reused by an agent rather than isolating the effect of the tokenizer itself. Appendix~\ref{sec:drawer-milestones} gives the milestone-level progression.

\begin{figure}[t]
\centering
\includegraphics[width=\linewidth]{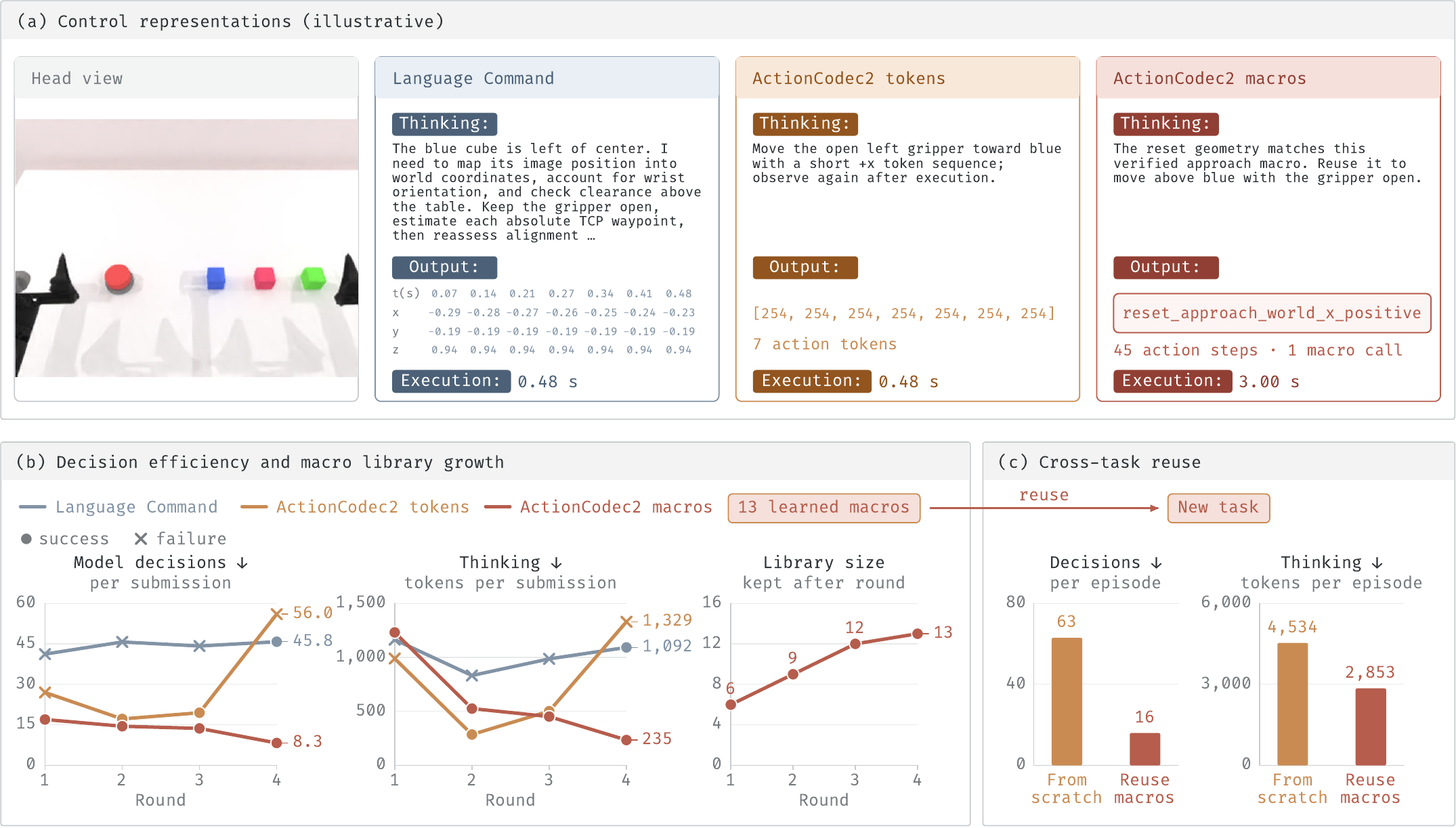}
\caption{\textbf{Agent control interfaces on RMBench.} (\emph{a})~The three control representations. (\emph{b})~Decision count, thinking-token usage, and macro-library growth over four chained rounds. (\emph{c})~Transfer of the learned macro library to a held-out rearrangement task.}
\label{fig:rmbench}
\end{figure}

\textbf{Controlled comparison on RMBench.}
We next hold the agent, task, and episodic memory fixed and vary only the control interface. GPT-6 Astra~\citep{gpt-6-astra} (low reasoning effort) performs RMBench \emph{blocks-try}~\citep{rmbench}, where it must identify the correct ordering of three blocks by trial among six possible permutations. We compare three interfaces (Figure~\ref{fig:rmbench}\emph{a}): (\emph{i})~\textbf{Language Command}, which specifies an absolute target TCP pose for each $0.5\,\mathrm{s}$ control window; (\emph{ii})~\textbf{ActionCodec2 tokens}, which directly specify the motion over the same window; and (\emph{iii})~\textbf{ActionCodec2 macros}, which additionally allows reusable token sequences to be called and carried across rounds. All three conditions receive the same episodic reflections.

Direct ActionCodec2 tokens require fewer control decisions and fewer thinking tokens than absolute pose commands (Figure~\ref{fig:rmbench}\emph{b}). With the macro library inherited across rounds, decisions per submission decrease from $17.0$ to $8.3$ over four rounds, compared with about $45$ under absolute pose control. The learned macros also transfer to a held-out block-rearrangement task, evaluated with Claude Opus 5.5 (low reasoning effort) in one episode per condition: starting without prior macros, the agent uses $63$ decisions and fails, whereas initialization with the $13$ macros acquired during training reduces this to $16$ decisions and succeeds (Figure~\ref{fig:rmbench}\emph{c}). These macros are named, text-described routines stored in the agent context rather than trained into the model weights (Table~\ref{tab:rmbench-macros}).

\section{Conclusion, limitations, and future work}
\label{sec:limitations}

In this paper, we introduce ActionCodec2, a streaming action tokenizer for AR VLAs. It quantizes each control step relative to the decoder's reconstruction, so every step corrects the error left by earlier steps and the tracking error stays within a prescribed tolerance. The primitives then only need to cover the one-step motion range rather than the workspace, and Set-BPE packs them losslessly into $8.8$ tokens per second at $15\,\mathrm{Hz}$. With the same VLA backbone, ActionCodec2 achieves state-of-the-art results on all evaluated benchmarks, and because each step has its own tokens, the policy can also use observations within a chunk. The pretrained tokenizers cover end-effector speeds up to about $1.2\,\mathrm{m/s}$, and on highly dynamic motion the compression may fall below that of chunk tokenizers, which use a longer window. In our experiments, frontier models merged ActionCodec2 tokens into macros with richer semantics and kept refining them across attempts, showing a trend toward recursive self-improvement; we leave further study of this to future work.

\newpage

\bibliography{iclr2027_conference}
\bibliographystyle{plainnat}

\newpage

\appendix

\section*{Appendix overview}
\begin{itemize}[topsep=0pt, itemsep=1pt, leftmargin=1.1em]
\item Appendix~\ref{app:related}: related work.
\item Appendix~\ref{app:proofs}: proofs for Sections~\ref{sec:general} and~\ref{sec:euclidean}.
\item Appendix~\ref{app:algorithms}: exact greedy fitting of Set-BPE.
\item Appendices~\ref{app:tokenizers}, \ref{app:benchmarks}, and~\ref{app:feedback}: baselines, benchmarks, per-suite real-robot results, and the feedback environments.
\item Appendix~\ref{app:agent-details}: macro library, Multi-Drawer milestones, and their narrative.
\item Appendices~\ref{sec:second-general}, \ref{sec:compare}, and~\ref{sec:lossless}: the second-order scheme (Section~\ref{sec:fitting}), the comparison with chunk tokenizers, and the losslessness of Set-BPE (Section~\ref{sec:setbpe}).
\end{itemize}

\section{Related Work}
\label{app:related}

\textbf{Robot foundation models.}
Two families of large models are used for general-purpose robot control. Vision--language--action (VLA) models are built on vision--language models \citep{rt2, openvla, pi0, pi05, gr00t-n1}, and world--action models (WAMs) are built on video generation models \citep{motus, fastwam, cosmospolicy}. Recent studies find that the training data affect performance more than the choice between the two \citep{openwam, xiaomirobotics1, shi2025diversity}. VLAs further differ in how they output actions. Some attach a separate diffusion or flow-matching action expert to the vision--language model \citep{pi0, pi05, gr00t-n1}. Others, which we call discrete VLAs, convert actions into tokens and predict them with the same autoregressive head that produces text \citep{rt2, openvla, fast, rdt2, walloss, walloss05, g05}. Because a discrete VLA is an ordinary language model, it can be trained and served with existing LLM infrastructure \citep{llamafactory, rlinf, sglang}, trains faster than a VLA with a flow-matching expert \citep{liu2026faster, pi06}, and supports chain-of-thought reasoning without modification \citep{cot-vla}. How well a discrete VLA trains and acts depends largely on its action tokenizer \citep{fast, dong2026actioncodec}.

\textbf{Action tokenizers.}
Per-dimension binning in RT-2 and OpenVLA \citep{rt2, openvla} decodes actions step by step. It spends one token on every action dimension, however, so the token sequences are long and training is inefficient \citep{fast}. Other tokenizers represent an action chunk of one to two seconds. FAST \citep{fast} applies a discrete cosine transform to the chunk and compresses the quantized coefficients with BPE. BEAST \citep{beast} fits B-splines to a chunk and quantizes temporally ordered control points; the local support of its spline basis allows early segments to be evaluated once the relevant control points are available. VQ-VAE-based tokenizers \citep{vq-bet, mete2024quest, vq-vla, liu2026faster, oat} learn an encoder that maps the chunk to a few discrete codes, and ActionCodec \citep{dong2026actioncodec} and X-Tokenizer \citep{x-tokenizer} add training objectives to this design that make the tokens easier for a VLA to predict. These tokenizers represent a fixed motion horizon. OAT \citep{oat} uses coarse-to-fine tokens for a complete action chunk rather than a stepwise symbol representation. Real-time chunking \citep{black2025real} improves reactivity by replanning overlapping chunks before the current chunk is exhausted, trading shorter reaction intervals for more frequent replanning. None of these tokenizers bounds the error that accumulates along the decoded trajectory. ActionCodec2 is streaming and keeps the trajectory error within a user-specified tolerance at every step.

\section{Proofs}
\label{app:proofs}
\hypertarget{proofs}{}

Each statement in Sections~\ref{sec:general} and~\ref{sec:euclidean} is restated here before its proof; the theorem number in the body links to the restatement, and the number in the restatement links back.

\hypertarget{proof:lem:onestep}{}
\paragraph{Lemma~\ref{lem:onestep} (One step suffices), restated.}
\emph{If the recoverability condition holds, then from every exact start
$\hat q_0=q_0$ and along every feasible trajectory, one can select a
symbol at each step such that $\xi_t\in\mathcal E$ for all $t$.
The symbol at step $t$ can be selected using only
$(q_t,\hat q_t,q_{t+1})$.}

\begin{proof}
Since $\hat q_0=q_0$ and the identity belongs to $\mathcal E$,
we have $(q_0,\hat q_0)\in\mathcal S$.
Suppose $(q_t,\hat q_t)\in\mathcal S$.
Because the trajectory is feasible,
$q_{t+1}\in\mathcal F(q_t)$.
By the recoverability condition, there exists a symbol $z_t\le K$ such that
\[
\bigl(q_{t+1},y_{z_t}(\hat q_t)\bigr)\in\mathcal S.
\]
Setting
$\hat q_{t+1}=y_{z_t}(\hat q_t)$
therefore gives $\xi_{t+1}\in\mathcal E$.
Applying the same argument at every subsequent step proves the claim
by induction.
The choice of $z_t$ requires only $(q_t,\hat q_t,q_{t+1})$ and no
information about later steps.
\end{proof}

\hypertarget{proof:thm:composition-optimal}{}
\paragraph{Theorem~\ref{thm:composition-optimal} (Optimality of fixed primitives), restated with part (4).}
\emph{Under the recoverability condition, (1) a fixed-primitive decoder with primitive set $\mathcal U\subseteq G$ meets it if and only if $\mathcal E\mathcal A\subseteq\bigcup_{u\in\mathcal U}u\mathcal E$; (2) every decoder of the general form \eqref{eq:general-step} that meets it uses at least $\min\bigl\{|\mathcal U|:\mathcal U\subseteq G\text{ finite},\ \mathcal E\mathcal A\subseteq\bigcup_{u\in\mathcal U}u\mathcal E\bigr\}$ symbols, the minimum over an empty family read as $+\infty$; (3) hence $K^\star$ equals that minimum, and a fixed-primitive decoder attains it whenever it is finite.  (4) All three survive a fixed distortion of the error between steps: if $q_{t+1}=\varphi(q_t)\delta_t$ for an automorphism $\varphi$ of $G$ and the decoder composes after it, $y_k(\hat q)=\varphi(\hat q)u_k$, then $\xi_{t+1}=u^{-1}\varphi(\xi_t)\delta_t$ and (1)--(3) hold with $\mathcal E\mathcal A$ replaced by $\varphi(\mathcal E)\mathcal A$.}

\begin{proof}
\textbf{Part 1.}
For a fixed-primitive decoder the error recursion closes on itself: with $\xi_t=\hat q_t^{-1}q_t$ and $q_{t+1}=q_t\delta_t$,
\begin{equation}
\xi_{t+1}
=(\hat q_tu_{z_t})^{-1}(q_t\delta_t)
=u_{z_t}^{-1}\xi_t\delta_t ,
\label{eq:pointwise-group-condition}
\end{equation}
in which the reconstructed configuration has cancelled, so the next error depends on the current error and the realized motion alone.  Hence $u$ answers the pair $(\xi,\delta)$ iff $u^{-1}\xi\delta\in\mathcal E$, i.e.\ iff $\xi\delta\in u\mathcal E$.

A pair in $\mathcal S$ is $q=\hat q\xi$ with $\xi\in\mathcal E$, and by the motion model its successors are $q^+=\hat q\xi\delta$ with $\delta\in\mathcal A$; by the previous paragraph the pair reached after executing $u$ lies in $\mathcal S$ iff $\xi\delta\in u\mathcal E$.  Quantifying over all $\xi\in\mathcal E$ and all $\delta\in\mathcal A$, the recoverability condition \eqref{eq:full-tolerance} is therefore exactly $\mathcal E\mathcal A\subseteq\bigcup_{u\in\mathcal U}u\mathcal E$, and Lemma~\ref{lem:onestep} turns it into tracking over an infinite horizon.  Conversely, if that inclusion fails there are $\xi\in\mathcal E$ and $\delta\in\mathcal A$ that no primitive corrects, so \eqref{eq:full-tolerance} fails at every pair carrying error $\xi$.

\textbf{Part 2.}
Fix any reconstruction $\hat q$ and set $c_k:=\hat q^{-1}y_k(\hat q)$, so that $y_k(\hat q)=\hat qc_k$.  This change of coordinates is available at every $\hat q$ because $G$ is a group, and it requires no assumption on the maps: at a single configuration, every decoder \eqref{eq:general-step} \emph{is} a fixed-primitive decoder, with a list $c_1,\ldots,c_K$ that may depend on $\hat q$.  Part 2 shows that this dependence does not reduce the number of symbols.

Take any $\xi\in\mathcal E$, so the real configuration is $q=\hat q\xi$, and any $\delta\in\mathcal A$, so $q^+=\hat q\xi\delta$.  If symbol $k$ is chosen, the next error is
$y_k(\hat q)^{-1}q^+=(\hat qc_k)^{-1}(\hat q\xi\delta)=c_k^{-1}\xi\delta$, in which $\hat q$ cancels as in \eqref{eq:pointwise-group-condition}, and it is acceptable precisely when $\xi\delta\in c_k\mathcal E$.  Because the tolerance is complete, every $\xi\in\mathcal E$ and every $\delta\in\mathcal A$ must be handled at this same $\hat q$, so
\begin{equation}
\mathcal E\mathcal A
\subseteq
\bigcup_{k=1}^{K}c_k\mathcal E .
\label{eq:arbitrary-cover}
\end{equation}
The left-hand side does not depend on $\hat q$: every reconstruction faces one and the same covering problem, so adapting the list to $\hat q$ does not reduce $K$.  The set $\{c_1,\ldots,c_K\}$ satisfies $\mathcal E\mathcal A\subseteq\bigcup_{k=1}^{K}c_k\mathcal E$, so $K$ is at least the stated minimum.

\textbf{Part 3.}
Take $\mathcal U$ of minimum size with $\mathcal E\mathcal A\subseteq\bigcup_{u\in\mathcal U}u\mathcal E$.  Part 1 shows the decoder it builds meets the recoverability condition, Part 2 shows no such decoder is smaller.

\textbf{Part 4.}
Only two steps above used $\varphi=\mathrm{id}$, and both survive because $\varphi$ is a homomorphism.  In Part 1, $\xi_{t+1}=(\varphi(\hat q_t)u_{z_t})^{-1}\varphi(q_t)\delta_t=u_{z_t}^{-1}\varphi(\hat q_t^{-1}q_t)\delta_t=u_{z_t}^{-1}\varphi(\xi_t)\delta_t$, so $\hat q_t$ still cancels and $u$ answers $(\xi,\delta)$ iff $\varphi(\xi)\delta\in u\mathcal E$; quantifying over $\mathcal E$ and $\mathcal A$ replaces $\mathcal E\mathcal A$ by $\varphi(\mathcal E)\mathcal A$.  In Part 2, set $c_k:=\varphi(\hat q)^{-1}y_k(\hat q)$ instead, which is again available at every $\hat q$, and the same cancellation gives $\varphi(\mathcal E)\mathcal A\subseteq\bigcup_kc_k\mathcal E$ at every reconstruction.  Part 3 is unchanged.  The set being covered still does not depend on $\hat q$, which is what the lower bound uses.
\end{proof}

\hypertarget{proof:thm:translation-exact}{}
\paragraph{Corollary~\ref{thm:translation-exact} (Translation), restated.}
\emph{Let $G=(\R^d,+)$ with $\underline a_j\le\overline a_j$, box motion set $\mathcal A=\prod_j[\underline a_j,\overline a_j]$ with $\Delta a_j:=\overline a_j-\underline a_j$, and tolerance $\mathcal E=\prod_j[-\epsilon_j,\epsilon_j]$ with $\epsilon_j>0$. In this case, the covering condition $\mathcal E\mathcal A\subseteq\bigcup_{u\in\mathcal U}u\mathcal E$ becomes $\mathcal E+\mathcal A\subseteq\bigcup_{u\in\mathcal U}(u+\mathcal E)$, and the minimum alphabet size is $K^\star_{\rm trans}=\prod_jN_j$ with $N_j=\lceil1+\Delta a_j/2\epsilon_j\rceil$, attained by the axiswise uniform grid \eqref{eq:translation-grid} of Algorithm~\ref{alg:primitives}.  If the real trajectory stays in a workspace box, clipping each reconstructed successor to that box preserves the error guarantee.}

\begin{proof}
The claim is about the decoder class \eqref{eq:general-step}, not about fixed primitives alone, so start from Theorem~\ref{thm:composition-optimal}: by Parts 2 and 3, every decoder meeting the recoverability condition supplies a finite $\{c_1,\ldots,c_K\}\subseteq\R^d$ satisfying $\mathcal E+\mathcal A\subseteq\bigcup_{k}(c_k+\mathcal E)$, and $K^\star_{\rm trans}$ is the least cardinality of such a set.  It is therefore enough to bound arbitrary translate covers.  Here $\xi_{t+1}=\xi_t+\delta_t-u$, so the set to be covered is the box $\prod_j[\underline a_j-\epsilon_j,\overline a_j+\epsilon_j]$, of side $L_j=\Delta a_j+2\epsilon_j$, and $N_j=\lceil L_j/2\epsilon_j\rceil$.  Take one point in the $j$th side if $N_j=1$ and $N_j$ equally spaced points spanning it otherwise; these are more than $2\epsilon_j$ apart, since $N_j-1<L_j/2\epsilon_j$, and a single translate $u+\mathcal E$ spans at most $2\epsilon_j$ per axis, so it contains at most one of the $\prod_jN_j$ product points: no cover, product or not, is smaller.  Conversely the points of \eqref{eq:translation-grid} are at most $2\epsilon_j$ apart and start and end at $\underline a_j$ and $\overline a_j$, so their $\epsilon_j$-intervals tile the $j$th side without a gap.  Clipping toward a target inside the workspace cannot increase any coordinate error, so the cover survives it, and the count stays necessary as long as every axis of the workspace is wider than $\Delta a_j+2\epsilon_j$, so that the separated points above still fit inside it.
\end{proof}

\hypertarget{proof:prop:so3-exact}{}
\paragraph{Corollary~\ref{prop:so3-exact} (Orientation), restated.}
\emph{Let $G=\SO(3)$ with $R_{t+1}=R_t\Delta_t$ and $\hat R_{t+1}=\hat R_tU$, $\mathcal E=\mathcal B_R(\epsilon_R)$, and $\mathcal A=\mathcal B_R(b_R)$, with $0<\epsilon_R\le\pi$ and $0\le b_R\le\pi$. Then $\mathcal E\mathcal A=\mathcal B_R(\rho_R)$, the bounds \eqref{eq:so3-bounds} hold, $K^\star_{\rm rot}=\Theta\bigl((1+\beta_R)^3\bigr)$ whenever $\rho_R+\epsilon_R/2\le r_0<\pi$, and the lattice $\{\mathrm{Rot}(v):v\in L^3\}$ satisfies the covering condition.}

\begin{proof}
The bi-invariant angle metric gives $\mathcal B_R(\epsilon_R)\mathcal B_R(b_R)=\mathcal B_R(\rho_R)$: the triangle inequality one way, and splitting a rotation of angle at most $\rho_R$ about its own axis the other.  So $\mathcal B_R(\rho_R)$ is the set to be covered by $K$ translates of volume $V_R(\epsilon_R)$, which gives the lower bound.  For the upper bound take a maximal $\epsilon_R$-separated set in $\mathcal B_R(\rho_R)$; maximality makes its $\epsilon_R$-balls a cover, and its disjoint $\epsilon_R/2$-balls all lie in $\mathcal B_R(\min\{\pi,\rho_R+\epsilon_R/2\})$.  Finally $V_R(r)=\int_0^r\sin^2(\vartheta/2)\,d\vartheta\big/\!\int_0^\pi\sin^2(\vartheta/2)\,d\vartheta$ is bounded above and below by constants times $r^3$ on $[0,r_0]$, and $\rho_R/\epsilon_R=1+\beta_R$.  For the lattice, $\theta(\mathrm{Rot}(x)^{-1}\mathrm{Rot}(v))\le\lVert x-v\rVert$ for all $x,v\in\R^3$: the differential of $\mathrm{Rot}$ at $x$, trivialized by right translation, has singular values $1$ and $\lvert\sin(\lVert x\rVert/2)\rvert/(\lVert x\rVert/2)\le1$, so the image of the segment from $v$ to $x$ is a curve of length at most $\lVert x-v\rVert$, and the angle metric is the length of the shortest such curve.  Every $w\in\mathcal B_R(\rho_R)$ is $\mathrm{Rot}(x)$ for some $x$ with $\lVert x\rVert\le\rho_R$, hence $x\in[-\rho_R,\rho_R]^3$, and the nearest point $v\in L^3$ satisfies $\lVert x-v\rVert\le\frac{\sqrt3}{2}\cdot\frac{2\epsilon_R}{\sqrt3}=\epsilon_R$; so $\mathrm{Rot}(v)^{-1}w\in\mathcal B_R(\epsilon_R)$, which is the covering condition.  The count is $n_R\le\sqrt3\rho_R/\epsilon_R+3$ per axis, and $\rho_R/\epsilon_R=1+\beta_R$.  A list larger than $K^\star_{\rm rot}$ still meets the recoverability condition by Theorem~\ref{thm:composition-optimal}, so a larger list increases the alphabet size and leaves the error bound unchanged.
\end{proof}

\hypertarget{proof:prop:so3-frames}{}
\paragraph{Proposition~\ref{prop:so3-frames} (Frame equivalence), restated.}
\emph{With $R_{t+1}=\Delta^w_tR_t$ and $\hat R_{t+1}=U\hat R_t$, the minimum alphabet equals the body-frame $K^\star_{\rm rot}$ for the motion set $(\mathcal A^w_R)^\top$; the two coincide when $\mathcal A^w_R$ is closed under inversion.}

\begin{proof}
Transposition is a bijection of $\SO(3)$ that reverses products and preserves the intrinsic angle, so $\mathcal B_R(\epsilon_R)^\top=\mathcal B_R(\epsilon_R)$; applying it termwise turns the world-frame inclusion into the body-frame cover for $(\mathcal A^w_R)^\top$ with primitives $U_k^\top$, a size-preserving correspondence of the two feasible sets.
\end{proof}

\section{Exact greedy fitting of Set-BPE}
\label{app:algorithms}

\label{app:bpe-training}
Algorithm~\ref{alg:setbpe} states the greedy rule over all legal candidates; the lazy loop certifies the same winner without enumerating them.  Identical steps are aggregated into weighted records, and counters are kept only for candidates that occur.  A committed merge changes only the counts that involve the instances it replaces: every new count involves the new symbol, and the counts of surviving candidates can only decrease, so a stale priority is an upper bound on the current score, and a lazy priority queue processed in decreasing bound returns the exact maximum.  A self-merge, in which a symbol is merged with itself at offset $\delta$, cannot use one instance as both endpoints; its exact gain is the size of a maximum matching on the time-ordered paths induced by the offset, and left-to-right greedy matching is optimal on these paths.  One vectorized pass initializes the counters.  A few further details make the certificate exact in practice.  A spatial pair's exact score is the weighted intersection of the occurrence records of its two symbols, \eqref{eq:spatial-frequency}, and a temporal candidate $(P,R,\delta)$ occurs only where a start of $P$ is followed by a start of $R$ at offset $\delta$, so intersecting shifted start sets yields exactly its instance pairs.  Every unexamined candidate carries a stored bound $U\ge F$ on its current score, and version numbers keep a bound computed before an earlier merge from being reused after the records supporting it change; a candidate wins once its exact score matches or exceeds every remaining bound, with the tie-breaking of exhaustive enumeration.  The reference implementation and an optimized native kernel are kept merge-for-merge identical: both maintain exact counters over observed events only, break ties by count and then by the smallest symbol pair, and apply merges in creation order at encode time, so the trained vocabulary is reproducible and independent of the backend.  Legality of a temporal candidate is decided from per-symbol time-shifted support masks in time proportional to the span, so no table over the vocabulary is materialized.  Sparse and lazy enumeration reduce memory, not time.

\section{Experimental setup}

\subsection{Tokenizers}
\label{app:tokenizers}

The paragraphs below give per-method details for the tokenizers of Table~\ref{tab:tokenizer-properties}.

\paragraph{FAST.}
We use the official FAST release\footnote{\url{https://huggingface.co/physical-intelligence/fast}} \citep{fast} and follow its recommended best practices in all experiments: encoded action chunks span $1$--$2\,\mathrm{s}$, and every action dimension is normalized to $[-1,1]$ by per-dataset $q_{01}/q_{99}$ quantiles. FAST requires the predicted action tokens, after BPE inversion, to form an integer array of exactly shape $T\times D$; if the length differs, the reshape fails, decoding aborts, and the policy emits a stationary rest action for the chunk. A single extra or missing token therefore discards the whole chunk. We wrap the decoder with a \emph{soft decode}: an inverted sequence shorter than $T\times D$ is zero-padded to length, and an overlong one is truncated. The quantized DCT coefficients are concentrated at low frequencies and decay toward zero, so padding with zeros or trimming the tail changes the decoded signal little. Soft decoding improves FAST's downstream performance in our experiments, and every reported FAST number uses it. It changes only how a malformed decode is salvaged, never how actions are tokenized.

\paragraph{BEAST.}
BEAST \citep{beast} represents action chunks with quantized B-spline control points. We evaluate it with the same G0.5 backbone and training recipe as the other tokenizers on LIBERO and SimplerEnv.

\paragraph{X-Tokenizer.}
We use the officially released code and pretrained weights\footnote{\url{https://github.com/X-Square-Robot/X-Tokenizer}} \citep{x-tokenizer}, aligning our action layouts with the official action-dimension slots and with physical units. We use the released weights as they are for every robot type with an official configuration, including WidowX and Franka. Used zero-shot, these weights give large reconstruction errors and policies with low success. The authors told us that X-Tokenizer is designed to accelerate policy learning inside the $\pi_{0.5}$ architecture \citep{pi05}, not as a standalone tokenizer for discrete VLAs, and recommended fine-tuning the pretrained weights on each target dataset. Every X-Tokenizer number in this paper uses the official weights fine-tuned on the task dataset.

\paragraph{OAT.}
For OAT \citep{oat} we use the official release\footnote{\url{https://github.com/Chaoqi-LIU/oat}}. No pretrained version of OAT exists; the tokenizer is learned from scratch on every dataset. OAT uses an ordered code in which each successive token refines the decoded action. The official training configuration sets a token budget of eight tokens per action chunk, which compresses strongly and lowers closed-loop VLA performance. We report the official default as $\text{OAT}_{\text{official}}$ throughout; we additionally evaluate an aligned variant, $\text{OAT}_{\text{aligned}}$, whose token budget matches the average compression rate of ACodec2, so that the comparison isolates the tokenizer design from the bit rate.

\paragraph{ActionCodec.}
ACodec \citep{dong2026actioncodec} is a VQ-based action tokenizer with auxiliary losses that align its codes with visual and language features. We use the officially released checkpoints\footnote{\url{https://huggingface.co/ZibinDong/ActionCodec-5e}}; because they are already trained on LIBERO, BridgeData, and DROID, we apply them without further fine-tuning. The G0.5 backbone \citep{g05} used in our experiments states in its technical report that its internal action tokenizer is a modified ACodec, differing in network architecture, pretraining corpus, and engineering choices. The released ACodec we evaluate therefore does not share G0.5's action-token prior.

\paragraph{ACodec2 variants.}
A first-order ACodec2 is fit on each benchmark corpus on a single CPU node (Section~\ref{sec:fitting}). \textbf{ACodec2+} is the universal instance pretrained once on the roughly $20$k-hour, $24$-embodiment corpus of Section~\ref{sec:fitting}, applied to every benchmark by resampling alone with no per-corpus fitting. The stateful variants \textbf{ACodec2$^{\mathrm{II}}$} and \textbf{ACodec2+$^{\mathrm{II}}$} additionally track a velocity state under the second-order feasibility analysis of Appendix~\ref{sec:second-general}.

\paragraph{Baselines and training details.}
Every tokenizer is plugged into the same released G0.5 VLA \citep{g05}\footnote{\url{https://huggingface.co/OpenGalaxea/G05/tree/main/g05-base}}. All experiments share AdamW with weight decay $0.01$, gradient clipping at $1.0$, and a cosine schedule decaying to $10\%$ of the peak after a short linear warmup. On LIBERO \citep{libero}, every tokenizer is trained for $10$k steps at learning rate $2.5{\times}10^{-5}$, global batch $128$, and action-chunk horizon $20$ (of which $10$ are executed). On SimplerEnv, policies are trained on BridgeData \citep{walke2023bridgedata} for $50$k steps at learning rate $5{\times}10^{-5}$, global batch $64$, and horizon $10$. On DROID-Franka, policies are trained on DROID \citep{droid} for $50$k steps at learning rate $5{\times}10^{-5}$, global batch $1{,}344$, and horizon $15$.

\subsection{Universal tokenizer configurations}
\label{app:configs}

The universal tokenizers use two shared action representations, one for end-effector (EEF) control and one for joint control. Robot trajectories are converted into these representations in physical units before fitting and encoding.

\paragraph{End-effector actions.}
EEF actions consist of six-degree-of-freedom relative pose changes and a gripper action. Translation increments $(\Delta x,\Delta y,\Delta z)$ are expressed in meters, and rotation increments as three-dimensional rotation vectors in radians. Both use a right-handed world frame with $+X$ forward, $+Y$ left, and $+Z$ up. Relative translations are independent of the choice of world-frame origin.

\paragraph{Joint actions.}
Joint actions consist of relative joint-angle changes in radians and a gripper action. The representation has seven joint slots; arms with fewer joints pad unused slots with zero, represented by the exact-zero primitive. For dual-arm robots, the same single-arm tokenizer is applied independently to the left and right arms.

\paragraph{Action scales and control frequency.}
Actions are expressed in physical units without dataset-specific min--max or $z$-score normalization. The default position and orientation tolerances are $5\,\mathrm{mm}$ and $0.02\,\mathrm{rad}$; joint tolerances range from $4$ to $20\,\mathrm{mrad}$ and are scaled using joint kinematics from URDF models, corresponding to approximately $5\,\mathrm{mm}$ of end-effector displacement. Training trajectories are resampled to $15\,\mathrm{Hz}$ before tokenization. At deployment, decoded actions are resampled to the controller frequency: $20\,\mathrm{Hz}$ for LIBERO, $5\,\mathrm{Hz}$ for SimplerEnv, and $15\,\mathrm{Hz}$ for DROID-Franka.

\paragraph{Tokenizer variants.}
Universal configurations cover position tolerances of $1$, $5$, and $10\,\mathrm{mm}$ and Set-BPE vocabulary sizes of $1{,}024$, $2{,}048$, $4{,}096$, $8{,}192$, and $16{,}384$. They include EEF and joint actions, first- and second-order schemes, and binary (2-bin) or continuous (11- or 21-bin) grippers. The default experiments use a $5\,\mathrm{mm}$ tolerance and a $4{,}096$-token vocabulary. Changing the tolerance changes the primitive alphabet and its fitted merge table; changing vocabulary size selects a different number of ordered Set-BPE merges. The VLA action head is configured for the selected vocabulary.

\begin{table}[h]
\centering
\small
\caption{\textbf{Tolerance and vocabulary size.} First-order universal tokenizers for end-effector (EEF) and joint actions. Token rates are for EEF actions at $15\,\mathrm{Hz}$; values are rounded. The tolerance comparison uses the pretraining dataset, and the vocabulary comparison uses an evaluation slice.}
\label{tab:tolerance-configs}
\begin{tabular}{llrrrr}
\toprule
Setting & Position tolerance & EEF atoms & Joint atoms & Tokens/step & Tokens/s \\
\midrule
\multicolumn{6}{l}{\emph{Varying tolerance, vocabulary size 4,096}} \\
1 mm  & 1 mm  & 836 & 1,701 & 1.38 & 20.7 \\
5 mm  & 5 mm  & 176 & 353   & 0.58 & 8.8 \\
10 mm & 10 mm & 96  & 187   & 0.45 & 6.8 \\
\midrule
\multicolumn{6}{l}{\emph{Varying vocabulary size, tolerance 5 mm}} \\
1,024  & 5 mm & 176 & 353 & 1.09 & 16.3 \\
4,096  & 5 mm & 176 & 353 & 0.81 & 12.2 \\
16,384 & 5 mm & 176 & 353 & 0.71 & 10.6 \\
\bottomrule
\end{tabular}
\end{table}

Table~\ref{tab:tolerance-configs} reports the effect of tolerance and vocabulary size on token rate. The two comparisons use different data slices.

\subsection{Additional tokenizer and control measurements}
\label{app:additional-measurements}

\paragraph{Within-chunk tracking error.}
For 20-step action chunks, the fraction exceeding $5\,\mathrm{mm}$ tracking error is $48.3\%$ for FAST, $80.0\%$ for fine-tuned X-Tokenizer, $87.9\%$ for OAT-aligned, $91.9\%$ for ACodec, and $0.0\%$ for ACodec2. The corresponding 95th-percentile errors are $10.6$, $32.1$, $37.2$, $53.4$, and $4.55\,\mathrm{mm}$, respectively.

\paragraph{Token release.}
A temporal token fills the corresponding slots in the current and subsequent steps as soon as it is emitted. A later step already filled by previous tokens needs no new token at release. Tables~\ref{tab:tokens-per-step} and~\ref{tab:token-span} give the token and span distributions on the pretraining and held-out trajectories. In the 2,000-episode real-time LIBERO evaluation, $95.9\%$ of server emissions release one step, $3.9\%$ release two, and $0.26\%$ release three or four steps, across $324{,}740$ executed steps.

\begin{table}[h]
\centering
\small
\caption{\textbf{New tokens required to release an action step (\%).} A step may require zero new tokens when earlier temporal tokens already filled its slots.}
\label{tab:tokens-per-step}
\begin{tabular}{lrrrrrr}
\toprule
Tokenizer & 0 & 1 & 2 & 3 & $\geq4$ & Mean $\pm$ std. \\
\midrule
ACodec2+ EEF & 53.4 & 43.4 & 3.1 & 0.1 & 0.005 & $0.50\pm0.56$ \\
ACodec2+$^{\mathrm{II}}$ EEF & 39.5 & 57.9 & 2.5 & 0.1 & 0.0 & $0.63\pm0.54$ \\
ACodec2+ joint & 53.4 & 29.8 & 11.2 & 3.9 & 1.7 & $0.71\pm0.96$ \\
ACodec2+$^{\mathrm{II}}$ joint & 41.4 & 43.4 & 12.6 & 1.7 & 1.0 & $0.78\pm0.83$ \\
\bottomrule
\end{tabular}
\end{table}

\begin{table}[h]
\centering
\small
\caption{\textbf{Active steps covered per emitted token (\%).}}
\label{tab:token-span}
\begin{tabular}{lrrrr}
\toprule
Tokenizer and dataset & 1 step & 2 steps & 3 steps & $\geq4$ steps \\
\midrule
ACodec2+ EEF, pretraining & 74.6 & 18.3 & 3.7 & 3.4 \\
ACodec2+ joint, pretraining & 71.7 & 19.3 & 2.7 & 6.2 \\
ACodec2+ EEF, held-out & 59.9 & 20.7 & 4.7 & 14.7 \\
ACodec2+$^{\mathrm{II}}$ EEF, held-out & 71.8 & 17.7 & 2.8 & 7.6 \\
\bottomrule
\end{tabular}
\end{table}

\paragraph{Real-time latency.}
In the $20\,\mathrm{Hz}$ LIBERO evaluation, the client requests a new action window when its FIFO is empty; the simulator continues to advance, issuing zero motion until an action arrives. Table~\ref{tab:latency-distribution} reports client first-action latency across 2,000 episodes. On the server, streaming ACodec2+ emits the first action in $95\,\mathrm{ms}$ on average (P50 $99\,\mathrm{ms}$, P99 $145\,\mathrm{ms}$); subsequent action releases are separated by $48\,\mathrm{ms}$ on average (P50 $57\,\mathrm{ms}$, P99 $99\,\mathrm{ms}$).

\begin{table}[h]
\centering
\small
\caption{\textbf{Client first-action latency in real-time LIBERO (ms).} The client control clock runs at $20\,\mathrm{Hz}$.}
\label{tab:latency-distribution}
\begin{tabular}{lrrrrrr}
\toprule
Tokenizer / release & Requests & Mean & P50 & P90 & P95 & P99 \\
\midrule
FAST / chunk & 13,748 & 1,348 & 914 & 4,128 & 4,244 & 4,647 \\
X-Tokenizer / chunk & 17,668 & 684 & 672 & 722 & 751 & 891 \\
OAT-aligned / chunk & 15,898 & 661 & 656 & 689 & 702 & 784 \\
ACodec / chunk & 16,652 & 574 & 565 & 603 & 627 & 1,006 \\
ACodec2+ / full chunk & 14,950 & 1,023 & 1,001 & 1,299 & 1,400 & 1,591 \\
ACodec2+ / streaming & 16,237 & 122 & 124 & 147 & 157 & 172 \\
\bottomrule
\end{tabular}
\end{table}

\paragraph{More frequent replanning.}
Table~\ref{tab:replan-horizon} compares execution horizons of 20 and 10 steps on the 500-episode LIBERO-Object-Basket-Shuffle PPO evaluation at step 200.
\begin{table}[h]
\centering
\small
\caption{\textbf{PPO evaluation with different execution horizons (\%).} Policies are evaluated over 500 episodes at PPO step 200.}
\label{tab:replan-horizon}
\begin{tabular}{lrrr}
\toprule
Tokenizer & Sampled $H=20$ & Sampled $H=10$ & Greedy $H=20$ \\
\midrule
FAST & 70.2 & 70.6 & 71.8 \\
OAT-aligned & 74.2 & 73.2 & 78.2 \\
X-Tokenizer & 72.0 & 77.8 & 76.0 \\
ACodec & 83.8 & 90.0 & 84.8 \\
ACodec2 & 89.0 & 90.8 & 92.6 \\
\bottomrule
\end{tabular}
\end{table}

\paragraph{Constrained decoding.}
Each action token fills a known set of time--coordinate slots. During generation, a mask admits tokens that fill the next unfilled slot without overlapping filled slots. Single-coordinate primitives remain available at every slot. Across 12,510 evaluation episodes, no malformed ACodec2 decodes were observed. FAST evaluations use its fault-tolerant decoding, which pads or truncates the coefficient sequence to the required shape.

\subsection{Evaluation benchmarks}
\label{app:benchmarks}

\begin{figure}[t]
\centering
\begin{minipage}[t]{0.4325\linewidth}
  \centering
  \begin{minipage}[t]{0.4917\linewidth}
    \centering
    \includegraphics[width=\linewidth]{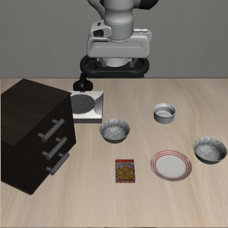}\\[1pt]
    {\scriptsize LIBERO-Spatial}
  \end{minipage}\hspace{0.0139\linewidth}%
  \begin{minipage}[t]{0.4917\linewidth}
    \centering
    \includegraphics[width=\linewidth]{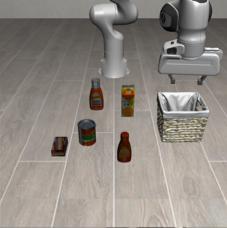}\\[1pt]
    {\scriptsize LIBERO-Object}
  \end{minipage}
\end{minipage}\hspace{0.035\linewidth}%
\begin{minipage}[t]{0.5325\linewidth}
  \centering
  \includegraphics[width=\linewidth]{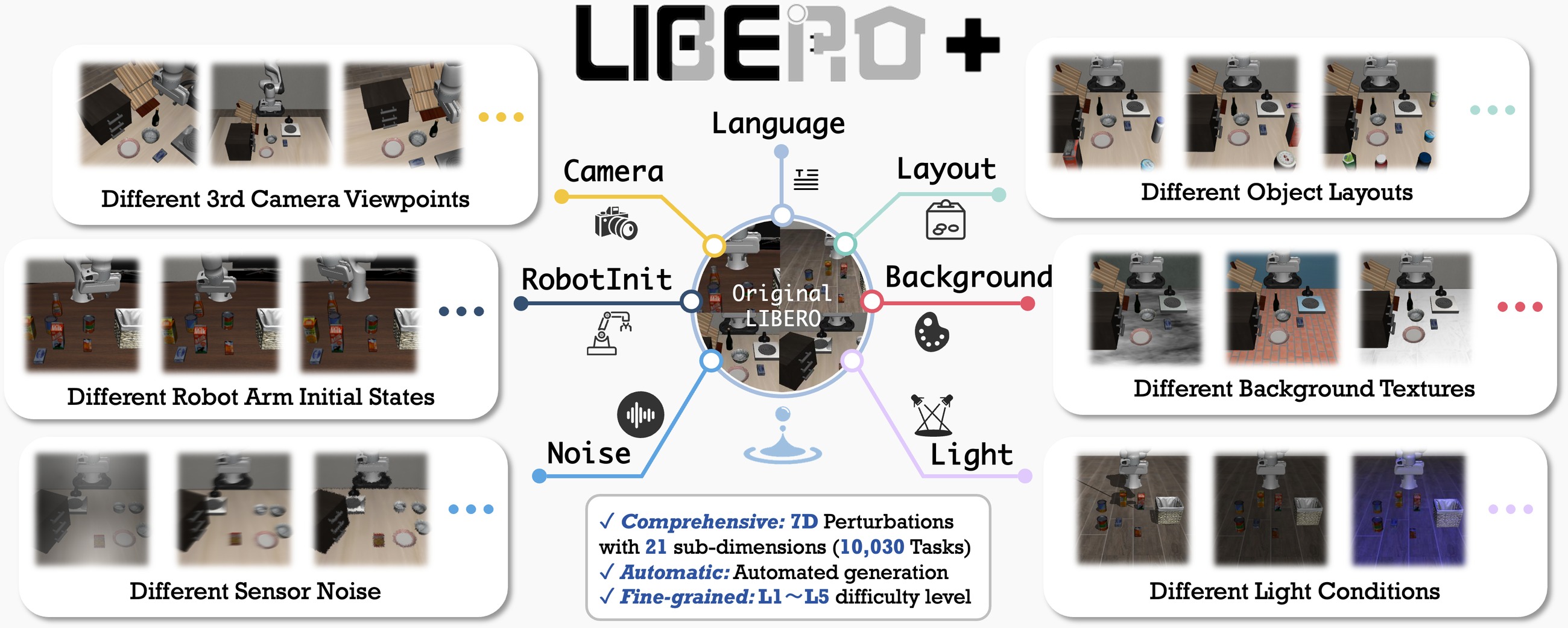}
\end{minipage}

\vspace{2pt}
\begin{minipage}[t]{0.4325\linewidth}
  \centering
  {\small\textbf{(a) LIBERO}}
\end{minipage}\hspace{0.035\linewidth}%
\begin{minipage}[t]{0.5325\linewidth}
  \centering
  {\small\textbf{(b) LIBERO-Plus} (seven perturbation axes)}
\end{minipage}

\vspace{5pt}
\begin{minipage}[t]{0.241\linewidth}
  \centering
  \includegraphics[width=\linewidth]{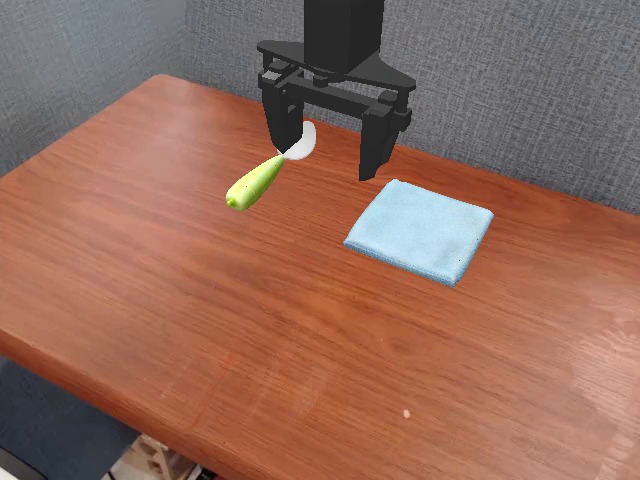}\\[1pt]
  {\scriptsize spoon on towel}
\end{minipage}\hspace{0.012\linewidth}%
\begin{minipage}[t]{0.241\linewidth}
  \centering
  \includegraphics[width=\linewidth]{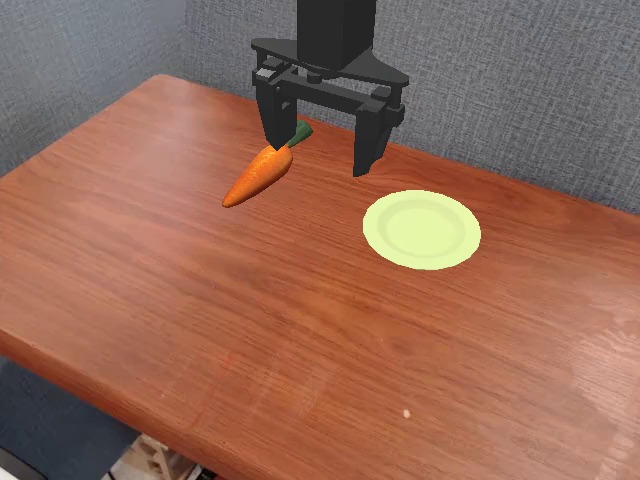}\\[1pt]
  {\scriptsize carrot on plate}
\end{minipage}\hspace{0.012\linewidth}%
\begin{minipage}[t]{0.241\linewidth}
  \centering
  \includegraphics[width=\linewidth]{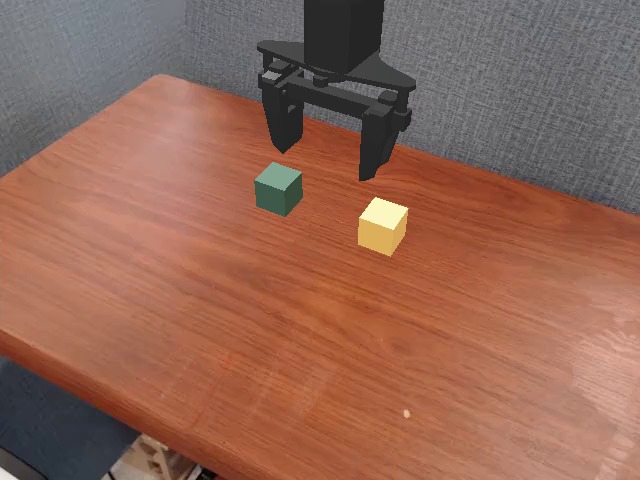}\\[1pt]
  {\scriptsize stack blocks}
\end{minipage}\hspace{0.012\linewidth}%
\begin{minipage}[t]{0.241\linewidth}
  \centering
  \includegraphics[width=\linewidth]{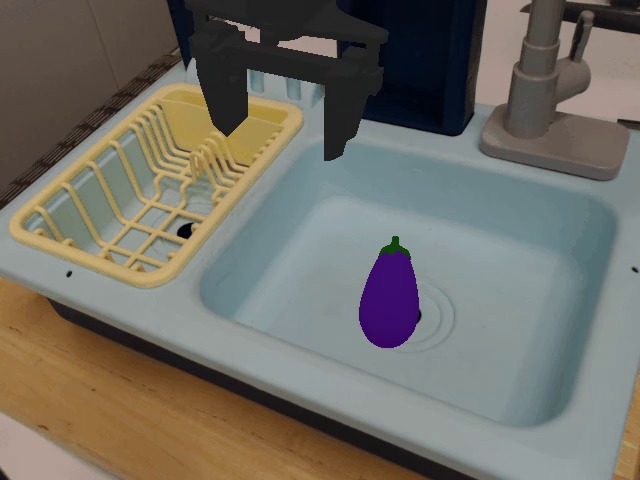}\\[1pt]
  {\scriptsize eggplant in basket}
\end{minipage}

\vspace{2pt}
{\small\textbf{(c) SimplerEnv-WidowX} (visual matching)}

\vspace{5pt}
\begin{minipage}[t]{0.241\linewidth}
  \centering
  \includegraphics[width=\linewidth]{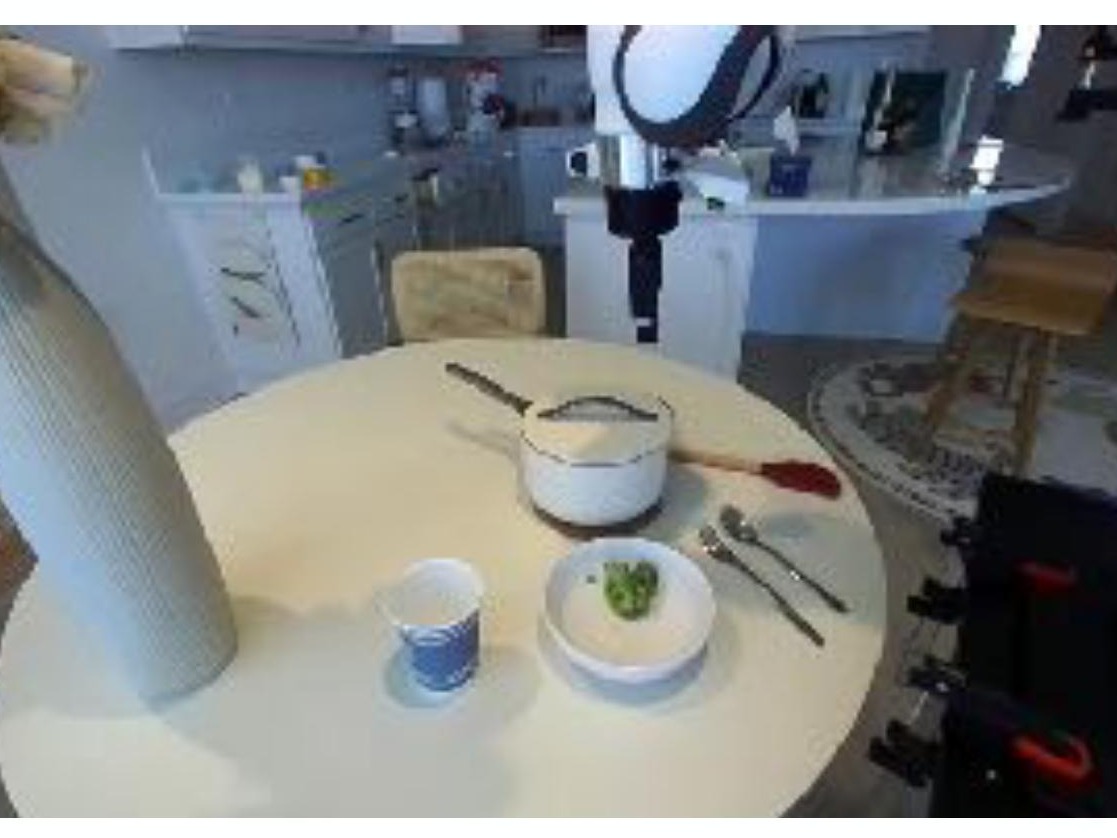}\\[1pt]
  {\scriptsize dining room}
\end{minipage}\hspace{0.012\linewidth}%
\begin{minipage}[t]{0.241\linewidth}
  \centering
  \includegraphics[width=\linewidth]{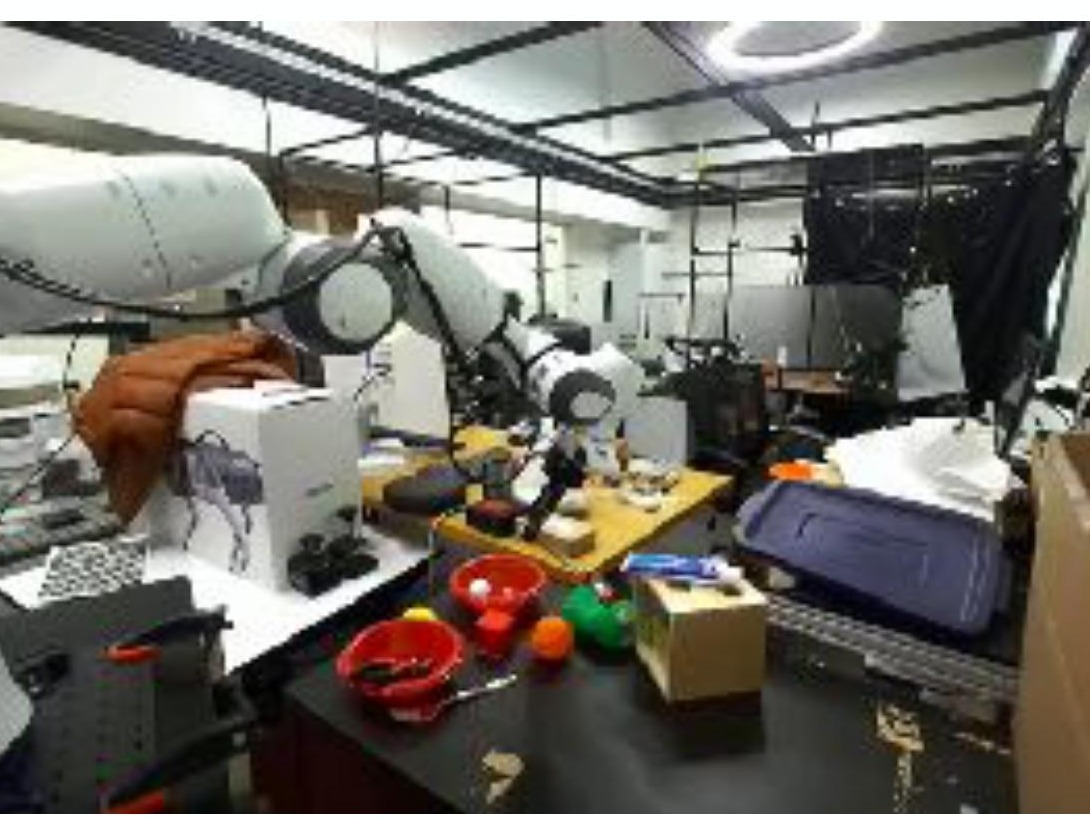}\\[1pt]
  {\scriptsize laboratory}
\end{minipage}\hspace{0.012\linewidth}%
\begin{minipage}[t]{0.241\linewidth}
  \centering
  \includegraphics[width=\linewidth]{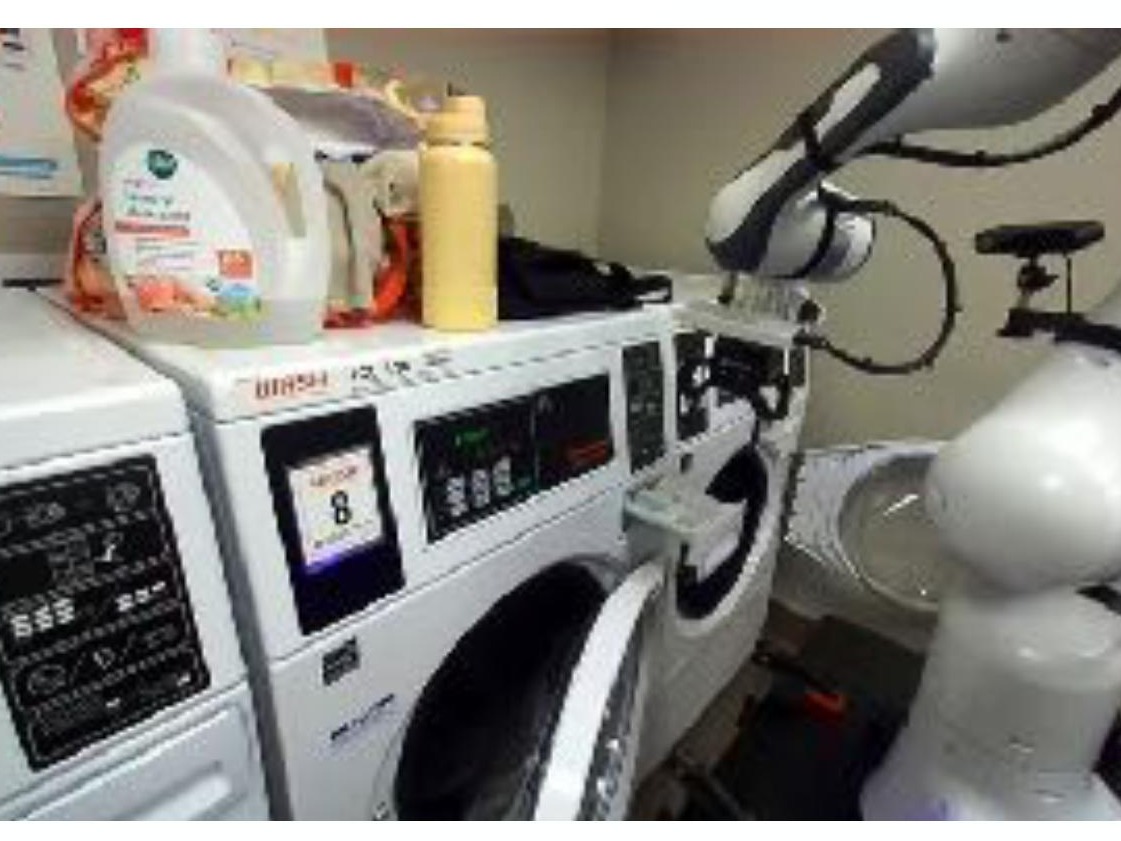}\\[1pt]
  {\scriptsize laundry room}
\end{minipage}\hspace{0.012\linewidth}%
\begin{minipage}[t]{0.241\linewidth}
  \centering
  \includegraphics[width=\linewidth]{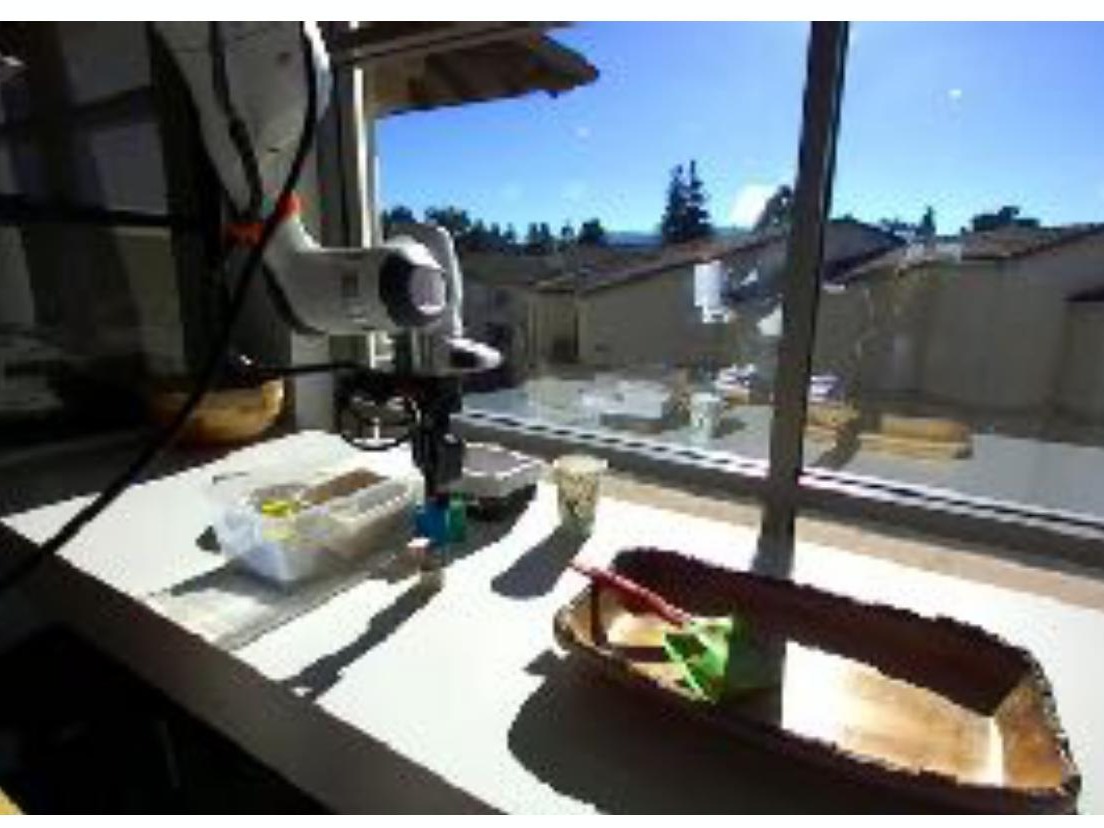}\\[1pt]
  {\scriptsize office}
\end{minipage}

\vspace{2pt}
{\small\textbf{(d) DROID-Franka} (in-the-wild scenes of the DROID training corpus)}
\vspace{5pt}

\begin{minipage}[t]{0.241\linewidth}
  \centering
  \includegraphics[width=\linewidth]{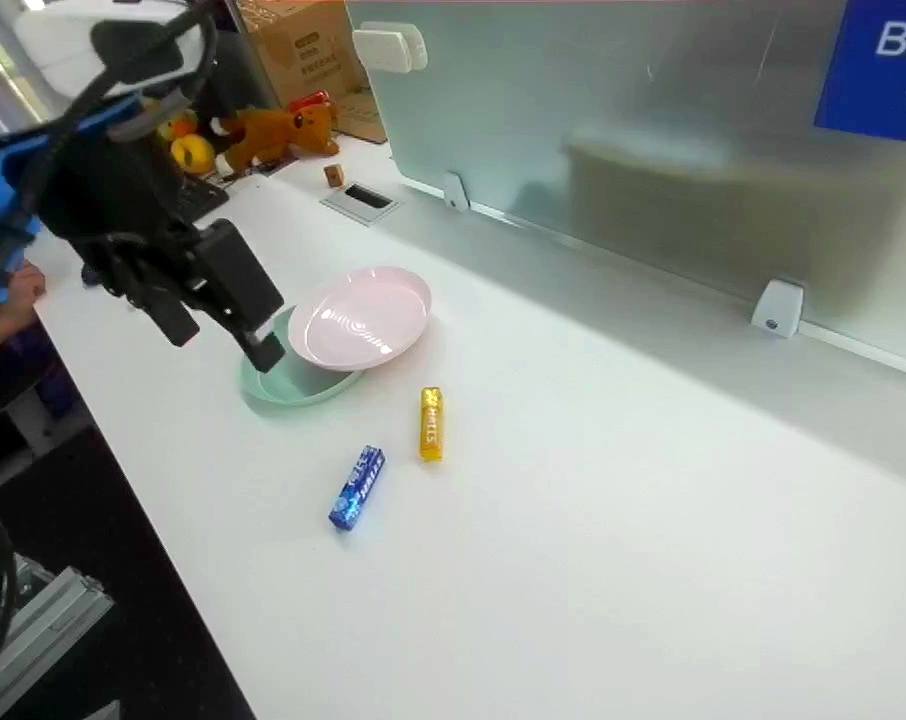}\\[1pt]
  {\scriptsize color pick-and-place}
\end{minipage}\hspace{0.012\linewidth}%
\begin{minipage}[t]{0.241\linewidth}
  \centering
  \includegraphics[width=\linewidth]{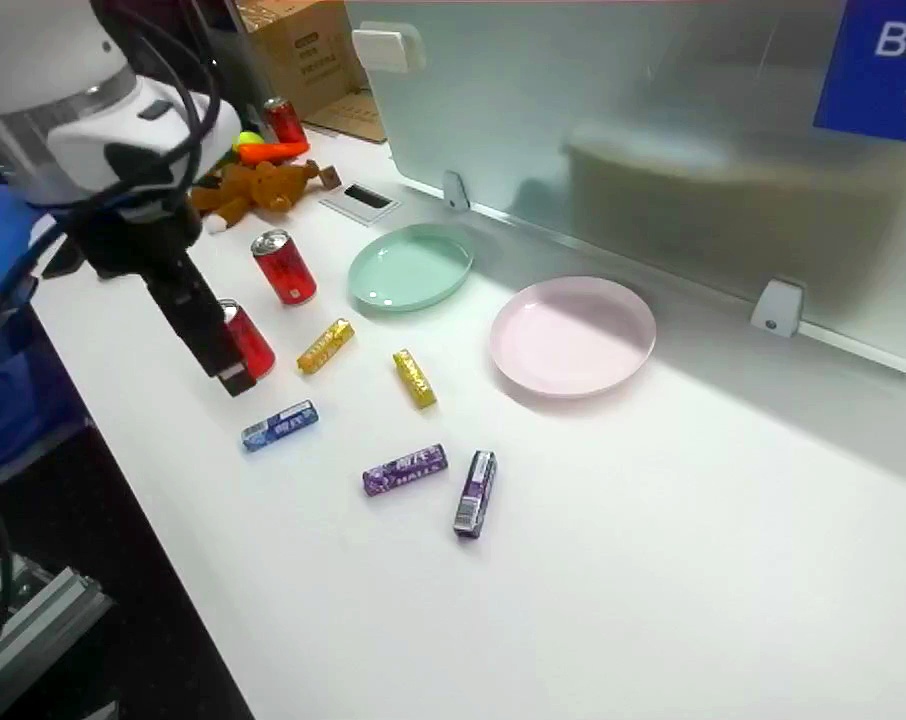}\\[1pt]
  {\scriptsize snack storing \& serving}
\end{minipage}\hspace{0.012\linewidth}%
\begin{minipage}[t]{0.241\linewidth}
  \centering
  \includegraphics[width=\linewidth]{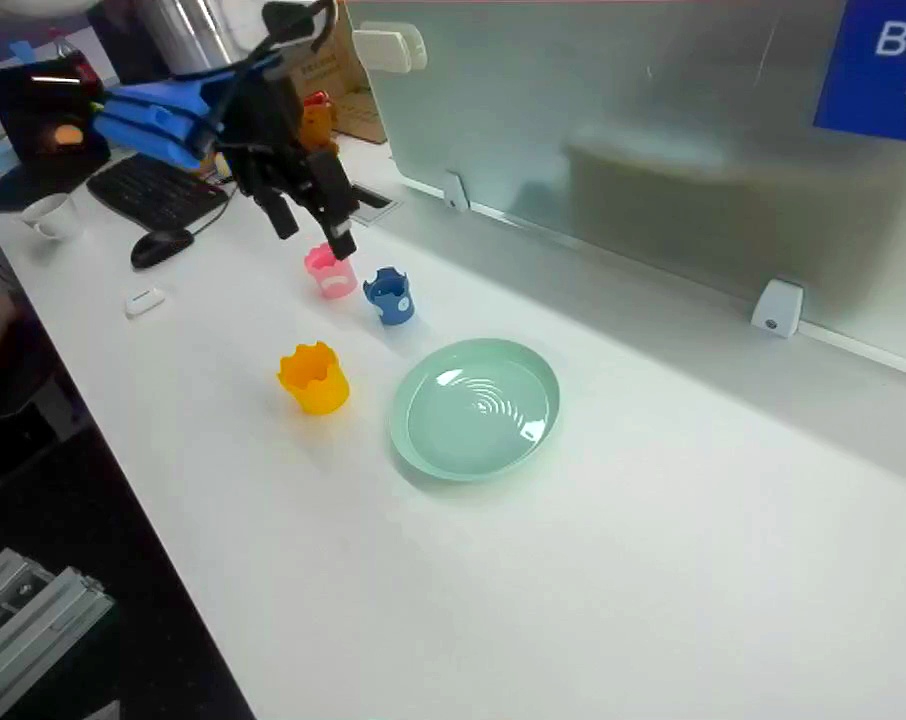}\\[1pt]
  {\scriptsize nested-cup arrangement}
\end{minipage}\hspace{0.012\linewidth}%
\begin{minipage}[t]{0.241\linewidth}
  \centering
  \includegraphics[width=\linewidth]{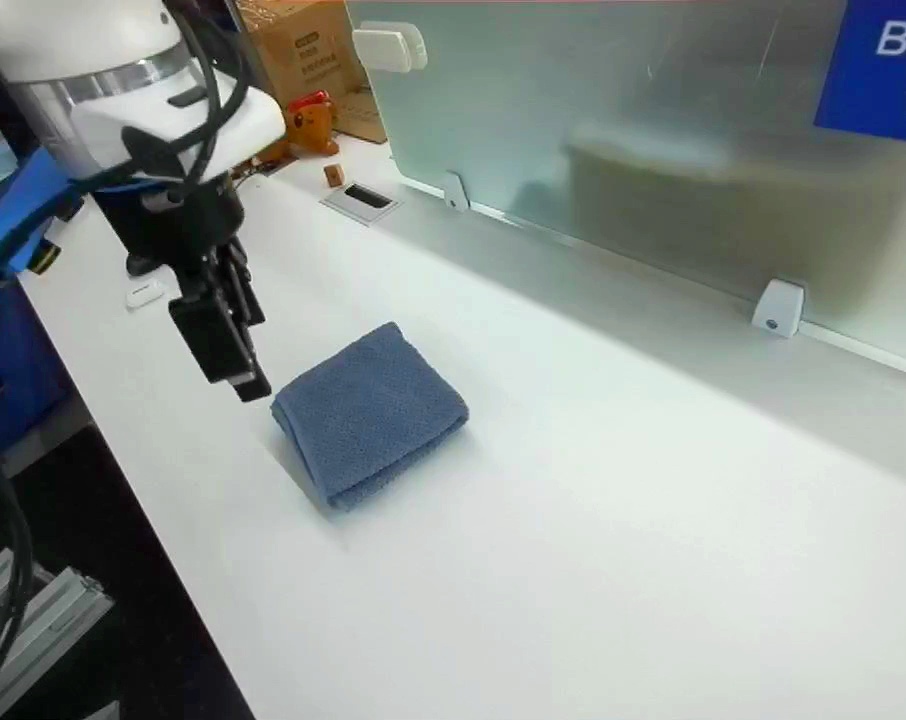}\\[1pt]
  {\scriptsize towel unfolding}
\end{minipage}

\vspace{2pt}
{\small\textbf{(e) DROID-Franka} (initial states of the four self-designed evaluation suites)}
\vspace{2pt}
\caption{\textbf{Example task observations of the evaluation benchmarks.} (a)~LIBERO \citep{libero}: MuJoCo Franka Panda tasks observed through the front camera; shown are one \emph{Spatial} and one \emph{Object} task out of the four suites. (b)~LIBERO-Plus \citep{libero-plus}: 10{,}030 variants of the LIBERO tasks generated along seven perturbation axes (camera, robot init, language, layout, background, light, noise). (c)~SimplerEnv-WidowX \citep{simplerenv}: the four BridgeData tasks \citep{walke2023bridgedata} rebuilt in SAPIEN under visual matching. (d)~DROID-Franka \citep{droid}: in-the-wild scenes of the DROID corpus on which the policies are fine-tuned. (e)~DROID-Franka: the initial states of the four self-designed suites (color-discriminated pick-and-place, snack storing and serving, nested-cup arrangement, towel unfolding) on which the fine-tuned policies are evaluated zero-shot; task progress is reported in Figure~\ref{fig:droid}. Images in (a)--(c) are reproduced from the official benchmark repositories.}
\label{fig:benchmarks}
\end{figure}

LIBERO, LIBERO-Plus, and SimplerEnv-WidowX score every rollout by binary task success from fixed initial states; DROID-Franka scores continuous \emph{task progress}. Figure~\ref{fig:benchmarks} shows example task observations.

\paragraph{LIBERO.}
LIBERO \citep{libero} is a MuJoCo benchmark built around a Franka Emika Panda arm observed through front and wrist cameras with proprioceptive state and a natural-language instruction. We use its 40 tasks in four suites that each isolate one type of distribution shift: \emph{LIBERO-Spatial} (10 tasks) varies the spatial arrangement of a fixed object set, \emph{LIBERO-Object} (10 tasks) replaces the object categories, \emph{LIBERO-Goal} (10 tasks) keeps the scene fixed while changing the goal, and \emph{LIBERO-Long} (10 tasks) composes these shifts into long-horizon tasks. Each task ships with 50 teleoperated demonstrations.

\paragraph{LIBERO-Plus.}
LIBERO-Plus \citep{libero-plus} automatically generates 10{,}030 perturbed variants of the LIBERO tasks along seven controlled axes: object layout with distractors, camera viewpoint, robot initial state, language instruction, light condition, background texture, and sensor noise, with severity progressing from mild to extreme within each axis. It is a drop-in replacement for LIBERO, so the same checkpoints are re-evaluated without modification; following the official protocol, each variant is rolled out from its prescribed initial state.

\paragraph{SimplerEnv-WidowX.}
SimplerEnv \citep{simplerenv} reconstructs the BridgeData V2 \citep{walke2023bridgedata} tabletop setup, based on a WidowX 250 arm, in SAPIEN, with lighting, textures, camera poses, and initial-state distributions tuned by system identification to match the real rig (\emph{visual matching}). We adopt its four standard BridgeData tasks: \emph{put spoon on towel}, \emph{put carrot on plate}, \emph{stack blocks}, and \emph{put eggplant in the yellow basket}. Because zero-shot outcomes are high-variance, we expand the official $24$ initial states (seeds) to $120$ per task.

\begin{figure}[t]
\centering
\scalebox{1.0}{\includegraphics[width=\linewidth]{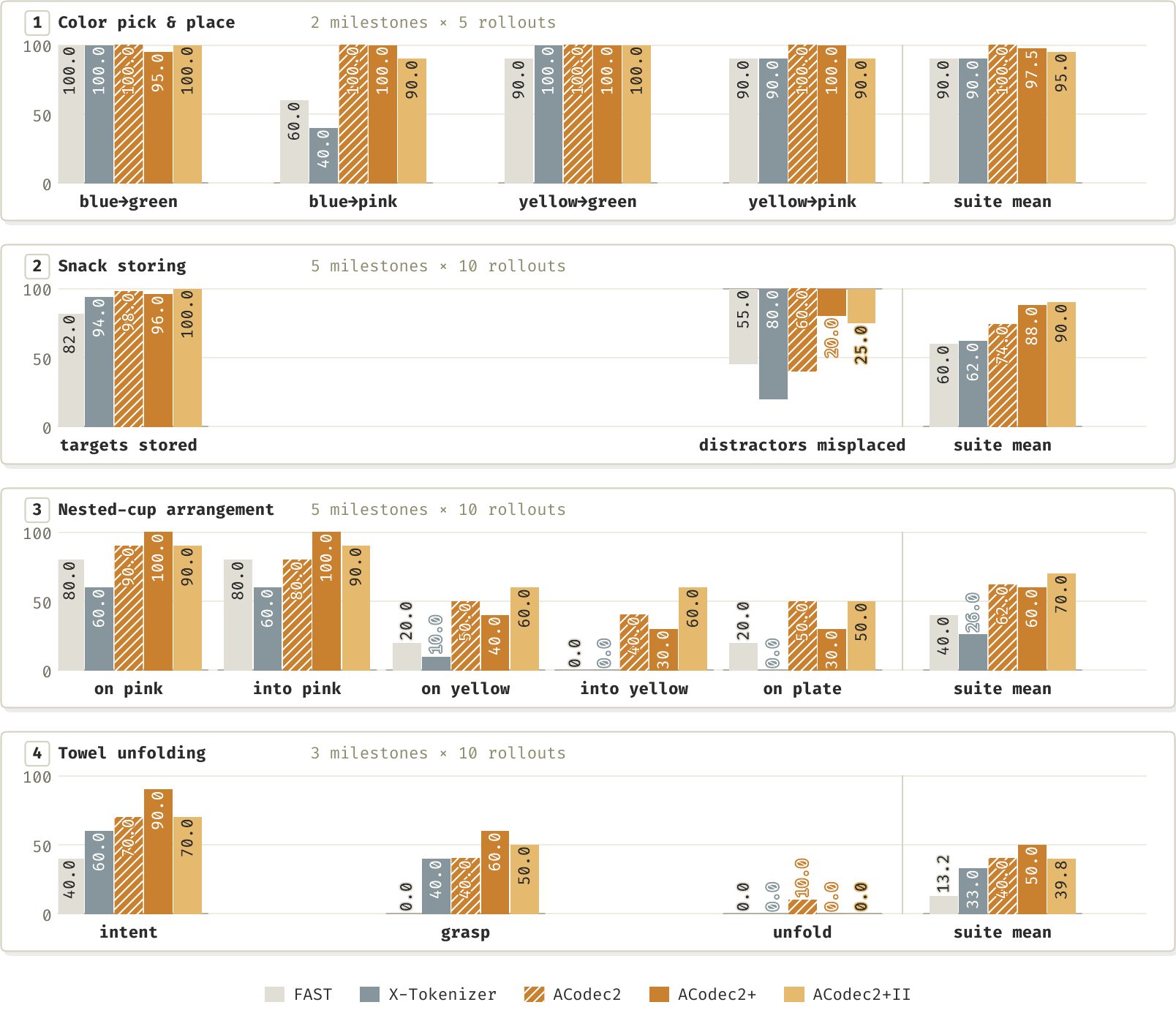}}
\caption{\textbf{DROID-Franka per-suite breakdown (\%).} The milestone-level process behind Figure~\ref{fig:droid}: task progress per instructed color pair, target items stored vs.\ distractor items misplaced (drawn from the inverted top axis; fewer is better), per-milestone nested-cup success, and towel-unfold attainment. Each row closes with the suite mean behind a divider; colors and legend match Figure~\ref{fig:droid}.}
\label{fig:droid-breakdown}
\end{figure}

\paragraph{DROID-Franka.}
In-the-wild rollouts are long-horizon and rarely end in one binary event, so each rollout is scored by \emph{task progress}: the fraction of prescribed milestones attained. A task group's score averages its rollouts (and the subtasks of the color group), and the overall score weights the four groups equally. \emph{Color-discriminated pick-and-place} uses candies and plates of two colors in four instructed combinations ($5$ rollouts each): a rollout earns $0.5$ for grasping the candy of the instructed color and the remaining $0.5$ for placing it on the plate of the instructed color. In the \emph{yellow}$\to$\emph{green} and \emph{blue}$\to$\emph{pink} combinations, a plate of the wrong color covers half of the instructed-color plate as a distractor, which makes the task harder. \emph{Snack storing} asks the robot to store five items, each worth $0.2$: one blue candy, two yellow candies, and two cola bottles. Two purple distractor candies must be left out at a penalty of $-0.2$ each, so the maximum rollout score of $1.0$ requires storing every correct item and no distractor ($10$ rollouts). \emph{Nested-cup arrangement} stacks three cups and places the stack on a plate, with five milestones of $0.2$ each: blue on pink, blue into pink, the pair on yellow, the pair into yellow, the stack onto the green plate ($10$ rollouts). \emph{Towel unfolding} carries three stages, intent, grasp, and unfold, worth $0.33$, $0.67$, and $1.0$ cumulatively ($10$ rollouts). The real-robot study compares five policies, FAST, X-Tokenizer, ACodec2, ACodec2+, and ACodec2+$^{\mathrm{II}}$, over $50$ rollouts each; Figure~\ref{fig:droid} reports the task progress of every policy on each suite, and Figure~\ref{fig:droid-breakdown} breaks each suite down to the milestone level.

\subsection{Feedback environments}
\label{app:feedback}

\begin{table}[h]
\centering
\small
\setlength{\tabcolsep}{3pt}
\caption{\textbf{Feedback under in-chunk perturbations (\%).} Random physics reports task success. For random force and motion, pick is grasp success and place is success conditional on a grasp. Each task is evaluated on 50 fixed seeds.}
\label{tab:feedback-results}
\begin{tabular}{lrrrrr}
\toprule
Setting and metric & ACodec2+ open & ACodec2+ closed & FAST & OAT-aligned & X-Tokenizer \\
\midrule
Physics: task success & 64.7 & 79.0 & 24.0 & 66.3 & 45.7 \\
Force: pick & 87.2 & 94.0 & 75.2 & 85.4 & 76.2 \\
Force: place given pick & 64.4 & 72.0 & 67.4 & 74.2 & 68.8 \\
Motion: pick & 27.6 & 46.2 & 9.2 & 18.0 & 15.8 \\
Motion: place given pick & 23.2 & 31.4 & 17.8 & 22.1 & 16.7 \\
\bottomrule
\end{tabular}
\end{table}

The three suites of Section~\ref{sec:exp-feedback} reuse the LIBERO scenes, objects, and collision models in MuJoCo. The Panda arm is driven by the \texttt{OSC\_POSE} controller with fixed impedance ($k_p=150$, damping ratio $1$); an action consists of a translation increment (at most $0.05\,\mathrm{m}$ per axis), a rotation increment (at most $0.5\,\mathrm{rad}$ per axis), and a binary gripper command, applied at $20\,\mathrm{Hz}$. Each episode begins with $10$ zero-motion steps with the gripper open, during which the perturbations already run. Physical parameters, external forces, and object motions are drawn from separate random streams seeded by the episode seed.

\paragraph{Random physics.}
Six LIBERO tasks keep their scenes, initial-layout sampling, and success predicates; at the start of each episode, the parameters in Table~\ref{tab:feedback-physics} are drawn uniformly from their ranges and held fixed for the episode. Mass changes scale the inertia proportionally. The reward is $1$ on native task success. Episodes are limited to $520$ steps for the microwave task and $300$ steps otherwise.

\begin{table}[h]
\centering
\small
\setlength{\tabcolsep}{4pt}
\caption{\textbf{Random-physics parameters.} Each parameter is drawn uniformly per episode.}
\label{tab:feedback-physics}
\begin{tabular}{llll}
\toprule
\textbf{Task (source suite)} & \textbf{Parameter} & \textbf{Easy} & \textbf{Hard} \\
\midrule
Bowl to cabinet top (Goal)         & bowl mass (kg)                        & $1.0$--$2.0$   & $2.25$--$4.5$ \\
Open middle drawer (Goal)          & slide damping (N$\cdot$s/m)           & $300$--$1000$  & $506$--$1688$ \\
                                   & slide friction (N)                    & $8$--$23$      & $13.5$--$38.8$ \\
Milk to basket (Object)            & finger-joint damping (N$\cdot$s/m)    & $3500$--$15000$ & $13125$--$56250$ \\
Mug in microwave, close door (Long) & door damping (N$\cdot$m$\cdot$s/rad) & $0.5$--$2.5$   & $0.75$--$3.75$ \\
                                   & door friction (N$\cdot$m)             & $0.05$--$0.25$ & $0.075$--$0.375$ \\
                                   & door spring (N$\cdot$m/rad)           & $0.15$--$0.4$  & $0.225$--$0.6$ \\
Wine bottle to rack (Goal)         & bottle mass (kg)                      & $0.8$--$2.0$   & $2.1$--$5.25$ \\
                                   & center-of-mass $x$ offset (m)         & $\pm0.01$      & $\pm0.01$ \\
Push plate to stove (Goal)         & sliding friction coefficient          & $0.03$--$0.55$ & $0.27$--$4.95$ \\
\bottomrule
\end{tabular}
\end{table}

\paragraph{Random force and random motion.}
Both suites use the ten LIBERO-Object tasks, in which the instructed object is placed in a basket. In random force, a translational force is applied to the gripper base in the world frame. A pulse starts every $1.0\,\mathrm{s}$, the first $0.10$--$0.20\,\mathrm{s}$ after reset; each pulse has a magnitude of $150$--$180\,\mathrm{N}$, a direction uniform on the sphere, and a duration of $0.08$--$0.14\,\mathrm{s}$, all resampled per pulse. In random motion, the initial planar positions of all objects and the basket are sampled jointly under reachability, clearance, and camera-visibility constraints. Each object and the basket then follows its own planar velocity command with speed $0.18$--$0.30\,\mathrm{m/s}$; the heading turns by up to $\pm60^{\circ}$ every $0.25$--$0.45\,\mathrm{s}$ and is reflected at the workspace boundary. The command is tracked by a planar force with a $0.15\,\mathrm{s}$ response time and an acceleration limit of $1.5\,\mathrm{m/s^2}$. An object stops being driven once it is grasped or lifted off the table. In both suites, the first two-finger grasp of the target earns $+1$ and the first native in-basket success earns $+1$ and ends the episode; episodes are limited to $280$ steps.

\paragraph{Policies and training.}
All policies are G0.5 models fine-tuned on LIBERO with the corresponding tokenizer (Appendix~\ref{app:tokenizers}). They generate $20$-step chunks with the first $10$ executed and receive camera images only at chunk boundaries. For ACodec2+ / closed-loop, the measured end-effector pose and gripper finger position after each executed action are normalized, projected to an embedding, and appended to the context before the next action token is predicted; in random motion, the state also contains the object positions. In random physics and random force, every token is confined to one control step, so a state is inserted after every action. In random motion, a token may span neighboring steps; all actions it decodes are executed, and then one state per executed action is inserted. The closed-loop starting checkpoint is obtained by one additional epoch of supervised fine-tuning on the LIBERO demonstrations with interleaved states; the open-loop variant and the baselines start from their standard LIBERO checkpoints. PPO uses $320$ rollouts per iteration, learning rate $5{\times}10^{-6}$, $\gamma=0.99$, GAE $\lambda=0.95$, clip ratio $0.2$, and entropy bonus $0.005$ for every policy. Random physics trains for $200$ iterations on the easy ranges and then $200$ on the hard ranges; random force trains for $150$ iterations and random motion for $200$. Checkpoints are evaluated every $50$ iterations with greedy decoding on $50$ fixed seeds per task; seeds, initial simulator states, and physical parameters are shared by all policies.

\section{Agent control details}
\label{app:agent-details}

This appendix collects the two tables behind the agent-control study of Section~\ref{sec:exp-agent}: the macro library the VLM induced on RMBench (Table~\ref{tab:rmbench-macros}) and the nine first-time milestones of the Multi-Drawer case study (Table~\ref{tab:drawer-milestones}), followed by a description of the milestones.

\begin{table}[t]
\centering
\small
\renewcommand{\arraystretch}{1.3}
\setlength{\tabcolsep}{5pt}
\caption{\textbf{Macros induced by the VLM on RMBench} (the library carried into Figure~\ref{fig:rmbench}\emph{c}). Names and semantics are the model's own words, not post-hoc human summaries.}
\label{tab:rmbench-macros}
\begin{tabular}{@{}>{\ttfamily\scriptsize\raggedright}p{0.30\textwidth}>{\raggedright\arraybackslash}p{0.45\textwidth}cr@{}}
\toprule
Macro & Semantics (VLM-authored) & Uses & Duration\,(s) \\
\midrule
grasp\_from\_row\_hover & From an aligned hover 10\,cm above grasp height, descend open, close, and test-lift. & 26 & 1.40 \\
park\_cube\_in\_front & With a verified cube lifted 4\,cm, buffer 15\,cm nearward, release, and withdraw. & 20 & 2.40 \\
set\_down\_and\_clear & With a held cube 4\,cm above placement height, lower, release, and withdraw 10\,cm. & 33 & 1.40 \\
buffer\_to\_row\_left\_hover & From a matching front buffer hover, approach the neighbouring row slot 9\,cm left and 15\,cm forward. & 2 & 1.60 \\
row\_to\_front\_buffer\_hover & From a row-slot hover, move open 15\,cm nearward to its front buffer. & 2 & 1.00 \\
buffer\_to\_row\_left\_and\_place & With a verified buffer cube lifted 4\,cm, add clearance, transfer 9\,cm left and 15\,cm forward, place, and withdraw. & 2 & 3.80 \\
adjacent\_cube\_right\_and\_place & With a verified cube lifted 4\,cm, transfer 9\,cm right into an empty slot, place, and withdraw. & 3 & 2.00 \\
descend\_open\_to\_cube & With downward fingers above a cube, descend open 7\,cm for alignment inspection. & 2 & 0.47 \\
orient\_down & Pitch about $-88^{\circ}$ open; only from a verified compatible orientation with clearance. & 7 & 2.27 \\
carry\_slot\_left & Move base $-x$ 10\,cm closed; require verified grasp, spacing and clearance. & 15 & 0.67 \\
carry\_slot\_right & Move base $+x$ 10\,cm closed; require verified grasp, spacing and clearance. & 13 & 0.67 \\
button\_press & Close, command 8\,cm descent and hold; require cap alignment, flat fingertips and both grippers previously opened. & 12 & 1.33 \\
carry\_forward\_12cm & Move base $+y$ 12\,cm closed; require verified grasp and clear route. & 14 & 0.80 \\
\bottomrule
\end{tabular}
\end{table}

\begin{table}[t]
\centering
\footnotesize
\renewcommand{\arraystretch}{1.25}
\setlength{\tabcolsep}{4pt}
\caption{\textbf{The nine first-time milestones of Figure~\ref{fig:agent-control}, in the model's own words.} Thinking is quoted (abridged) from the decision logs and end-of-episode reflections; macros are those born at, or executed during, the milestone (the library is append-only).}
\label{tab:drawer-milestones}
\begin{tabular}{@{}c>{\raggedright\arraybackslash}p{0.125\textwidth}>{\raggedright\arraybackslash\itshape}p{0.49\textwidth}>{\ttfamily\scriptsize\raggedright\arraybackslash}p{0.28\textwidth}@{}}
\toprule
Ep. & Milestone & Model's thinking & Macros \\
\midrule
2 & First drawer opened & ``Three closes have missed. The wrist view shows the fingers open left--right, while the handle bar also runs left--right, parallel to the finger axis, so the fingers can never pinch it. I'll roll the tool about $90^{\circ}$ so the fingers open vertically and can straddle the horizontal bar.'' & roll\_plus\_90\_open, rolled\_drawer\_grip\_pull \\
3 & Spotted the tongs & ``The left wrist view looks into the open top drawer of the left cabinet and shows only silver tongs, not the measuring cup.'' & open\_drawer\_from\_handle, tool\_axis\_advance\_pitched \\
5 & Found the measuring cup & ``The head view now shows the measuring cup in the open right-cabinet drawer: a white round bowl with a handle.'' & slow\_pull\_drawer\_10cm, pull\_drawer\_by\_side\_wall \\
8 & First cup grasp & ``The close stalled at 0.113, which suggests the thin tab is pinching again. \ldots{} this time I'm holding closed for 16 rows to build full grip force, then lifting very slowly.'' & pitched\_yaw\_center\_pinch, close\_lift\_slow\_2 \\
11 & Invented the playbook & ``Lessons were facts, not procedures. Each decision re-derived heights, drawer level, and the escape order. \ldots{} I kept guessing.'' (reflection) & drawer\_open\_rolled\_full, grasp\_close\_verify\_lift, carry\_pitchup\_back\_10, release\_open\_rise \\
13 & First cup into the bin & ``The cup is in the bin: the right wrist view shows it resting inside, and the instruction has advanced to the tongs.'' & carry\_pitchup\_back\_10, release\_open\_rise, table\_pinch\_vertical \\
16 & Devised the handover & ``When the target's drawer is on the far side from the bin, set the object down near the table centre early and let the bin-side arm fetch it.'' (hypothesis) & left\_carry\_minus\_y\_low, pitched\_hook\_handle\_extend \\
19 & First tongs grasp & ``The lift worked: the wrist moved back 3\,cm and up 2.6\,cm, and the reading holds at 0.232. The wrist view shows the tongs still clamped between the fingers.'' & grasp\_close\_verify\_lift, pitch\_up\_13\_closed \\
20 & First full success & ``The close stalled at 0.204 and the 4.6\,cm lift completed, so the tongs are grasped.'' & steep\_center\_pinch, bin\_lower\_release\_rise \\
\bottomrule
\end{tabular}
\end{table}

\subsection{Milestones of the Multi-Drawer case study}
\label{sec:drawer-milestones}

Table~\ref{tab:drawer-milestones} quotes the model at the nine first-time achievements marked in Figure~\ref{fig:agent-control}, with the macros created or executed there. The paragraphs below describe what each period contributed.

\textbf{Episodes 1--5.} Horizontal handle bars can be pinched only by vertically opening fingers, which requires a $+90^{\circ}$ wrist roll, and every later drawer interaction uses this grip. This period also fixed how gripper readings are interpreted: a full close near $0.004$ means the fingers closed on nothing, and $0.2$--$0.4$ means the gripper holds an object. The first positive score was a false trigger: the bin had been pushed under an open drawer, so the scorer registered the cup as delivered. The rules kept from this period are to never move the bin and to rely on the instruction line rather than the image.

\textbf{Episodes 6--10.} Reliable motions became macros: slow $1$\,cm-step pulls (fast pulls slip off), the handle standoff, and, in episode~8, the first cup grasp. Transporting the cup then caught on the drawer handle, which gave the rule to rise before carrying. Three protocol errors were found in the same period: a rejected IK solution for one arm voids the entire reply, a hold token issued to an open gripper drops the object, and forty decisions were spent grasping a peeler that the camera had labeled a cup. Each became a standing rule. The operator addendum at the end of this period gives the agent the layout ground truth.

\textbf{Episodes 11--14.} The reflection after episode~11 diagnosed the failure as ``lessons were facts, not procedures'': every decision re-derived heights, drawer level, and escape order. The agent wrote a phase playbook with named phases, phase macros, and a table from observations to actions. From episode~12 the cup chain runs on the playbook, and episode~13 completes the first cup delivery. When the left arm reached its reach limit with the tongs, it put them on the table and the right arm picked them up with a vertical pinch, the first left-to-right handover.

\textbf{Episodes 15--20.} Episode~15, the worst of the series, misidentified both targets and fixed two rules: the layout is ground truth, and the right arm uses the mirrored $-90^{\circ}$ clamp. Episode~16 introduced viewing each drawer at level height first, which removed the closes on empty space, and mapped the carry limit of the left arm. This gave the final strategy: put the object near the table center early and let the bin-side arm fetch it. Episode~19, with the decision budget doubled, achieved the first in-drawer tongs grasp. In episode~20 the agent pushed the overhanging drawer shut, which removed the interference and made the steep center pinch on the tongs reliable, and then released over the bin center, the first environment-confirmed success.

\textbf{Findings.} First, lessons improved performance only when written as procedures (the playbook) or as named token sequences (macros); the same facts written as prose did not, and the library is append-only. Second, most permanent rules concern interpretation, such as what a gripper reading means, what an image shows, and what invalidates a reply, rather than motion. Third, the final handover strategy came from the agent mapping its own kinematic limits, and it used primitives that were available from the first episodes.

\section{The second-order scheme on a group}
\label{sec:second-general}

\proofhere

At first order the feasible successors $\mathcal F(q)=q\mathcal A$ carry the same relative motions to every configuration, and Theorem~\ref{thm:composition-optimal} covers them in closed form.  Acceleration limits and workspace boundaries remove this property: the feasible next increment depends on the current one and on the distance to the boundary, state that the first-order error recursion cancels.  Folding every truncation into one enlarged $\mathcal A$ covers increments that are never realized, such as a one-step reversal from full speed, and increases the count.  The second-order variant of Section~\ref{sec:fitting} instead lets the decoder track the increment, so that a primitive names a bounded change of increment, a smaller set to cover.  This appendix has three parts.
\begin{enumerate}
\item Section~\ref{sec:second-criterion} defines the decoder, whose reconstruction obeys its own copy of the real limits widened by a margin, and replaces the recoverability condition, which no primitive set meets here, by an exact test and the optimum $K_2^\star$ it defines.
\item Section~\ref{sec:second-margin} shows why the margin is needed: without it a velocity error cannot decrease, while a margin of a few $\epsilon$ admits an explicit primitive grid on translation and on $\SO(3)$.  These are the grids the fit uses.
\item Section~\ref{sec:second-compute} computes $K_2^\star$ exactly on translation and explains why the search does not transfer to $\SO(3)$.
\end{enumerate}

\subsection{The decoder and the exact criterion}
\label{sec:second-criterion}

The configuration becomes the pair $(q,\delta)\in G\times G$ of position and the increment that produced it, $\delta_t=q_{t-1}^{-1}q_t$, so that a step first updates the increment by the change $\gamma_t=\delta_t^{-1}\delta_{t+1}$ and then applies it,
\begin{equation}
(q,\delta)\mapsto(q\,\delta\,\gamma,\ \delta\,\gamma).
\label{eq:second-general-step}
\end{equation}
An instance fixes closed sets $\mathcal W,\mathcal D,\mathcal C\subseteq G$, and a trajectory is admissible while $q_t\in\mathcal W$, $\delta_t\in\mathcal D$, $\gamma_t\in\mathcal C$.  A state can obey all three and still have no infinite admissible continuation, so we use the viable states $\mathcal V_2=\{(q,\delta):\ q\delta^{-1}\in\mathcal W\ \text{and an infinite admissible trajectory starts at }(q,\delta)\}$, the first clause asking that the step producing $q$ also started in the workspace, and their feasible changes $\mathcal F(q,\delta)=\{\gamma\in\mathcal C:\ (q\delta\gamma,\delta\gamma)\in\mathcal V_2\}$, nonempty at every viable state.

At first order the reconstruction is not required to be a configuration the robot could reach.  At second order its increment $\hat\delta$ is the rate the decoded stream commands, so we must say which limits it obeys.  We let it run its own instance, nested around the real one: closed $\mathcal W\subseteq\mathcal W^\mu$, $\mathcal D\subseteq\mathcal D^\mu$, $\mathcal C\subseteq\mathcal C^\mu$, with viable states $\mathcal V_2^\mu$ and feasible changes $\mathcal F^\mu$.  The \emph{margin} $\mu\ge0$ has one component per relaxed bound, and the \emph{strict} case $\mu=0$ holds the reconstruction to the robot's own limits.  The tolerance $\mathcal E$ is never enlarged, so a margin changes what the reconstruction may command and not how accurate it must be.

A primitive $u\in\mathcal U\subseteq G$ is a fixed request for the next change.  It is made without regard to the reconstructed state, so it is clipped to the nearest feasible value $\bar u=\Pi_{\mathcal F^\mu(\hat q,\hat\delta)}(u)$ and executed as
\begin{equation}
\hat\delta_{t+1}=\hat\delta_t\bar u_t,
\qquad
\hat q_{t+1}=\hat q_t\hat\delta_{t+1}.
\label{eq:second-general-recon}
\end{equation}
We assume $\mathcal F^\mu$ is compact and nonempty on $\mathcal V_2^\mu$ and fix one nearest-point selection; both instances below satisfy this.  Since $\bar u_t\in\mathcal F^\mu(\hat q_t,\hat\delta_t)$, the reconstruction never leaves $\mathcal V_2^\mu$.  With the errors $\xi_t=\hat q_t^{-1}q_t$ and $\psi_t=\hat\delta_t^{-1}\delta_t$, one step gives
\begin{equation}
\psi_{t+1}=\bar u_t^{-1}\,\psi_t\,\gamma_t,
\qquad
\xi_{t+1}=\bar u_t^{-1}\,c_{\hat\delta_t}(\xi_t)\,\psi_t\,\gamma_t,
\qquad
c_h(\xi)=h^{-1}\xi h,
\label{eq:second-general-error}
\end{equation}
the first-order step $\xi_{t+1}=u^{-1}\xi_t\delta_t$ conjugated into the frame of the current increment.  The tolerance applies to $\xi$ alone, and the accepted pairs are $\mathcal S_2=\{(q,\delta;\hat q,\hat\delta):\ (q,\delta)\in\mathcal V_2,\ (\hat q,\hat\delta)\in\mathcal V_2^\mu,\ \hat q^{-1}q\in\mathcal E\}$.

Translation takes $G=(\R,+)$ with $\mathcal W=[\ell,h]$, $\mathcal D=[-b,b]$, $\mathcal C=[-c,c]$, $\mathcal E=[-\epsilon,\epsilon]$.  Writing the pair as $(x,v)$ and the change as $g_t$, \eqref{eq:second-general-error} is plain accumulation,
\begin{equation}
\psi_{t+1}=\psi_t+g_t-\bar u_t,
\qquad
\xi_{t+1}=\xi_t+\psi_{t+1},
\label{eq:second-translation-error}
\end{equation}
so quantizing each change separately does not control the position.  For $c>0$ the extra distance needed to bring a positive displacement $v$ to rest under $|g|\le c$ is $B(v)=\sum_{i=1}^{\lceil v/c\rceil-1}(v-ic)$, with $B(v)=0$ for $v\le0$, and
\begin{equation}
\mathcal V_2=\bigl\{(x,v):\ |v|\le b,\quad x-v\in[\ell,h],\quad x+B(v)\le h,\quad x-B(-v)\ge\ell\bigr\}.
\label{eq:second-translation-viability}
\end{equation}
The margin has three components, replacing $[\ell,h]$, $b$, $c$ by $[\ell-\mu_x,h+\mu_x]$, $b+\mu_v$, $c+\mu_g$.  Orientation takes $G=\SO(3)$ with $\mathcal W=\SO(3)$, $\mathcal D=\mathcal B_R(b_R)$, $\mathcal C=\mathcal B_R(c_R)$, $\mathcal E=\mathcal B_R(\epsilon_R)$ in the notation of Corollary~\ref{prop:so3-exact}.  Without a boundary there is no braking constraint, since repeating $\gamma=I_3$ continues forever, so $\mathcal V_2=\SO(3)\times\mathcal B_R(b_R)$ and $\mathcal F(q,\delta)=\{\gamma\in\mathcal B_R(c_R):\ \theta(\delta\gamma)\le b_R\}$.  The margin has two components, one on the increment angle and one on the change angle.

At first order the optimum is defined by the recoverability condition \eqref{eq:full-tolerance}, a symbol for every accepted pair.  On $\mathcal S_2$ no primitive set satisfies it.  $\mathcal S_2$ tests $\xi$ alone, so it contains pairs whose $\psi$ is as large as the increment sets allow, while one symbol moves $\hat\delta$ by a bounded amount.  On translation a pair with $\xi=\epsilon$ and $\psi=2b$ leaves $\mathcal E$ for every symbol once $2b>2c+\mu_g$, by \eqref{eq:second-translation-error}, so the requirement rejects every primitive set, even those that never reach such a pair from an exact start.  The guarantee needs only a set of accepted pairs that contains the exact starts and that the decoder can always remain inside.  Proposition~\ref{prop:viability} shows that this condition is exact and obtains the largest such set, the viability kernel of the accepted pairs, as the limit of a decreasing recursion.  It holds for the general decoder \eqref{eq:general-step} on any $G$, and the recoverability condition is its condition (3) with $\mathcal R=\mathcal S$.

\proofhere
\begin{proposition}[Invariant-set criterion]
\label{prop:viability}
Assume every configuration has an infinite feasible continuation.

\emph{(a) The criterion.}  The symbol $z_t$ can be selected at every step from $q_{0:t+1}$ so that the decoder \eqref{eq:general-step} keeps $(q_t,\hat q_t)\in\mathcal S$ on every infinite feasible trajectory, from every exact start, if and only if there is a set $\mathcal R$ of pairs with (1) $\mathcal R\subseteq\mathcal S$; (2) $(q,q)\in\mathcal R$ for every $q$; (3) $\forall(q,\hat q)\in\mathcal R,\ \forall q^+\in\mathcal F(q),\ \exists k\le K:\bigl(q^+,y_k(\hat q)\bigr)\in\mathcal R$.  If such a selection exists, one that uses only $(q_t,\hat q_t,q_{t+1})$ also exists.

\emph{(b) The kernel.}  Put $\mathcal Q_0=\mathcal S$ and
\begin{equation}
\mathcal Q_{n+1}=\bigl\{(q,\hat q)\in\mathcal Q_n:\ \forall q^+\in\mathcal F(q)\ \exists k\le K\ \text{with}\ \bigl(q^+,y_k(\hat q)\bigr)\in\mathcal Q_n\bigr\},
\qquad
\mathcal Q_\infty=\bigcap_n\mathcal Q_n .
\label{eq:viability-kernel}
\end{equation}
Then $\mathcal Q_n$ is exactly the set of pairs from which the symbols can be selected so that the pair stays accepted for $n$ more steps; every $\mathcal R$ obeying (1) and (3) lies in $\mathcal Q_\infty$; and when $K$ is finite $\mathcal Q_\infty$ itself obeys (1) and (3).  A largest admissible $\mathcal R$ therefore exists and equals $\mathcal Q_\infty$, and the criterion of (a) is $\{(q,q)\}\subseteq\mathcal Q_\infty$.
\end{proposition}

\begin{proof}
\textbf{(a) Sufficiency.}
Repeat the proof of Lemma~\ref{lem:onestep} with $\mathcal R$ in place of $\mathcal S$: requirement 2 starts the induction at $(q_0,q_0)$, requirement 3 continues it, and requirement 1 returns the conclusion to $\mathcal S$.  The symbol chosen there depends on $(q,\hat q)$ and $q^+$ and not on $t$, which is what makes the policy stationary.

\textbf{(a) Necessity.}
Suppose the decoder succeeds from every exact start, and let $\mathcal R$ be the set of pairs $(q_t,\hat q_t)$ it reaches over all exact starts and all infinite feasible trajectories.  Requirement 1 holds because success means every reached pair lies in $\mathcal S$, and requirement 2 because the exact start at $q$ reaches $(q,q)$ at $t=0$.  For requirement 3, take $(q,\hat q)\in\mathcal R$, reached at some time $t$ along a feasible history $q_{0:t}$ with $q_t=q$, and take any $q^+\in\mathcal F(q)$.  Causality is what makes $q^+$ free to choose here: $\pi_s$ reads only $q_{0:s+1}$, so $z_{0:t-1}$ is determined by $q_{0:t}$ and hence so is $\hat q_t=\hat q$.  Extend $q_{0:t}q^+$ by an infinite feasible continuation, which exists by assumption.  The result is an infinite feasible trajectory agreeing with the history on $q_{0:t}$, so the encoder reaches the same $\hat q_t=\hat q$ and then emits some $z_t=k$ after observing $q^+$; the decoder moves to $y_k(\hat q)$.  The pair $\bigl(q^+,y_k(\hat q)\bigr)$ is therefore reached and lies in $\mathcal R$.

\textbf{(b).}
That $\mathcal Q_n$ is the $n$-step survivor set is induction on $n$: $\mathcal Q_0=\mathcal S$, and a pair permits $n+1$ safe steps exactly when it is safe now and, after every feasible successor, some symbol reaches a pair permitting $n$ more, which is \eqref{eq:viability-kernel}, the rule observing $q^+$ before choosing the symbol.  If $\mathcal R$ obeys (1) and (3) then $\mathcal R\subseteq\mathcal Q_0$, and $\mathcal R\subseteq\mathcal Q_n$ gives $\mathcal R\subseteq\mathcal Q_{n+1}$ by (3), so $\mathcal R\subseteq\mathcal Q_\infty$.  For the converse, $\mathcal Q_\infty\subseteq\mathcal Q_0=\mathcal S$ is (1); for (3) fix $(q,\hat q)\in\mathcal Q_\infty$ and $q^+\in\mathcal F(q)$, so each membership $(q,\hat q)\in\mathcal Q_{n+1}$ supplies a symbol $k_n$ whose successor lies in $\mathcal Q_n$.  With $K$ finite one symbol $k^\star$ recurs for infinitely many $n$, and the sets decrease, so its successor lies in every $\mathcal Q_m$.
\end{proof}

On $\mathcal S_2$ the recursion reads as follows.  Write $y^+(\gamma,u)$ for the pair reached from $y$ when the real system takes $\gamma\in\mathcal F(q,\delta)$ and the decoder, having observed $\gamma$, takes $u\in\mathcal U$.  Starting from $Q_0(\mathcal U)=\mathcal S_2$,
\begin{equation}
Q_{n+1}(\mathcal U)=\bigl\{y\in Q_n(\mathcal U):\ \forall\gamma\in\mathcal F(q,\delta)\ \exists u\in\mathcal U\ \text{with }y^+(\gamma,u)\in Q_n(\mathcal U)\bigr\},
\qquad
Q_\infty(\mathcal U)=\bigcap_n Q_n(\mathcal U),
\label{eq:second-general-kernel}
\end{equation}
and exact initialization is $\mathcal I=\{(q,\delta;\,q,\delta):(q,\delta)\in\mathcal V_2\}$.

\begin{theorem}[Exact test for a fixed second-order primitive set]
\label{thm:second-general-test}
Let $\mathcal U\subseteq G$ be finite.  Then (1) $Q_n(\mathcal U)$ is exactly the set of pairs from which the symbols can be selected, in the sense of Lemma~\ref{lem:onestep}, so that the pair stays in $\mathcal S_2$ for the next $n$ steps; (2) every $y\in Q_\infty(\mathcal U)$ has, for each real $\gamma\in\mathcal F(q,\delta)$, some $u\in\mathcal U$ with $y^+(\gamma,u)\in Q_\infty(\mathcal U)$; and (3) $\mathcal U$ tracks every infinite admissible trajectory from exact initialization, with $\xi_t\in\mathcal E$ for all $t$, if and only if
\begin{equation}
\mathcal I\subseteq Q_\infty(\mathcal U).
\label{eq:second-general-condition}
\end{equation}
\end{theorem}

\begin{proof}
This is Proposition~\ref{prop:viability} on the pair state space.  The real configuration is $(q,\delta)$ and the decoder \eqref{eq:second-general-recon} is \eqref{eq:general-step} on $(\hat q,\hat\delta)$ with the finite symbol set $\mathcal U$; $\mathcal S_2$ plays the role of $\mathcal S$, $\mathcal F(q,\delta)$ that of $\mathcal F(q)$, and \eqref{eq:second-general-kernel} is then literally \eqref{eq:viability-kernel}.  Restricting the two state spaces to $\mathcal V_2$ and $\mathcal V_2^\mu$ is how that proposition's infinite-continuation assumption is met, and its exact starts are $\mathcal I$.  Parts (1) and (2) are Proposition~\ref{prop:viability}(b), the finiteness of $\mathcal U$ being what (2) needs; part (3) is its criterion $\mathcal I\subseteq Q_\infty(\mathcal U)$ together with (a).
\end{proof}

The optimum is therefore
\begin{equation}
K_2^\star(\mu)=\min\bigl\{|\mathcal U|:\ \mathcal U\subseteq G\text{ finite},\ \mathcal I\subseteq Q_\infty(\mathcal U)\bigr\},
\label{eq:second-general-optimum}
\end{equation}
$+\infty$ if no finite set works.

\subsection{The margin and explicit primitive grids}
\label{sec:second-margin}

\begin{lemma}[Velocity error without a margin]
\label{lem:strict-lag}
Take the translational instance with $\mu=0$ and $c>0$.  Suppose the displacement error satisfies $\psi_t\ge\rho>0$ at some time $t$, and the real trajectory takes $g_s=c$ for $s=t,\ldots,t+n-1$.  Then $\psi_s\ge\rho$ for every $s=t,\ldots,t+n$, and
\begin{equation}
\xi_{t+n}\ \ge\ \xi_t+n\rho .
\label{eq:strict-lag}
\end{equation}
The same holds with every sign reversed when $\psi_t\le-\rho$ and $g_s=-c$.
\end{lemma}

\begin{proof}
At $\mu=0$ the reconstruction's feasible set satisfies $\mathcal F(\hat x,\hat v)\subseteq\mathcal C=[-c,c]$, so every executed value obeys $\bar u_s\le c$, for every primitive set.  With $g_s=c$, the first half of \eqref{eq:second-translation-error} gives $\psi_{s+1}=\psi_s+c-\bar u_s\ge\psi_s$, and induction from $\psi_t\ge\rho$ gives $\psi_s\ge\rho$ for all $s$ up to $t+n$.  Summing the second half over $s=t,\ldots,t+n-1$ gives $\xi_{t+n}=\xi_t+\sum_{s=t+1}^{t+n}\psi_s\ge\xi_t+n\rho$.
\end{proof}

Since $\mathcal E$ bounds $\xi$, \eqref{eq:strict-lag} caps $n\rho$: the longer the robot can keep its velocity change at $c$, the smaller the velocity error a strict decoder may ever have, and its primitives must be fine enough to ensure this.  In particular no box $\lvert\xi\rvert\le\epsilon_\xi$, $\lvert\psi\rvert\le\epsilon_\psi$ with $\epsilon_\psi>0$ is invariant in the strict case, since its corner $\xi=\epsilon_\xi$, $\psi=\epsilon_\psi$ away from the workspace boundary gives $\xi_{t+1}\ge\epsilon_\xi+\epsilon_\psi$ under $g_t=c$.  With a margin $\mu_g>0$ an executed value may exceed $c$, so $\psi$ can strictly decrease.  On both instances the sufficient margin is explicit.  In each, the encoder sends the primitive nearest to the request $W_t=\hat\delta_t^{-1}\hat q_t^{-1}q_{t+1}$, the change that would land the reconstruction exactly on the next configuration; on translation $W_t=\xi_t+\psi_t+g_t$.

\begin{proposition}[Explicit grids under a margin]
\label{prop:margin-witness}
\emph{(a) Translation.}  Let $\mu^\dagger=(\epsilon,2\epsilon,2\epsilon)$ and $\mu\ge\mu^\dagger$ componentwise.  Place $K=\lceil c/\epsilon\rceil+3$ points uniformly on $[-c-2\epsilon,\,c+2\epsilon]$, endpoints included.  Then $\lvert\xi_t\rvert\le\epsilon$ for every $t$ from exact initialization, so $\mathcal I\subseteq Q_\infty(\mathcal U)$ and $K_2^\star(\mu)\le\lceil c/\epsilon\rceil+3$.

\emph{(b) Orientation.}  Let the margin widen the increment angle by at least $2\epsilon_R$ and the change angle by at least $3\epsilon_R$, put $\rho=\min\{\pi,c_R+3\epsilon_R\}$, and let $\mathcal U\subseteq\mathcal B_R(\rho)$ satisfy $\mathcal B_R(\rho)\subseteq\bigcup_{u\in\mathcal U}u\,\mathcal B_R(\epsilon_R)$.  Then $\theta(\xi_t)\le\epsilon_R$ for every $t$ from exact initialization, so $\mathcal I\subseteq Q_\infty(\mathcal U)$, and $K_2^\star(\mu)$ is at most the right-hand side of \eqref{eq:so3-bounds} with $\rho_R=\rho$.
\end{proposition}

\begin{proof}
Both parts prove by induction the joint invariant that $\xi_t$ is within $\epsilon$ of the identity and $\psi_t$ within $2\epsilon$, which exact initialization starts, and both show that the chosen primitive is feasible, so nothing is clipped.

\emph{(a).}  The spacing is at most $2\epsilon$, since $(2c+4\epsilon)/(\lceil c/\epsilon\rceil+2)\le2\epsilon$ is $c\le\epsilon\lceil c/\epsilon\rceil$.  The invariant puts $W_t=\xi_t+\psi_t+g_t$ in $[-c-3\epsilon,\,c+3\epsilon]$, and the grid spans $[-c-2\epsilon,c+2\epsilon]$ with spacing at most $2\epsilon$, so the nearest point $u_t$ leaves $\xi^\circ:=W_t-u_t$ with $\lvert\xi^\circ\rvert\le\epsilon$, the error the step leaves if $u_t$ is executed unclipped.  Write $\hat x^\circ,\hat v^\circ$ for the reconstructed state that step produces, so $x_{t+1}-\hat x^\circ=\xi^\circ$ and $v_{t+1}-\hat v^\circ=\xi^\circ-\xi_t$.  Then $\lvert x_{t+1}-\hat x^\circ\rvert\le\epsilon$, $\lvert v_{t+1}-\hat v^\circ\rvert\le2\epsilon$ and $\lvert u_t\rvert\le c+2\epsilon$, which put $\hat x^\circ$ in $[\ell-\epsilon,h+\epsilon]$, $\lvert\hat v^\circ\rvert$ below $b+2\epsilon$ and $\hat x^\circ-\hat v^\circ=\hat x_t$ in the enlarged workspace.  The remaining constraint is braking, where the three components of $\mu^\dagger$ act together: enlarging the change bound from $c$ to $c+2\epsilon$ and the displacement by $2\epsilon$ shortens the stopping distance, because
\[
B_{c+2\epsilon}(v+2\epsilon)=\sum_{i\ge1}\bigl(v-ic-(i-1)2\epsilon\bigr)_+\ \le\ \sum_{i\ge1}(v-ic)_+=B_c(v),
\]
so with $B$ nondecreasing, $\hat x^\circ+B_{c+2\epsilon}(\hat v^\circ)\le x_{t+1}+\epsilon+B_c(v_{t+1})\le h+\epsilon$, and symmetrically at the lower end.  Hence $u_t\in\mathcal F^{\mu^\dagger}(\hat x_t,\hat v_t)$, $\bar u_t=u_t$, $\xi_{t+1}=\xi^\circ$, and $\psi_{t+1}=\xi_{t+1}-\xi_t$ gives $\lvert\psi_{t+1}\rvert\le2\epsilon$.  A larger margin admits the same requests and executes the same trajectory.

\emph{(b).}  Write $\epsilon=\epsilon_R$.  Since $q_{t+1}=q_t\delta_t\gamma_t$, the request is $W_t=\hat\delta_t^{-1}\xi_t\delta_t\gamma_t=c_{\hat\delta_t}(\xi_t)\,\psi_t\,\gamma_t$, and the angle is invariant under conjugation and subadditive, so $\theta(W_t)\le\epsilon+2\epsilon+c_R$ and $W_t\in\mathcal B_R(\rho)$.  The cover supplies $u_t$ with $\theta(u_t^{-1}W_t)\le\epsilon$.  Executed unclipped, it gives $\xi_{t+1}=u_t^{-1}W_t$, so $\theta(\xi_{t+1})\le\epsilon$; and $\xi_{t+1}=\hat\delta_{t+1}^{-1}\xi_t\delta_{t+1}=c_{\hat\delta_{t+1}}(\xi_t)\,\psi_{t+1}$ gives $\theta(\psi_{t+1})\le2\epsilon$.  Feasibility follows: $\theta(u_t)\le c_R+3\epsilon$ is within the widened change bound, and $\theta(\hat\delta_{t+1})\le\theta(\delta_{t+1})+\theta(\psi_{t+1})\le b_R+2\epsilon$ is within the widened increment bound, which is all $\mathcal V_2^\mu$ asks on a group without boundary.  So $\bar u_t=u_t$ and the induction closes.  The count is the upper bound of Corollary~\ref{prop:so3-exact}, whose maximal $\epsilon_R$-separated subset of $\mathcal B_R(\rho)$ is such a $\mathcal U$.
\end{proof}

These are the grids of Section~\ref{sec:actioncodec2}: part (a) on each translation axis and part (b) on orientation, the latter built as in Corollary~\ref{prop:so3-exact} from a lattice in logarithmic coordinates, radially projected onto the ball of radius $\rho$; the projection keeps the cover because projection onto a Euclidean ball is nonexpansive and $\exp$ is $1$-Lipschitz.

\subsection{Computing the optimum on translation}
\label{sec:second-compute}

Theorem~\ref{thm:second-general-test} tests a given primitive set; this section finds the smallest on translation, where every object is a polytope or an interval.  We carry the pair $y=(x,v;\hat x,\hat v)$ in the error coordinates $(x,v,\xi,\psi)$ of \eqref{eq:second-translation-error}.  Two lemmas reduce the search over all finite subsets of $\R$ to a compact box.

\begin{lemma}[Polytope form and finite termination]
\label{lem:v2-polytope}
Let $c>0$, put $N_b=\lfloor b/c\rfloor+1$, and for $N=1,\ldots,N_b$ let $f_N(v)=(N-1)v-cN(N-1)/2$.  Then $B(v)=\max_{1\le N\le N_b}f_N(v)$ for every $|v|\le b$, so \eqref{eq:second-translation-viability} reads
\begin{equation}
\mathcal V_2=\bigl\{(x,v):\ |v|\le b,\ \ x-v\in[\ell,h],\ \ x+f_N(v)\le h,\ \ x-f_N(-v)\ge\ell,\ \ N=1,\ldots,N_b\bigr\}.
\label{eq:v2-polytope}
\end{equation}
Consequently (i) $\mathcal V_2$ is a polytope with at most $2N_b+4$ faces; (ii) writing $\widetilde{\mathcal V}_2=\{(x,v):x\in[\ell,h],|v|\le b,x-v\in[\ell,h]\}$ and iterating $\mathcal V^{(0)}=\widetilde{\mathcal V}_2$, $\mathcal V^{(m+1)}=\{(x,v)\in\mathcal V^{(m)}:\exists|g|\le c,\ (x+v+g,v+g)\in\mathcal V^{(m)}\}$, one has $\mathcal V^{(m)}=\mathcal V_2$ for every $m\ge N_b-1$; and (iii) $\mathcal F(x,v)$ is a nonempty compact interval whose endpoints are finite minima and maxima of affine functions of $(x,v)$.  Replacing $([\ell,h],b,c)$ by $([\ell-\mu_x,h+\mu_x],b+\mu_v,c+\mu_g)$ gives $\mathcal V_2^\mu$ and $\mathcal F^\mu$ in the same form.
\end{lemma}

\begin{proof}
Since $f_{N+1}(v)-f_N(v)=v-Nc$, the sequence $f_N(v)$ increases while $N<v/c$ and decreases afterwards, so its maximum over $N\ge1$ sits at $N=\lfloor v/c\rfloor+1$ and equals $\sum_{i=1}^{\lceil v/c\rceil-1}(v-ic)=B(v)$ for $v>0$; for $v\le0$ every $f_N(v)\le f_1(v)=0=B(v)$.  The state constraint $|v|\le b$ caps the active index at $N_b$, giving \eqref{eq:v2-polytope} and (i).

For (ii), write $\mathcal V^{\rm cl}$ for the right-hand side of \eqref{eq:second-translation-viability}.  If $(x,v)$ admits $N_b-1$ admissible steps inside $\widetilde{\mathcal V}_2$ and $v>0$, put $N=\lceil v/c\rceil\le N_b$; from $|g_i|\le c$ we get $v_i\ge v-ic>0$ for $i\le N-1$, hence $x_{N-1}=x+\sum_{i\le N-1}v_i\ge x+B(v)$, and admissibility forces $x+B(v)\le h$.  The case $v<0$ is symmetric and $v\le0$ is trivial, so $\mathcal V^{(N_b-1)}\subseteq\mathcal V^{\rm cl}$.  Conversely $\mathcal V^{\rm cl}$ is forward-invariant under braking: for $v>0$ take $g=-\min\{c,v\}$, so $v'=\max\{v-c,0\}$ satisfies $v'+B(v')=B(v)$, whence $x'+B(v')=x+v'+B(v')=x+B(v)\le h$, while $v'\ge0$ gives $x'-B(-v')=x'\ge x\ge\ell$ and $x'-v'=x\in[\ell,h]$.  So $\mathcal V^{\rm cl}$ is a set of states with infinite admissible continuations, giving $\mathcal V^{\rm cl}\subseteq\mathcal V_2$; with $\mathcal V_2\subseteq\mathcal V^{(m)}$ for all $m$ the three sets coincide from $m=N_b-1$ on.

For (iii), $\mathcal F(x,v)$ intersects $[-c,c]$ with the preimage of the polytope \eqref{eq:v2-polytope} under the affine map $g\mapsto(x+v+g,v+g)$, an intersection of finitely many half-lines in $g$; it is nonempty because $\mathcal V_2$ consists of states with an admissible continuation.
\end{proof}

\begin{lemma}[The search domain is a compact interval]
\label{lem:search-domain}
Let $\mathcal J^\mu=[-c-\mu_g,\,c+\mu_g]$.  Replacing any finite $\mathcal U\subset\R$ by $\Pi_{\mathcal J^\mu}(\mathcal U)$ changes no executed value at any reconstructed state, hence leaves $Q_\infty$ unchanged, and does not increase cardinality.  Restricting the minimization in \eqref{eq:second-general-optimum} to $\mathcal U\subseteq\mathcal J^\mu$ therefore loses no optimality.
\end{lemma}

\begin{proof}
$\mathcal F^\mu(\hat x,\hat v)\subseteq\mathcal C^\mu=\mathcal J^\mu$ at every reconstructed state.  For nested closed intervals $I\subseteq J$ the projections compose, $\Pi_I=\Pi_I\circ\Pi_J$; taking $I=\mathcal F^\mu(\hat x,\hat v)$ and $J=\mathcal J^\mu$ gives $\Pi_I(u)=\Pi_I(\Pi_{\mathcal J^\mu}(u))$, so $u$ and its projection execute identically at every state.
\end{proof}

Testing a fixed $\mathcal U$ is exact.  The executed value $\bar u_k=\Pi_{\mathcal F^\mu(\hat x,\hat v)}(u_k)$ does not depend on $g$, so every coordinate of $y^+(g,u_k)$ is affine in $g$, and the real changes answered by $u_k$,
\begin{equation}
\mathcal F_k(y;Q_n):=\{g\in\mathcal F(x,v):\ y^+(g,u_k)\in Q_n\},
\label{eq:answered-set}
\end{equation}
form a finite union of closed intervals, computable exactly by Lemma~\ref{lem:v2-polytope}.  The deletion test of \eqref{eq:second-general-kernel} at $y$ is then the one-dimensional covering $\mathcal F(x,v)\subseteq\bigcup_{k\le K}\mathcal F_k(y;Q_n)$.  An implementation that retains whole cells $\mathcal N$ of states must keep the two quantifiers joint: it subdivides $\{(y,g):y\in\mathcal N,\ g\in\mathcal F(x,v)\}$ and certifies by interval arithmetic that each piece is answered by some symbol, rather than asking one symbol to answer a given $g$ at every state of the cell.

Searching is a branch and bound.  By Lemma~\ref{lem:search-domain} the candidates at budget $K$ form the compact box $\{u\in(\mathcal J^\mu)^K:u_1\le\cdots\le u_K\}$.  For a sub-box $X_1\times\cdots\times X_K$, letting each use of the $k$th symbol re-select its value anywhere in $X_k$ gives a relaxation at least as capable as any $\mathcal U$ in the box, so if its kernel misses $\mathcal I$ the whole box is pruned soundly; the relaxed vocabulary is infinite, so its kernel serves only as the outer approximation that pruning needs.  A concrete $\mathcal U$ that passes the certified test is a witness.  Feasibility at budget $K$ is monotone in $K$, so bisection on $K$ is sound, and both endpoints are explicit.  For the lower one put $L=\max_{(x,v)\in\mathcal V_2}\lvert\mathcal F(x,v)\rvert$, attained because the graph of $\mathcal F$ is compact and its length is upper semicontinuous; at exact initialization $\xi^+=g-\bar u$, and each executed value keeps at most an interval of length $2\epsilon$ of $\mathcal F(x,v)$ inside the tolerance, whence
\begin{equation}
K_2^\star(\mu)\ \ge\ \max\bigl\{1,\ \lceil L/2\epsilon\rceil\bigr\}.
\label{eq:compute-lower}
\end{equation}
The upper one is Proposition~\ref{prop:margin-witness}(a) when $\mu\ge\mu^\dagger$.

\begin{algorithm}[t]
\caption{Optimal second-order primitive set for the translational instance}
\label{alg:second-optimum}
\small
\begin{algorithmic}[1]
\Procedure{Optimum}{$\ell,h,b,c,\epsilon,\mu$}
  \Require $c>0$, $\epsilon>0$, $\mu\ge0$
  \State build $\mathcal V_2,\mathcal V_2^\mu$ and $\mathcal F,\mathcal F^\mu$ from \eqref{eq:v2-polytope} \Comment{Lemma~\ref{lem:v2-polytope}: polytopes, $\lfloor b/c\rfloor$ steps}
  \State $\mathcal J^\mu\gets[-c-\mu_g,\,c+\mu_g]$ \Comment{Lemma~\ref{lem:search-domain}: no loss of optimality}
  \State $\textit{lo}\gets\max\{1,\lceil L/2\epsilon\rceil\}$ \Comment{\eqref{eq:compute-lower}}
  \If{$\mu\ge\mu^\dagger$}
    \State $\textit{hi}\gets\lceil c/\epsilon\rceil+3$; \ $\mathcal U^\star\gets$ the grid of Proposition~\ref{prop:margin-witness}(a)
  \Else
    \State $\textit{hi},\mathcal U^\star\gets$ any certified feasible set \Comment{Proposition~\ref{prop:margin-witness} does not apply}
  \EndIf
  \While{$\textit{lo}<\textit{hi}$}
    \State $K\gets\lfloor(\textit{lo}+\textit{hi})/2\rfloor$
    \State $(\textit{status},\mathcal U)\gets$\Call{Feasible}{$K$} \Comment{branch and bound over $(\mathcal J^\mu)^K$}
    \If{$\textit{status}=\textsc{yes}$} \State $\textit{hi},\mathcal U^\star\gets K,\mathcal U$
    \ElsIf{$\textit{status}=\textsc{no}$} \State $\textit{lo}\gets K+1$
    \Else\ \Return $[\textit{lo},\textit{hi}],\ \mathcal U^\star$ \Comment{certified integer gap; no branch may be taken}
    \EndIf
  \EndWhile
  \State \Return $\textit{lo},\ \mathcal U^\star,\ Q_\infty(\mathcal U^\star)$
\EndProcedure
\end{algorithmic}
\end{algorithm}

Algorithm~\ref{alg:second-optimum} assembles these pieces.  When it returns a single integer, the witness $\mathcal U^\star$ with $Q_\infty(\mathcal U^\star)\supseteq\mathcal I$ and a certified prune of the whole box at budget $K^\star-1$ together certify optimality, and the encoder is read off the kernel by part (2) of Theorem~\ref{thm:second-general-test}.  If the optimum sits on a non-robust boundary, no finite refinement need reach either verdict, and the algorithm returns the surviving integer interval rather than reading a failed search as infeasibility.

On $\SO(3)$, Theorem~\ref{thm:second-general-test} and Proposition~\ref{prop:margin-witness}(b) still apply.  The interval-projection and polytope reductions above do not transfer directly, so we use the explicit bound rather than this search.

\section{Comparison with chunk tokenizers}
\label{sec:compare}

\proofhere

A chunk tokenizer obtains compression by buffering $\tau$ control steps, encoding the block jointly, and releasing it as a unit, whereas a streaming tokenizer releases every control step when it arrives. This appendix bounds the number of bits that joint encoding can save under the same per-step tolerance. For $\beta_j=16$, the translation example of Section~\ref{sec:euclidean}, buffering saves at most $0.09$ bits per step and axis (Proposition~\ref{prop:chunk-band}). A FAST-style transform emits the same number of symbols as streaming, $\tau-1$ steps later, and needs additional bits to meet the tolerance (Theorem~\ref{thm:fast-dominated}). A learned codebook has a fixed number of distinct blocks, which covers at most a $2^{-304}$ fraction of the legal motions in the same example (Eq.~\eqref{eq:vq-budget}).

\paragraph{Setup.}
Everything is stated for the box model of Section~\ref{sec:euclidean}: increments in $\mathcal A=\prod_j[-b_j,b_j]$, per-step tolerance $\mathcal E=\prod_j[-\epsilon_j,\epsilon_j]$, and the ratio $\beta_j=b_j/\epsilon_j$---how many tolerance widths one control step can travel along axis $j$.  Two volume ratios recur, $\mathcal B=\prod_j\beta_j=\Vol(\mathcal A)/\Vol(\mathcal E)$ and $\mathcal B_+=\prod_j(1+\beta_j)$.  The streaming optimum is known exactly (Corollary~\ref{thm:translation-exact}): $K^\star(1)=\prod_jN_j$ symbols, $N_j=\lceil1+\beta_j\rceil$, at $R(1)=\sum_j\log_2N_j$ bits per step.  For a chunk decoder releasing $\tau$ steps at a time, $K^\star(\tau)$ is the smallest dictionary of reconstructed blocks that covers within tolerance, at every step of the block, every legal path (residual in $\mathcal E+\mathcal A$, increments in $\mathcal A$), the final constraint returning the residual to $\mathcal E$ so that blocks concatenate indefinitely; its rate is $R(\tau)=\tau^{-1}\log_2K^\star(\tau)$, and $R(1)-R(\tau)$ is the saving obtained by delaying release by $\tau-1$ steps.  The bounds do not assume a memoryless decoder: the legal paths do not depend on the state the previous block left behind, so every decoder state faces the same covering problem and the bounds hold for every decoder architecture.

\paragraph{Joint encoding.}
Every buffered step still costs at least $\log_2\mathcal B$ bits, so longer blocks do not reduce the per-step cost below this value.

\begin{proposition}[Saving of joint encoding]
\label{prop:chunk-band}
$K^\star(1)=\prod_jN_j$ and $K^\star(\tau+\tau')\le K^\star(\tau)K^\star(\tau')$, so $R(\tau)\le R(1)$ and $R(\tau)\to R_\infty=\inf_\tau R(\tau)$; moreover
\begin{equation}
\begin{gathered}
K^\star(\tau)\ \ge\ \mathcal B_+\,\mathcal B^{\,\tau-1},
\qquad\text{hence}\qquad
\log_2\mathcal B\ \le\ R(\tau)\ \le\ R(1),\\
\text{so}\qquad
R(1)-R_\infty
\ \le\ \sum_{j=1}^{d}\log_2\frac{\lceil1+\beta_j\rceil}{\beta_j}
\ \le\ \sum_{j=1}^{d}\log_2\Bigl(1+\frac2{\beta_j}\Bigr).
\end{gathered}
\label{eq:band}
\end{equation}
\end{proposition}

\begin{proof}
Concatenating an optimal $\tau$-code with an optimal $\tau'$-code covers the $(\tau+\tau')$ paths, so $\log_2K^\star$ is subadditive and Fekete's lemma applies.  Each reconstruction covers one translate of $\mathcal E^{\,\tau}$, while the legal paths shear to $(\mathcal E+\mathcal A)\times\mathcal A^{\tau-1}$, so covering them forces $K^\star(\tau)\Vol(\mathcal E)^\tau\ge\Vol(\mathcal E+\mathcal A)\Vol(\mathcal A)^{\tau-1}$; divide by $\Vol(\mathcal E)^\tau$ and take $\tau^{-1}\log_2$, using $\lceil1+\beta\rceil\le\beta+2$ for the last line.
\end{proof}

For example, at $\beta_j=16$ the saving of joint encoding is $\log_2(17/16)=0.09$ bits per step per axis, against a streaming cost of $\log_217=4.09$.  When $\beta_j<1$, one step cannot cross its own tolerance, the integer ceiling gives $N_j=2$, and the bound is weak; the deployed tokenizers do not operate in this regime. The next two paragraphs show that the DCT quantizer needs more bits per step than $R(1)$, and that a learned codebook covers at most a $2^{-304}$ fraction of the legal paths in the same example.

\paragraph{DCT quantization (FAST).}
FAST~\citep{fast} normalizes a block, applies a temporal DCT, rounds the coefficients, and runs BPE.  The DCT is invertible, so the quantization stage emits exactly $\tau$ integers per axis per block---one per control step, the same count as streaming, delivered $\tau-1$ steps late.  The token saving that FAST reports comes from its BPE stage, which is lossless and corresponds to the packing of Section~\ref{sec:setbpe}; the same packing applies to our symbol stream, so we compare the two at the quantization stage.  There, rounding in the frequency basis makes the error cell a cube rotated by the DCT, and a rotated cube must be smaller than the axis-aligned tolerance box to fit inside it.

\begin{theorem}[A FAST-style quantizer is dominated by streaming]
\label{thm:fast-dominated}
Assume $\beta_j>2$ for every axis.  Let $\Lambda_\tau:=\max_l\sum_k|C_{kl}|$ be the largest column $\ell_1$ norm of the orthonormal DCT-II; $\sqrt{\tau/2}\le\Lambda_\tau\le\sqrt\tau$.  If the quantizer $\hat q=\alpha^{-1}C^\top\operatorname{round}(\alpha Cq)$, with any per-axis scales $\alpha_j$, meets the per-step tolerance on every legal path, then it spends at least
\begin{equation}
R_{\rm DCT}(\tau)-R(1)\ \ge\ \frac d2\log_2\frac\tau2-\sum_{j=1}^{d}\log_2\Bigl(1+\frac2{\beta_j}\Bigr)
\label{eq:fast-gap}
\end{equation}
bits per control step more than streaming.
\end{theorem}

\begin{proof}
On each axis write $s=\alpha_j^{-1}$, $b=b_j$, and $\epsilon=\epsilon_j$.  If $s\Lambda_\tau>2\epsilon$, choose a maximizing column $l$, $\sigma_k=\operatorname{sign}(C_{kl})$, and $\epsilon/\Lambda_\tau<t<\min\{s/2,b/(2\Lambda_\tau)\}$.  Then $q=tC^\top\sigma$ is a legal path, since $|q_1|<b/2$ and $|q_{i+1}-q_i|<b$, yet rounds to zero with $\|q\|_\infty=t\Lambda_\tau>\epsilon$.  Hence $s\le2\epsilon/\Lambda_\tau$.  Cells of volume $\prod_j\alpha_j^{-\tau}$ must then cover the legal paths, of volume $\prod_j(2b_j+2\epsilon_j)(2b_j)^{\tau-1}$, so at least $\Lambda_\tau^{d\tau}\mathcal B_+\mathcal B^{\tau-1}$ reconstructions are used; take $\tau^{-1}\log_2$ and subtract $R(1)\le\log_2\mathcal B+\sum_j\log_2(1+2/\beta_j)$, using $\Lambda_\tau\ge\sqrt{\tau/2}$.
\end{proof}

At $\beta_j=16$ on the three translation axes, \eqref{eq:fast-gap} is at least $2.4$ bits per step already at $\tau=8$ and at least $6.9$ at $\tau=64$; no per-axis scale removes it, because $\Lambda_\tau$ grows with $\tau$.

\paragraph{Learned codebooks.}
Learned chunk codebooks---the OAT~\citep{oat}, X-Tokenizer~\citep{x-tokenizer}, and ACodec~\citep{dong2026actioncodec} baselines---encode a block into $m$ symbols over a vocabulary of size $S$ and decode with a learned network.  The compression comes from using $m\ll\tau$ symbols per block, which limits the number of distinct blocks.  For any parameters, it can emit at most $M\le S^m$ distinct blocks, each covering one translate of the tolerance box, so the fraction of legal paths reconstructible within tolerance is
\begin{equation}
\gamma_\theta\ \le\ \frac{M}{\mathcal B_+\mathcal B^{\,\tau-1}}
\label{eq:vq-budget}
\end{equation}
for every parameter setting $\theta$, and the bound is fixed before training.  For $S=1024$, $m\le8$, and a $32$-step block on the three translation axes at $\beta_j=16$, it gives $\gamma_\theta\le2^{-304}$.  Training chooses which cells are covered: if the legal paths are split into a training region and its complement, the covered fractions of the two regions trade off linearly, so higher coverage of the training region reduces coverage elsewhere.  In our study, OAT is learned from scratch on each dataset, X-Tokenizer is fine-tuned on it, and the released ACodec is trained on LIBERO, BridgeData, and DROID (Appendix~\ref{app:tokenizers}), whereas the pretrained streaming instance is applied zero-shot across embodiments.  A worst-case guarantee, by \eqref{eq:vq-budget}, forces $M\ge K^\star(\tau)$: the streaming rate.

\paragraph{Summary.}
The number of bits per step needed for a tolerance guarantee is determined by the step range $b_j$ and the tolerance $\epsilon_j$ through $\beta_j$, not by how symbols are grouped. Joint encoding of $\tau$ steps saves at most $\sum_j\log_2\bigl(\lceil1+\beta_j\rceil/\beta_j\bigr)$ bits per step ($0.09$ per axis at $\beta_j=16$). The FAST-style quantizer needs at least $2.4$ to $6.9$ additional bits per step at $\beta_j=16$ on three translation axes, for $\tau=8$ to $64$. A learned codebook with the parameters above covers at most $2^{-304}$ of the legal paths. Streaming composition holds the tolerance on all legal paths.

\section{Lossless packing after physical tokenization}
\label{sec:lossless}

\proofhere

Everything here starts from the already fixed primitive sequence: a packer may change how that sequence is written for an AR model, but not which primitive was chosen or which state was reconstructed.

At step $t$ the emitted token instances $T_t$ partition the step's joint assignment,
\begin{equation}
\biguplus_{C\in T_t}\Exp(C)=S_t,
\label{eq:spatial-partition}
\end{equation}
every atom covered exactly once, and the weighted count of a pair of token types over training steps is
\begin{equation}
F(A,B)=\sum_t w_t\,\mathbf 1\{A\in T_t,\ B\in T_t\}.
\label{eq:spatial-frequency}
\end{equation}
A temporal instance starts at a step $a$ and covers occurrences at nonnegative offsets within the horizon $\tau$; the stream orders the final instances by start step and, at equal starts, by the smallest coordinate of the leading spatial token.

\begin{proposition}[Set-BPE is lossless and entropy-preserving]
\label{prop:setbpe}
Every merge of Algorithm~\ref{alg:setbpe} replaces two token instances by one whose expansion is their disjoint union.  The emitted instances therefore partition the episode's assignments, the stream is a strict total order that the decoder inverts without timestamps or separators, the trained packer is a bijection with $H(Y)=H(Z)$ for every source distribution and with the reconstructed trajectory of Corollaries~\ref{thm:translation-exact} and \ref{prop:so3-exact} unchanged, and each applied merge removes exactly one emitted token per affected occurrence, so the greedy scores in Algorithm~\ref{alg:setbpe} are the exact one-step gains.
\end{proposition}

\begin{proof}
A spatial merge replaces two disjoint expansions by their union, so \eqref{eq:spatial-partition} keeps returning $S_t$ and the step holds one token fewer; a temporal merge replaces two disjoint subsets of the episode's occurrences by their union, so induction from the singletons keeps the emitted instances a partition.  Two final instances never share a leading occurrence, and at a fixed start the leading tokens have disjoint coordinate supports, so the serialization order is strict.  An instance covers only steps from its start onward, so when the earliest incomplete step is $a$, the next instance in order leads there and its root sits at the lowest coordinate of $a$ still unassigned; decoding therefore advances over completed steps with no timestamp or separator, and expansion restores every $S_t$.  The packer is thus injective, the trajectory and its distribution are unchanged, and each merge's drop in emitted tokens is exact: a spatial rule's weighted count falls by exactly $F(A,B)$, and the largest reduction of a temporal rule $(P,R,\delta)$ is the maximum matching of its instance graph, which for a fixed offset has degree at most two---time-ordered paths on which left-to-right greedy selection is exact.
\end{proof}

Appendix~\ref{app:bpe-training} describes the exact sparse and lazy implementations of this loop.

\end{document}